\documentclass[12pt]{article}
\usepackage[total={6.5in, 9.5in}]{geometry}

\usepackage{amsmath}
\usepackage{mathtools}
\usepackage{amsfonts}
\usepackage{amsthm}
\usepackage{thmtools}

\usepackage[T1]{fontenc}
\usepackage{newtxtext}
\usepackage{newtxmath}
\usepackage[bb=boondox,cal=boondox]{mathalpha}

\usepackage[usenames,dvipsnames]{xcolor}

\usepackage{graphicx}
\usepackage[labelfont=bf]{caption}
\usepackage[format=hang]{subcaption}

\usepackage{algpseudocode}
\usepackage{algorithm}

\usepackage{enumitem}

\usepackage{booktabs}
\usepackage{multirow}
\usepackage{longtable}
\usepackage{etoolbox,siunitx}
\robustify\bfseries
\usepackage{pdflscape}
\usepackage{makecell}

\usepackage[bf,tiny]{titlesec} 

\usepackage{tikz}
\usetikzlibrary{bayesnet}

\usepackage{natbib}

\newtheorem{assumption}{Assumption}
\newtheorem{proposition}{Proposition}
\newtheorem{theorem}{Theorem}
\newtheorem{definition}{Definition}

\usepackage[colorlinks,linktoc=all]{hyperref}
\usepackage[all]{hypcap}
\hypersetup{
    citecolor=MidnightBlue,
    urlcolor=MidnightBlue,
    linkcolor=black
}

\usepackage{cleveref}
\crefname{assumption}{assumption}{assumptions}
\Crefname{assumption}{Assumption}{Assumptions}

\newcommand{\parhead}[1]{\vspace{0.1in} \noindent \textbf{#1}. \,\,}

\newcommand{\rmdo}{\mathrm{do}}
\newcommand{\s}{\, ; \,}
\newcommand{\g}{\,\vert\,}
\newcommand{\pr}{\mathrm{p}}
\newcommand{\f}{\mathrm{f}}
\newcommand{\di}{\mathrm{d}}
\newcommand{\X}{\mathcal{X}}
\newcommand{\Bc}{\mathcal{B}}
\newcommand{\con}{\mathrm{con}}
\newcommand{\cau}{\mathrm{cau}}
\newcommand{\ateest}{\widehat{\textsc{ate}}}

\newcommand{\mba}{\mathbf{a}}
\newcommand{\mbz}{\mathbf{z}} \newcommand{\add}[1]{{{#1}}}
\newcommand{\citepapx}{\citep}
\newcommand{\citeapx}{\cite}

\begin{document}

\title{Estimating the Causal Effects of T Cell Receptors}

\author{Eli N. Weinstein\thanks{Department of Chemistry, Technical University of Denmark, Lyngby, DK. Contact: enawe@dtu.dk}\textsuperscript{\,\,\,$\dagger$}, Elizabeth B. Wood\thanks{Jura Bio, Boston, MA}, David M. Blei\thanks{Department of Computer Science and Department of Statistics, Columbia University, New York, NY. Contact: david.blei@columbia.edu}}\date{\today}

\maketitle

\begin{abstract}
	
A central question in human immunology is \add{how a patient's T cell receptors impacts disease}. Here, we introduce a method to infer the causal effects of T cell receptor (TCR) sequences on patient
  outcomes using observational TCR sequencing data and
  clinical outcomes data. Our approach corrects for unobserved
  confounders, such as a patient's environment and life history,
  using the patient's pre-selection TCR repertoire. This
  pre-selection repertoire is generated by a \add{programmed stochastic
  recombination process, V(D)J recombination}, which provides a natural
  experiment. We first derive a causal identification result that leverages biological theory to semiparametrically constrain a hierarchical causal model, enabling causal inference.
  \add{We then develop a causal estimation strategy that uses permutation invariant neural networks and representation learning to scale to millions of sequences from hundreds of patients.
  Given sequence data,
  our method produces an
  estimate of the effect of interventions that add a specific TCR
  sequence to patient repertoires.} 
  On semisynthetic data, we demonstrate that our method can correct for unobserved confounding.
  We use it to analyze the effects of TCRs on COVID-19 severity, uncovering
  potentially therapeutic TCRs that are (1) observed in patients, (2)
  bind SARS-CoV-2 antigens \textit{in vitro} and (3) have strong
  positive effects on clinical outcomes.
 \end{abstract}

Individual T cells' ability to respond to disease depends crucially on
their T cell receptor (TCR), a protein on the surface of the cell
\add{that} recognizes antigens such as viral proteins. A patient's
collection or \textit{repertoire} of diverse TCRs can shape the
course of their disease, whether infections, cancers, or other
conditions.

In this paper, we develop a method to estimate the causal effects of
TCRs on clinical outcomes. We are not merely interested in which TCRs
are associated with a disease, or which TCRs recognize disease-related
antigens. Rather, our goal is to predict what would happen to patients
if we \textit{intervene} on them by adding a particular TCR to their
repertoires. An accurate causal
map of TCR sequences would provide a deeper understanding of human
disease and directly inform the design of TCR-T cell therapies, TCR bispecifics, T cell
vaccines, and other medicines~\citep{Baulu2023-vt,Klebanoff2023-bx}.

How can we estimate the causal effects of TCRs without actually
intervening on patients? Advances in high-throughput sequencing has
allowed hundreds of thousands of TCRs to be sequenced from individual
donors, providing a detailed picture of patients'
repertoires~\citep{Freeman2009-ej,Robins2012-zj}. Using this
technology, several studies have collected TCR repertoire sequences
from large cohorts of patients with different diseases, conditions,
and
outcomes~\citep{Nielsen2018-hz,Davis2019-ec,Joshi2022-oe,Mhanna2024-il}.
We derive a way to analyze such \textit{observational
  data}---observed repertoire sequences and clinical outcomes---to
estimate the effects of TCR interventions.

The fundamental challenge to estimating causal effects from
observational studies is \textit{unobserved confounders}, unmeasured
variables that affect both TCR repertoires, which is the treatment
variable, and patient outcomes. Confounders lead to correlation
without causation, so that the TCRs associated with an outcome are not
necessarily responsible for that outcome. Confounding is a particular
threat to repertoire studies because the human immune system is
adaptive, and possesses a memory of past diseases and exposures.

Consider a healthcare worker who is repeatedly exposed to
a variety of infectious diseases. They are likely to possess T cells
protective against these previous infections, which may be unrelated
to their current disease. But they
may also be more likely to have appropriate diagnosis and treatment
for their disease, compared to those who do not work in healthcare. So
a patient's job can affect both their TCR repertoire and their disease
outcome. Consequently, TCRs that are protective against a previous
disease can be associated with clinical outcomes in a current disease,
without causing those outcomes.

\add{

To address this confounding, we look back into the TCR repertoire's biological history. 
Each TCR is first created by a programmed genetic recombination process, \textit{V(D)J recombination}, which stochastically generates a TCR gene through enzymatic reactions.
If the TCR gene is \textit{productive}, it is transcribed into a functional TCR protein expressed on the cell surface. There, it encounters ongoing \textit{selection} processes that can increase or decrease the number of cells with that TCR. 
Some selection happens in the thymus, where, for example, TCRs that recognize self antigens are eliminated.
Other selection processes happen later, when the TCR encounters a foreign antigen, and its population expands to mount an immune response. Selection processes are how confounders such as past infections shape the repertoire, and create immunological memory.

To correct for confounding, we make use of data about the repertoire before it undergoes selection. 
Some of the TCR genes produced by V(D)J recombination are \textit{nonproductive}, meaning they have truncation or frameshift mutations that disable the TCR protein, rendering it unable to recognize antigens. 
\citet{Murugan2012-zg} show that data from nonproductive TCRs can be used to infer the \textit{pre-selection repertoire}: the collection of productive TCRs produced by V(D)J recombination before any selection.
(In contrast to the TCR repertoire, defined as the collection of all TCRs in a patient at a given moment in time, the term \textit{pre-selection repertoire} refers to an inferred constant, namely, the distribution of TCRs that a person's V(D)J recombination process can produce before selection \citep{Murugan2012-zg}.)
So, we can learn about a patient's repertoire before memory and confounding enter.

Our key idea is to use the pre-selection repertoire as an \textit{instrumental variable} (IV), a source of randomness affecting the treatment that can be used to correct for confounding~\citep{Imbens1994-pk}. 
The pre-selection repertoire is generated by V(D)J recombination, providing natural randomization. It affects the current repertoire but is unaffected by confounders, since selection and immunological memory do not enter. We use the pre-selection repertoire as an IV to correct for confounding and estimate TCRs' causal effects.

}

\begin{figure*}[t]
\centering
\includegraphics[width=\textwidth]{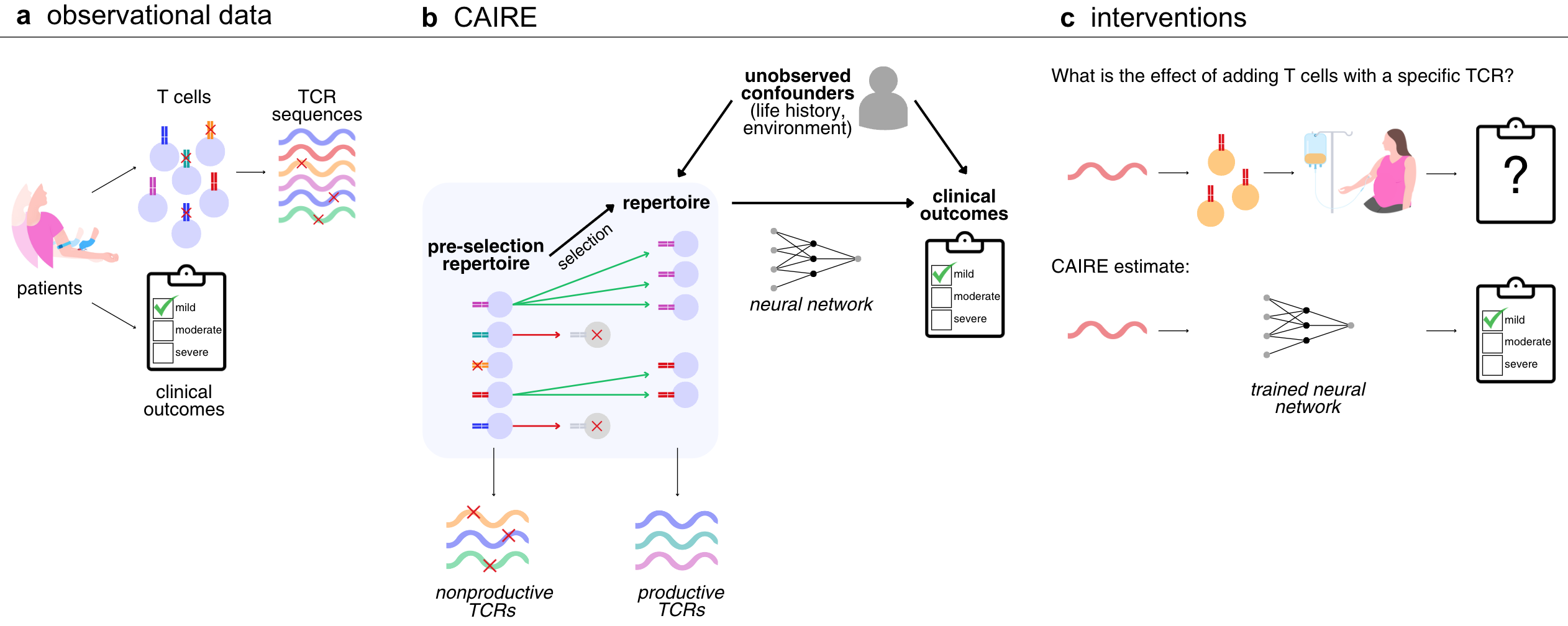}
\caption{
\textbf{Estimating the causal effects of TCRs with CAIRE.} 
\add{(a) CAIRE uses TCR repertoire sequencing and clinical outcomes data from patients. (b) CAIRE trains a neural network to estimate the effect of TCR repertoires on clinical outcomes. 
Unobserved confounders such as previous diseases affect the repertoire through a process of natural \textit{selection}, in which specific TCRs increase or decrease in number over time, encoding immunological memory.
CAIRE uses the \textit{pre-selection repertoire}, the repertoire created by V(D)J recombination before any selection, as an instrumental variable to correct for confounding.
It uses data from \textit{nonproductive} TCR genes, meaning those with protein-disabling mutations, to infer the pre-selection repertoire \citep[applying][]{Murugan2012-zg}.
(c) CAIRE provides an estimate of the effect of adding T cells with a specific TCR to patients.}} \label{fig:overview}
\end{figure*}

\parhead{Contribution} We develop CAIRE (causal adaptive immune
receptor effect estimator), a method to estimate the causal effects of
TCRs on clinical outcomes (\Cref{fig:overview}). 
We first show formally how information about
patients' pre-selection repertoire can be used to correct for
unobserved confounders. 
Then we develop an estimation algorithm that uses deep representation learning to scale to large observational datasets. On semisynthetic data, we demonstrate that CAIRE can provide accurate
estimates of the effects of individual TCRs on patient outcomes, even
when non-causal methods are led astray. 
We apply CAIRE to estimate the
effects of TCRs on COVID-19. 
These estimates show how different TCRs within a patients' repertoire contribute to the severity of their disease, and uncover potential therapeutic and vaccine candidates.

\parhead{Related Work} \label{sec:related}
Previous work on observational repertoire sequencing data has
considered the problem of predicting a patient's past or present
disease from their repertoire sequences, and learning which TCRs are
associated with that disease~\citep{Mhanna2024-il}. In particular,
\cite{Emerson2017-je} pioneers diagnostic methods based on high-throughput TCR sequencing, developing a test for cytomegalovirus
infection based on associations with commonly seen (\textit{public}) TCRs.
\cite{Widrich2020-hj} develops a transformer-based
approach for predicting disease status from TCR repertoires, working
in the framework of multiple instance learning.
\cite{Slabodkin2023-ba} employs a semi-supervised prediction approach,
and shows that it can also be used to predict and generate unobserved
receptor sequences associated with a disease.

Other approaches have been developed based on a variety of machine
learning techniques, such as hand-crafted features, distance-based
clustering, or representations derived from protein language
models~\citep{Mariotti-Ferrandiz2016-hu,Pavlovic2021-sn,Zaslavsky2022-qg,Liu2024-hh}.
Such methods have been applied to diseases, including COVID-19, to
diagnose the disease and to identify clusters of TCRs associated with
it~\citep{Snyder2020-qg,Schultheis2020-mj,Dannebaum2022-df,Kockelbergh2022-ci}.
As highlighted by \cite{Pavlovic2024-ic}, however, these methods for
prediction and association do not address causal questions about the
effects of interventions.

Another line of work focuses on identifying receptors in an
individual's repertoire that are under strong selective pressure. In
particular, \cite{Pogorelyy2018-ah,Pogorelyy2019-zb} compare the
mature repertoire to an estimate of the pre-selection repertoire
derived from nonproductive TCRs~\citep{Murugan2012-zg,Marcou2018-so}.
Their purpose is to identify TCRs that have reacted against antigens
the patient has encountered in the past. Here we use the pre-selection
repertoire for a different purpose, to estimate causal effects.

\cite{Pradier2023-ah} estimates causal effects \textit{on} TCR
repertoires, such as the effect of disease exposure, using deep
generative models and causal representation learning. We consider the
effects \textit{of} repertoires, and advance methods to account for
unobserved confounding.

Outside of immune repertoire studies, our work relates to causal
inference methods in human genetics. The digital twin test uses parent
and offspring genotyping data to infer the causal effects of genetic
variants~\citep{Bates2020-uo}. Like our method, it exploits genetic
recombination as a source of randomization. It uses meiotic
recombination; we use V(D)J recombination.

Our approach extends instrumental variable (IV) methods, a technique
for causal inference that has been widely used in genetics, economics,
epidemiology, and other
fields~\citep{Imbens1994-pk,Newey2003-dl,Davey_Smith2014-ai,Saengkyongam2022-se}. \add{Our
identification results and estimation methods are distinct from existing IV methods, in that our results do not rely on
assumptions about how the outcome variable is generated, e.g. we do
not assume an additive or independent residual. Instead, we use
domain-specific assumptions, stemming from the biology, to semiparametrically restrict how the instrument affects the
treatment; similar semiparametric restrictions appear in \citet{Puli2020-ud}. } Moreover,
our results are not limited to binary or continuous treatments, but
rather extend to complex, high-dimensional treatments, namely TCR
repertoires.

Finally, this work builds on recent developments in causal machine learning
methods for high dimensional and structured data.
\cite{Kaddour2021-uu} develop efficient semiparametric estimators for
the effects of structured treatment data when confounding is observed;
they focus on applications to graph data such as small molecules.
\cite{Hartford2017-rr,Xu2020-kv} and \citep{Puli2020-ud} extend semiparametric IV
methods to high-dimensional treatments, focusing especially
on images. 
\citet{Weinstein2024-cg} study causal inference in nested datasets with hierarchical structure, developing hierarchical causal models (HCMs).
We build neural IV methods for HCMs, exploiting the nested structure of immune repertoire data. 

  \section{Learning the Effects of TCRs} \label{sec:id}

\begin{figure*}
\centering
\begin{subfigure}[c]{0.36\textwidth}
\centering
\begin{tikzpicture}

\node[obs] (y) {$y_{i}$};
  \node[obs, left=1.5cm of y] (a) {$a_{ij}$};
  \node[obs, left=1cm of a] (z) {$z_{ik}$};
  \node[latent, above=.5cm of a] (r) {$r_{i}$};
  \node[latent, above=of a, xshift=1.5cm]  (u) {$u_i$};

\edge {a,u} {y} ; \edge {u} {r} ;
  \edge {r} {a} ;
  \edge {z} {a} ;

\plate[dashed] {ap} {(a)} {$m$} ;
  \plate[dashed] {zp} {(z)} {$b$} ;
  \plate {zayu} {(ap)(zp)(u)(y)} {$n$} ;

\end{tikzpicture}
\caption{} \label{fig:hcm}
\end{subfigure}
\begin{subfigure}[c]{0.36\textwidth}
\centering
\begin{tikzpicture}

\node[obs] (y) {$y_{i}$};
  \node[obs, left=1.5cm of y] (a) {$q^{a}_i$};
  \node[obs, left=1cm of a] (z) {$q^{z}_i$};
  \node[obs, above=.5cm of a] (r) {$r_{i}$};
  \node[latent, above=of a, xshift=1.5cm]  (u) {$u_i$};

\edge {a,u} {y} ; 
  \edge {u} {r} ;
  \edge {r} {a} ;
  \edge {z} {a} ;

\plate {zayu} {(a)(z)(u)(y)} {$n$} ;

\end{tikzpicture}
\caption{} \label{fig:chcm}
\end{subfigure}
\caption{\textbf{Causal graphs.} (a) Hierarchical causal model. (b) Collapsed causal model. $u_i$: unobserved confounding. $z_{ik}$: pre-selection repertoire sequences. $a_{ij}$: mature repertoire sequences. $y_i$: patient outcomes. $q^z_i$: pre-selection repertoire distribution. $q^a_i$: repertoire distribution. $r_i$: relative fitness. }
\end{figure*}
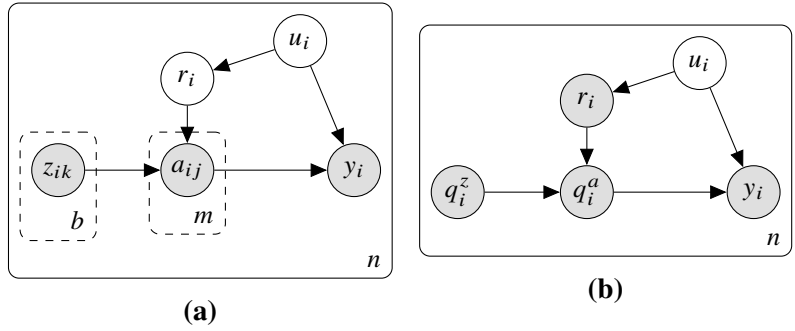
 
Our goal is to estimate the causal effect of TCRs on patients' outcomes.
A causal effect is a prediction under a hypothetical intervention.
Here the intervention is \add{to add T cells to the repertoire that have a given TCR sequence}.

There are three stages to applied causal analysis. The first is to write down our assumptions in a causal model (\Cref{sec:causal-model}). 
What are the variables we observe? What \add{relevant} variables do we not observe? 
What is the intervention we want to understand?

The second stage is to \emph{identify} the causal effect (\Cref{sec:id_details}). This means: if I observed an infinite amount of data from the model, \add{such that there is no uncertainty about its underlying distribution, how would I calculate the causal effect?}
Identification is a question about how observations relate to interventions.
Specifically, the causal model implies a probability distribution of the observed variables; the \emph{identification formula} calculates the causal effect as a function of that distribution.

Finally, with the assumptions and identification formula in place, \add{we develop an approximate inference method to estimate the effect} (\Cref{sec:estimation}).
How can we analyze finite data to estimate the causal effect?
This is the algorithm, as a culmination of the causal analysis.

\subsection{A Causal Model of Repertoires and Outcomes} \label{sec:causal-model}
Our causal model is drawn as a graph in \Cref{fig:hcm}. Nodes are
variables, edges denote a causal relationship between them, shaded
nodes are observed, and unshaded nodes are unobserved. \add{This causal
graph depicts a hierarchical causal model: the solid outer plate indicates replication of
patients in the data; the dashed inner plates indicates replication of
cells within them, with each copy of each variable having the same arrows in and out~\citep{Weinstein2024-cg}.}
We describe each variable in turn, discussing their role in
the causal model \add{and the biological motivation for the model's assumptions}. Formal mathematical details are in
\Cref{apx:hcm}.

\begin{itemize}[leftmargin=*]
\item \textit{Disease outcome: $y_i$.} The disease outcome of patient $i$ is
  $y_i$. It is a scalar measurement, such as survival. The disease
  outcome is our outcome variable.

\item \textit{TCR repertoire: $\mba_i$.} The TCR repertoire of patient $i$ is a
  collection of TCRs $\mba_i = \{a_{ij}\}_{j=1}^{m_i}$. TCR $a_{ij}$
  is a sequence of amino acids, that is, a variable-length string made
  from a 20-letter alphabet. The variable $a_{ij}$ falls within an
  inner plate of the causal graph because each patient has many T
  cells. \add{The T cells are unordered, and there can be multiple T cells with the same TCR sequence, so $\{a_{ij}\}_{j=1}^{m_i}$ is a \textit{multiset}, i.e. a set where there can be multiple of the same item.} There is an arrow from $\mba_i$ to $y_i$ because we expect
  the patient's repertoire influences their disease outcome. The
  repertoire is our treatment variable.

\item \textit{Unobserved confounders: $u_i$.} A central challenge to causal
  inference from observational data is unobserved confounders,
  variables that affect both the treatment and the outcome. For
  example, a patient's environment and life history can affect both
  their disease outcomes and T cell repertoire. In the graph, the
  unobserved confounders are $u_i$. They connect both to the disease
  outcome $y_i$ and the repertoire $\mba_i$. Unobserved confounders
  can induce spurious (i.e., non-causal) correlation between
  repertoires and disease outcomes.

\item \textit{Pre-selection repertoire $\mbz_i$.} Each patient's pre-selection
  repertoire is a collection of TCRs $\mbz_i = \{z_{ij}\}_{j=1}^{b_i}$. \add{By \textit{pre-selection repertoire} we refer to TCRs produced by V(D)J recombination prior to any selection process \citep{Murugan2012-zg,Elhanati2014-ak}.
We infer the pre-selection repertoire using techniques that exploit information about V(D)J recombination in nonproductive TCR data (\Cref{sec:estimation}) \citep{Murugan2012-zg}.} We will use the pre-selection
  repertoire as an external source of randomness that can help correct
  for unobserved confounding. In particular, we assume this repertoire
  is an \textit{instrument}, a variable that affects the treatment but
  is unaffected by confounders and does not have a direct effect on
  the outcome~\citep[][Chap. 7]{Imbens1994-pk,Newey2003-dl,Pearl2009-fh}.

  Why is it a valid instrument? The pre-selection repertoire develops
  into the current repertoire $\mba_i$, and thus shapes its content; so
  we draw an arrow from $\mbz_i$ to $\mba_i$. \add{The pre-selection
  repertoire is randomized, via V(D)J recombination, and does not respond to antigens; 
  so we assume it is not affected by
  confounders, and there is no arrow from $u_i$ to $\mbz_i$. 
    The developing cells that contain pre-selection TCRs do not themselves mount immune responses, until they develop into mature T cells~\citep[Chap. 8]{Abbas2018-zj};
thus there is no arrow from $\mbz_i$ to $y_i$, and the only path goes through $\mba_i$. 
  Finally, the pre-selection repertoire varies from person to person, and this variation impacts the mature repertoire~\citep{Slabodkin2021-wr}; thus the instrument is relevant.} 
  Taken together, these conditions imply that
  $\mbz_i$ is a valid instrument. We discuss these assumptions and their
  limits further in \Cref{sec:discussion}.

\item \textit{Natural selection and the fitness function: $r_i$.} The process
  of repertoire development is one of natural selection~\citep{Elhanati2014-ak,Abbas2018-zj}. Different
  TCRs in the pre-selection repertoire have different levels of fitness, and
  increase or decrease in number over time in response to antigens. The result of this
  selection is the current repertoire we see today.

  We explicitly account for this antigenic selection process using a latent
  variable $r_i$, which describes the growth (\textit{fitness}) of the
  TCRs in a patient's repertoire, \add{specifying how the pre-selection repertoire $\mbz$ matures into the current repertoire $\mba$. $r_i$ is a function which assigns a fitness to each sequence.} Suppose patient $i$ contracts
  influenza. If TCR $x$ recognizes the influenza virus, its population will expand, so $r_i(x)$ will be large. 
  Conversely, a TCR $x$ that recognizes patient
  $i$'s own proteins will have small $r_i(x)$, as it is destroyed to avoid autoimmunity. 
  \add{The full function $r_i$ effectively summarizes all the selective
  forces that have acted on the patient's
  repertoire over the course of their life, to create the current repertoire $\mba$.
  We formalize this mathematically in \Cref{asm:selection} below.}

  We assume that the process of antigenic selection \textit{mediates} confounding: the confounders $u_i$ only affect the mature repertoire
  $\mba_i$ through their effect on $r_i$. The reason is that this is the biological mechanism for immune memory. A patient's life history of antigen exposure \add{creates immune memory by placing selective pressure on the TCRs in their repertoire}, so
  there must be an arrow from $u_i$ to $r_i$ and from $r_i$ to
  $\mba_i$~\citep[Chap.
  8]{Abbas2018-zj}. The only other way for confounders to affect the mature repertoire is by changing the pre-selection TCR
  repertoire, and this we have already excluded; hence, there is no
  arrow from $u_i$ to $\mba_i$. In short, we assume that the effect
  that any confounder -- be it environment, life history, or another
  variable -- has on the patient repertoire boils down to an effect on
  the relative fitness of different TCRs.
\end{itemize}

Our causal model accounts for the process of selection mathematically using a population genetics
model of evolution under natural
selection~\citep{Neher2011-ls,Bertram2019-ok}. \add{\citet{Elhanati2014-ak} proposed applying this population genetics model to describe T cell development, modeling a population of T cells whose genotypes correspond to
TCR sequences~\citep{Isacchini2021-na}.}
We use $q_i^z(x)$ to denote the distribution over TCR sequences $\mbz$ in a patient's pre-selection repertoire, and $q_i^a(x)$ for the mature repertoire.
We use $\Bc = \{A, C, \ldots, Y\}$ to denote the amino acids. 
\begin{assumption}[Maturation via selection] \label{asm:selection}
	The distribution of the mature repertoire $q^a_i$ is generated from $q^z_i$ and $r_i$ as:
	\begin{equation} \label{eqn:fitness} 
	q^{a}_i(x) = \f(q^{z}_i, r_i) = \frac{r_i(x)}{\sum_{x' \in \mathcal{X}} r_i(x') q^{z}_i(x')} q^{z}_i(x),
	\end{equation}
where $\X = \cup_{\ell=1}^\infty \Bc^\ell$ is the space of TCR amino acid sequences and $r_i(x)$ is a function $\mathcal{X} \to \mathbb{R}_+$. \end{assumption}
\noindent \Cref{eqn:fitness} says that the fraction of TCRs with sequence $x$ in the mature repertoire, $q^a_i(x)$, is proportional to the fraction of TCRs with that sequence in the pre-selection repertoire, $q_i^z(x)$, times the fitness, $r_i(x)$.
The denominator normalizes the distribution, so $\sum_{x\in\mathcal{X}} q^a_i(x) = 1$.
In sum, we assume the underlying causal mechanism generating the mature repertoire $\mba$ is governed by a mechanistic model of evolution under natural selection. 
We do not make assumptions about the causal mechanisms generating the other variables in our causal graph.

\add{Overall, the core distinguishing assumption of our approach is that the pre-selection repertoire is an instrumental variable for the current repertoire.
The causal graph (\Cref{fig:hcm}) and \Cref{asm:selection} reflect textbook immunology: antigen exposure and immunological memory are encoded into the immune repertoire through a process of natural selection, in which TCR subpopulations expand and contract over time \citep{Abbas2018-zj}. 
Writing this idea as a formal causal model, we see it suggests the repertoire \textit{before} selection is an instrument. 
But biology is complex, and models are simplifications. The causal graph encodes assumptions in the arrows that are absent, and \Cref{eqn:fitness} assumes a specific functional form.
For example, if V(D)J recombination in early life is impacted by a disease or exposure, there should be an arrow from $u$ to $\mbz$. If V(D)J recombination impacted patient outcomes without affecting their TCR repertoire, there should be an arrow from $\mbz$ to $y$. 
If mutations occur in TCRs after V(D)J recombination, \Cref{eqn:fitness} fails.
These or other biological mechanisms could violate the model's assumptions, and degrade the accuracy of its estimates.

}

\parhead{Intervention}
We established a causal model of TCR repertoires and disease outcomes,
with the goal of estimating the causal effect of TCRs. Formally, the
causal effect is \add{the result of a hypothetical
\textit{intervention}} on this model. 

We are interested in the effects
of interventions that add TCRs with sequence $a_\star$ to each
patient's repertoire. 
\add{One reason to study such interventions is to learn about the effects of possible therapeutic interventions, such as TCR-T cell therapy, where T cells genetically
engineered to possess a chosen TCR $a_\star$ are transferred into a
patient (\Cref{fig:overview}c)~\citep{Baulu2023-vt}.
}
Mathematically, we add the TCR sequence $a_\star$ to make up a fraction $\epsilon$ of the repertoire, modifying $\mba = \{a_{ij}\}_{j=1}^{m_i}$ to $\{a_{ij}\}_{j=1}^{m_i} \cup 
\{a_\star\}_{j=1}^{\frac{\epsilon}{1-\epsilon} m_i}$ (\Cref{apx:hcm}).
Adapting Pearl's do-notation, we write this intervention as $\rmdo_\epsilon(a_\star)$~\citep{Pearl2009-fh}.
Our goal is to learn what will happen to patients after this intervention.
So the distribution of $y$ after intervention, $\pr(y \s \rmdo_\epsilon(a_\star))$, is our \emph{causal effect} of interest.

\add{Note this is an average treatment effect: it does not specify other biologically relevant factors inside and outside the T cell that modulate a TCR's effects on patient outcomes, such as cell type (e.g. CD4+ or CD8+) or cell state, instead marginalizing out these factors to obtain the average effect (\Cref{apx:id-cell-context,sec:discussion}).
Note also this effect is limited to changes in the proportion rather than absolute number of TCRs. Changes in proportion are easier to learn about from TCR sequencing data, and the two effects in general will converge in the limit of infinite T cells \citep{Weinstein2024-cg}. The number of T cells in an adult is finite though large, roughly 500 billion \citep{Sender2023-sd}.}

\subsection{Identifying TCR Effects} \label{sec:id_details}

The next step towards estimating the causal effect is to
\textit{causally identify} it. In causal identification, we derive a
formula for the interventional distribution
$\pr(y \s \rmdo_\epsilon(a_\star))$ in terms of the
distribution of observable variables $\pr(y, \mba, \mbz)$. From this formula, we then
develop estimation methods to approximate the causal effect from data
(\Cref{sec:estimation}).
Causal identification usually proceeds by assuming that we see an
infinite amount of data from the observational model. \Cref{fig:hcm}
is a \textit{hierarchical} causal model, with an inner plate. So to
study identification, we consider the limit where we have data from an
infinite number of TCRs within each patient ($m \to \infty$ and
$b \to \infty$), as well as from an infinite number of patients
($n \to \infty$).

With this infinite data, we can effectively reconstruct the underlying
\textit{distributions} over TCR sequences within each patient's
repertoire. Each patient's TCR repertoire consists of samples
$a_{i1}, a_{i2}, \ldots$ from their underlying TCR distribution
$q_i^a$; with infinite data we effectively observe $q_i^a$. Likewise
for pre-selection sequences $z_{ij} \sim q_i^z$, we effectively
observe $q_i^z$. Finally, with data $(q_i^a, q_i^z, y_i)$ from
infinite patients, we effectively observe the underlying joint
distribution $\pr(q^a, q^z, y)$. Our goal now is to write the causal
effect in terms of $\pr(q^a, q^z, y)$.

\parhead{The collapsed model} Following \citet{Weinstein2024-cg}, the first step of
identification in a hierarchical causal model is to \textit{collapse}
the model, equating it to a flat causal model without inner plates,
\Cref{fig:chcm}. In the collapsed model, the repertoire distributions
$q^a$ and $q^z$ are causal variables, in place of the TCR sequences
$a$ and $z$. 
\begin{definition}[Collapsed repertoire IV
  model] \label{def:collapsed} 
  \textit{
  The collapsed model has
  the graph in \Cref{fig:chcm} and the mechanisms:
  \begin{equation}
    \label{eqn:chcm}
    \begin{split}
      u_i &\sim \pr(u)\\
      r_i &\sim \pr(r \mid u_i)\\
      q_i^z &\sim \pr(q^z)\\
      q_i^a(x) &= \frac{r_i(x)}{\sum_{x' \in \mathcal{X}} r_i(x') q^{z}_i(x')} q^{z}_i(x) \\
      y_i &\sim \pr(y \mid q^a_i, u_i).
    \end{split}
  \end{equation}
  Given $q^z_i$ and $q^a_i$, TCR sequences in the pre-selection and
  mature repertoires are generated as $z_{ik} \sim q_i^z$ and
  $a_{ij} \sim q_i^a$. }
\end{definition} 

A key consequence of the evolutionary mechanism generating $q^a$ (\Cref{asm:selection}) is that the selective pressures on the repertoire can be reconstructed.
Given $q^z$ and
$q^a$, we can compute $r$ as (\Cref{apx:selection}),
\begin{align}
  \label{eqn:relative-fitness}
  r(x) = \frac{q^{a}(x)/q^{z}(x)}{q^{a}(x_0)/q^{z}(x_0)}.
\end{align}
So we can calculate the relative fitness
of each TCR by examining the likelihood ratio between the
pre-selection and mature distributions. (The sequence $x_0$
is a reference point that can be chosen arbitrarily, since $q^a$ and $q^z$ are normalized to one.) So, the variable $r$
is marked as observed in \Cref{fig:chcm}. To be clear,
\Cref{eqn:relative-fitness} does not describe a causal mechanism, i.e.
$r$ is not caused by $q^a$ and $q^z$. Rather, once we know $q^a$ and
$q^z$, we can reconstruct the value of $r$ which led to $q^a$.
\add{(In principle there might be selective pressures outside the support of $q^z(x)$, where \Cref{eqn:relative-fitness} is ill-defined.
But this selective pressure has no observable consequence, i.e. the value of $r(x)$ outside the support of $q^z(x)$ does not affect $q^a$, $y$ or any other variable in the model, so we can safely assume $r(x)=1$ wherever $q^z(x) = 0$ for the purposes of causal identification (\Cref{apx:id}).}

\parhead{Intervention in the collapsed model} 
We are interested in interventions that add TCRs with a specific sequence. 
In the collapsed model, we can write the intervention $\rmdo_\epsilon(a_\star)$ as a change to the mature repertoire's distribution, modifying $q_a$ to
$q_\star^a = (1-\epsilon)q^a + \epsilon \delta_{a_\star}$ (\Cref{apx:id}).
Post-intervention, patients' repertoires are generated by (1) sampling  $q^a \sim \pr(q^a \mid q^z, r)$, and then (2)
setting
$q_\star^a = (1-\epsilon)q^a + \epsilon \delta_{a_\star}$.
This is a \emph{soft} intervention, meaning there is randomness in the treatment $q^a$ even after intervention \citep{Correa2020-df}.

\parhead{Identification} We can use the collapsed model to compute the effect of the intervention.
Relying on the graphical rules of do-calculus, and its generalization to soft interventions ($\sigma$-calculus), we derive an identification formula. 
This formula tells us how to calculate the effect of an intervention, $\pr(y ; \rmdo_\epsilon(a_\star))$, from the observational distribution $\pr(q^a, q^z, y)$.

\begin{theorem}[TCR effects are identified] \label{thm:softID}
Assume positivity: $\pr(q^a = q^a_\star \mid r) > 0$ for $r \sim \pr(r)$, $q^a_\star \sim \pr(q^a_\star \g r ; \rmdo_\epsilon(a_\star))$. Then, 
\begin{equation} \label{eqn:softID}
  \begin{split} \pr(y \s  \rmdo_\epsilon(a_\star)) = \int \int \pr(y \mid (1 - \epsilon) q^a + \epsilon \delta_{a_\star}, r) \pr(q^a, r) \di q^a \di r,
\end{split}
  \end{equation}
where $\pr(q^a, r)$ is derived from $\pr(q^a, q^z)$ via \Cref{eqn:relative-fitness}.
\end{theorem}
\noindent The proof is in \Cref{apx:id}\add{; it relies most critically on the assumption that $q^a$ depends on $q^z$ and $r$ through a known deterministic relation (\Cref{asm:selection})}. We further discuss the
positivity assumption in \Cref{apx:positivity}, and the
result's relationship to other hierarchical causal models in
\Cref{apx:relate_hcm}.

\Cref{thm:softID} identifies the entire outcome distribution after an
intervention. To summarize the effect of an intervention, we focus on
the \textit{average treatment effect}, defined as the change in the
average outcome after intervention:
\begin{equation} \label{eqn:ate} \textsc{ate}(a_\star, \epsilon)
  \triangleq \mathbb{E}[Y \s \rmdo_\epsilon(a_\star)] - \mathbb{E}[Y \s \rmdo_0(a_\star)].
\end{equation}
The ATE compares the average outcome after adding TCR $a_\star$ at
dosage $\epsilon$ versus the average outcome when no new TCRs are
added, dosage $\epsilon = 0$. \add{Other estimands are also possible. For example, a quantile of the outcome might be more relevant than the average if the goal is to understand the risk of toxicity. If patient covariates are available, we could consider a conditional average treatment effect, stratifying the intervention's effect across the patient population (\Cref{sec:discussion}).}

There are two complementary ways to understand
the identification result in \Cref{thm:softID}. One perspective is that we exploit
natural variation in the pre-selection repertoire as a source of
randomization, since the pre-selection repertoire is created by V(D)J
recombination and unaffected by antigen exposure. Another perspective is that we
exploit the imprint that a patient's history of antigen exposures has
left on their repertoire. We correct for a patient's history of antigen
exposures -- and hence, confounders -- by correcting for the selective forces that shaped their repertoire's development.
Following the second interpretation, we refer to our identification
formula as an ``antigenic history correction.'' We further discuss the
assumptions and limitations of the result in \Cref{sec:discussion}.

 \subsection{Estimating TCR Effects} \label{sec:estimation}

We now show how to estimate the causal effects of TCRs from repertoire
sequences and clinical data. Our method approximates the average
treatment effect (\Cref{eqn:ate}) using the identification formula on
the right hand side of \Cref{eqn:softID}.

For each person $i$ we observe \add{several pieces of data}. We observe $m_i$ TCR
sequences $a_{i1}, \ldots, a_{i m_i}$ from their mature repertoire. 
For bulk TCR sequencing data, which are the largest datasets currently available, each sequence $a_{ij}$ measures only part of the TCR protein, the CDR3$\beta$ region, which is roughly 10-20 amino acids long. Sequencing provides about
$m_i \approx 100{,}000$ sequences per patient.
\add{Sequencing also provides a separate collection
of $\tilde{k}_i$ nonproductive TCR DNA sequences
$\tilde{z}_{i1}, \ldots, \tilde{z}_{i \tilde{k}_i}$, defined as those TCR$\beta$ gene sequences with protein-disabling mutations (specifically, mutations that truncate the protein or shift the reading frame) \citep{Murugan2012-zg}.}
Finally, for each patient, we
observe a clinical outcome $y_i$, a scalar.

\add{
An important limitation of bulk sequencing datasets is that they only provide information from the CDR3 region of the TCR's $\beta$ chain, and in particular provide no information about the $\alpha$ chain of the TCR. As a result, we are limited to estimating the average effect of the CDR3$\beta$, marginalizing over the rest of the protein (\Cref{apx:id-cell-context}).
With large scale single cell data, $a_{ij}$ and $\tilde{z}_{ij}$ could be extended to include the the TCR's CDR3$\alpha$ region, and the treatment effect could also be conditioned on cell type information extracted from gene expression levels (\Cref{sec:discussion}).}

The outline of our approach is as follows. First we construct a
dataset of observations from the collapsed model, \Cref{fig:chcm}. For each patient, we
estimate the mature repertoire distribution $q_i^a$ from their
productive TCR sequences and the pre-selection repertoire distribution
$q_i^z$ from their nonproductive TCR sequences\add{; this latter estimate depends on the assumption that the nonproductive TCR sequences are informative of the pre-selection repertoire, through an assumed biophysical model of V(D)J recombination \citep{Murugan2012-zg}}. We then estimate the
patient's fitness function $r_i$ using \Cref{eqn:relative-fitness}\add{; this estimate depends on the assumption that the repertoire develops via selection (\Cref{asm:selection}).}
From these estimates, we construct a dataset
$\{(\hat{q}_i^a, \hat{r}_i, y_i)\}_{i=1}^n$, with which we estimate
the right side of \Cref{eqn:softID}. We regress $y$ on $\hat q^a$ and
$\hat r$ to estimate the conditional distribution of the outcome
$\pr(y \mid q^a, r)$\add{; this gives us $\pr(y \mid r \s \rmdo(q^a))$ under the assumed causal graph (\Cref{fig:chcm}).} We use $\hat q^a$ and $\hat r$ to estimate
$\pr(q^a, r)$. Finally we combine these estimates via
\Cref{eqn:softID} to obtain the causal effect\add{, according to \Cref{thm:softID}}.

A significant challenge to this estimation is that the data is high
dimensional. The TCR subsequences we analyze are strings of amino
acids, i.e., with an alphabet of 20 characters. The identification
formula in \Cref{eqn:softID} involves distributions over distributions
over high-dimensional discrete objects.
We use representation learning to address this challenge, employing
neural networks to embed the high-dimensional sequences into a
lower-dimensional space. Neural-network representations make it easier
to estimate the distributions of the data; they take advantage of the
assumption that similar TCR sequences have similar effects; and they
allow us to estimate effects of arbitrary sequences.
\add{As a consequence, our estimator is not restricted to \textit{public TCRs}, i.e. those exact amino acid sequences that are found in multiple patients;
instead, it assumes both public and private TCRs come from the same underlying recombination and selection process, and applies the same neural network~\citep{Emerson2017-je}.}

Our method is called \textit{CAIRE: causal adaptive immune receptor
  effect estimator}. In this section, we describe the statistical models underlying CAIRE, how they are fit to data, and the resulting effect estimates. Code and data is available at
\url{https://github.com/EWeinstein/causal-tcrs}.

\parhead{Estimating the repertoire distributions and fitness
  functions} For each patient, we need to estimate the mature
repertoire distribution $q^a_i$ and the pre-selection repertoire
distribution $q^z_i$. To estimate $q^a_i$ we use the empirical
distribution of mature TCR sequences
$\hat{q}^a_i = \frac{1}{m_i} \sum_{j=1}^{m_i} \delta_{a_{ij}}$. 

\add{To estimate $q^z_i$ we apply previous work and fit a generative biophysical model of V(D)J recombination, IGoR, to the nonproductive TCR data \citep{Marcou2018-so}. 
The V(D)J model, $q_\theta^{vdj}(\tilde z)$, describes the full distribution of DNA sequences produced by V(D)J recombination, with parameters $\theta$ that describe the probability of different recombination and mutation events, including gene usage, insertions and deletions. 
These mechanistic parameters can be identified from nonproductive TCR sequences alone, i.e. we can learn $\theta$ given only samples from $q_\theta^{vdj}(\tilde z \mid \mathrm{np}(\tilde z) = 1)$ where $\mathrm{np}(\tilde z)$ is an indicator function for whether $\tilde z$ has a truncation or frameshift mutation. 
So, even when trained only on nonproductive TCRs, the model can extrapolate to describe the distribution over all TCR DNA sequences, on the basis of its biophysical inductive biases~\citep[SI Sec. 10]{Murugan2012-zg}.
This model does not provide direct access to the likelihood $\hat{q}^z_i(x)$ of productive pre-selection TCR amino acid sequences.
However, one can draw samples from $\hat{q}^z_i(x)$ by (1) fitting the parameters $\theta$ of the V(D)J model to the nonproductive TCR DNA sequences, (2) generating samples i.i.d. from $q_{\hat \theta}^{vdj}(\tilde z)$, (3) accepting only sequences without frameshift or truncation mutations, to condition on productivity ($\mathrm{np}(\tilde z) = 0$), and (4) translating the DNA sequences into protein sequences via the standard genetic code.
Since the extrapolation is done on the basis of the mechanics of V(D)J recombination, using data from nonproductive sequences that are not subject to selection, $\hat{q}^z_i(x)$ provides an estimate of the pre-selection repertoire \citep{Murugan2012-zg}.}

We next estimate the fitness function $r_i$ for each patient.
\Cref{eqn:relative-fitness} writes the fitness function as a density
ratio between two high-dimensional distributions, $q_i^a$ and $q_i^z$.
Directly estimating a high-dimensional density ratio is challenging,
so we use the methods of \cite{Sugiyama2010-kk,Mohamed2016-yc} to
reduce the problem to estimating a low-dimensional parameter in a
classifier model.

In particular, we train a classifier to distinguish between sequences
sampled from $\hat{q}^a_i$ and those sampled from $\hat{q}^z_i$.
\add{The probability of any given sequence $x$ coming from $q^a$ under the classifier provides an estimate of the likelihood ratio between the distributions, $\hat{q}^a_i(x)/\hat{q}^z_i$, which we can turn into estimate of $r_i$ \citep{Sugiyama2010-kk,Mohamed2016-yc}.}
Consider a candidate sequence $x_{ij}$ and let $s_{ij}$ denote whether
it comes from the mature repertoire or the pre-selection repertoire.
The classifier model is
\begin{align}
  \label{eqn:classifier}
  s_{ij}
  \sim
  \mathrm{Bernoulli}(\sigma(\rho_i^\top h_r(x_{ij}; \phi) + \beta_i)),
\end{align}
where $\sigma(x) = 1/(1 + \exp(-x))$. Here, $h_r(x ; \phi)$ is a
convolutional neural network parameterized by $\phi$ that extracts low-dimensional features of a sequence $x$
(\Cref{apx:semisynth_arch}), while $\rho_i \in \mathbb{R}^{d_r}$ and
$\beta_i \in \mathbb{R}$ are latent per-patient coefficients.

\add{There is a one-to-one mapping between $\rho_i$ and $r_i$, so we use the learned value of
$\rho_i$ as a low-dimensional summary or \textit{representation} of $r_i$
(\Cref{apx:selection_est_details}).} In particular, if the
classifier is trained accurately on an equal number of samples from
$q^a_i$ and from $q^z_i$ then \Cref{eqn:relative-fitness} implies that
$r_i (x) = \exp( \rho_i^\top [h_r(x; \phi) - h_r(x_0;\phi)])$.
Intuitively, $\rho_i$ describes the amount of selective pressure on
each sequence feature extracted by $h_r(x; \phi)$.
\add{We discard the latent offset $\beta_i$, since we only need a relative likelihood ratio (\Cref{eqn:relative-fitness}) not the absolute likelihood ratio.}  

\parhead{Estimating the intervention} With estimates of $q^a$ and
$r$ in hand, we now estimate the right side of \Cref{eqn:softID}.

First we fit a model of the outcome, $\pr(y \g q^a, r)$:
\begin{align}
  \label{eqn:outcome}
  y_i \sim \mathrm{Normal}\left (\gamma_a^\top \mathbb{E}_{\hat{q}^a_i}[h_a(A ;\theta)] + \gamma_r^\top \rho_i + \gamma_0, \tau_y\right),
\end{align}
where $h_a(x ;\theta)$ is another convolutional neural network,
parameterized by $\theta$. This model uses $\rho_i$ as
\add{a low-dimensional representation of $r_i$, since $r$ can be reconstructed from $\rho$}. It uses $h_a(x ;\theta)$ to extract a
low-dimensional representation of sequences, and then averages
individual sequence representations together to produce an overall
representation of the entire repertoire,
$\mathbb{E}_{\hat{q}^a_i}[h_a(A ;\theta)]$~\citep{Zaheer2017-vs}. \add{The
repertoire's representation is linear in $q^a_i$, without interaction terms between $q^a(x)_i$ and $q^a(x')_i$ for $x' \neq x$.
We assume linearity for simplicity; the lack of interaction terms is motivated by the biological assumption that individual TCR proteins do not interact physically with one another through their variable region \citep{Abbas2018-zj}. }

Finally we use \Cref{thm:softID} to estimate the average treatment
effect (\Cref{eqn:ate}). We approximate the integral over
$\pr(q^a, r)$ from the empirical distribution of $\hat{q}^a$ and of $\hat r$, as represented by $\rho$.

\begin{align}
&\textsc{ate}(a_\star, \epsilon) = \int  \mathbb{E}[Y \mid (1 - \epsilon) q^a + \epsilon \delta_{a_\star}, r] \pr(q^a, r) \di q^a \di r - \int \mathbb{E}[Y \mid q^a, r] \pr(q^a, r) \di q^a \di r\\
    &\approx  \left(\frac{1}{n} \sum_{i=1}^n \gamma_a^\top (\epsilon  h_a(a_\star \s \theta) + (1- \epsilon)\mathbb{E}_{\hat{q}^a_i}[h_a(A ;\theta)]) + \gamma_r^\top \rho_i + \gamma_y\right) 
     \quad - \left( \frac{1}{n} \sum_{i=1}^n \gamma_a^\top (\mathbb{E}_{\hat{q}^a_i}[h_a(A ;\theta)]) + \gamma_r^\top \rho_i + \gamma_y \right) \\
    &= \epsilon \gamma_a^\top \left( h_a(a_\star \s \theta) - \frac{1}{n} \sum_{i=1}^n \mathbb{E}_{\hat{q}_i^a}[h_a(A \s \theta)]\right). \label{eqn:ate_est}
\end{align}

\parhead{Summary} The method takes the following two steps. (1) Fit
the parameters $\rho_i$, $\beta_i$ and $\phi$ in the classifier model
(\Cref{eqn:classifier}) to estimate relative fitness $r_i$. (2) Fit
the parameters $\gamma_a$, $\gamma_r$, $\gamma_0$, $\theta$ and
$\tau_y$ in the outcome model (\Cref{eqn:outcome}) to predict $y_i$.
The final result is a trained neural network
$\hat \gamma_a^\top h_a(x ; \hat \theta\,)$ that predicts the effect
of adding an arbitrary TCR sequence to patient repertoires
(\Cref{eqn:ate_est}). 
\add{By cross-fitting on different subsets of patients, or training an ensemble of neural networks, we can obtain an ensemble of estimates $\hat \gamma_{a}^{(1)\top} h_a(x ; \hat \theta^{(1)}\,), \ldots, \hat \gamma_a^{(K)\top} h_a(x ; \hat \theta^{(K)}\,)$, providing a rough uncertainty measure \citep{Lakshminarayanan2017-jy,Wilson2020-wb}.} 

In practice, we fit the parameters of the classifier model and the
outcome model simultaneously. We use stochastic optimization to scale
to large datasets, drawing batches of patients and batches of
repertoire sequences for each patient. 
\add{To enable scalable inference of the $n$ per-patient latent variables, $\rho_1, \ldots, \rho_n$ and $\beta_1, \ldots, \beta_n$, we apply an \textit{amortized inference} technique based on \textit{autoencoders}~\citep{Rezende2014-ad,Kingma2014-sp,Amos2023-mw}.
In particular, rather than optimize all $n$ latent variables separately, we train a neural network (an \textit{encoder}) to output an estimate of $\rho_i$ and $\beta_i$ given a batch of repertoire sequences from patient $i$.
This allows the inference method to scale to large $n$. } 
We also
employ a propensity score correction, using a propensity model of the
treatment $q^a$ given the confounder $r$~\citep{Kaddour2021-uu}.
Further description is in \Cref{apx:reg_train} and \Cref{apx:models}.

 \section{Semisynthetic Data Study} \label{sec:semisynthetic}

We first evaluate CAIRE on semisynthetic data, where we have access to
\add{true causal effects}. We demonstrate that CAIRE is capable of
inferring which TCRs in a patient repertoire affect the outcome, even
in the presence of confounding, and even when those TCRs are rare
within repertoires. We also show that the existing state of the art
for repertoire classification~\citep{Widrich2020-hj} does not provide
similarly accurate causal inferences.
\add{CAIRE's success hinges on its distinguishing assumption that the pre-selection repertoire is an instrumental variable, along with its practical ability to exploit this assumption to correct for unobserved confounders.}

\subsection{Semisynthetic data}

To construct semisynthetic data, we extend previous methods designed
for evaluating non-causal TCR repertoire classification
methods~\citep{Widrich2020-hj,Pavlovic2021-sn,Slabodkin2023-ba}.
Following these approaches, we ``inject motifs'' (short subsequences)
into a small fraction of TCRs in a subset of patients. The presence of
each motif will be associated with the outcome variable. However,
unlike in previous semi-synthetic studies, only one motif actually causes the outcome. The idea that short sequence motifs are
responsible for the biological activity of TCRs has been motivated by
structural studies of the binding interaction between immune receptors
and antigens~\citep{Ostmeyer2019-ak,Akbar2021-mk}.

In more detail, we start with values of the pre-selection repertoire
distribution $q^z_i$ learned from a real TCR
dataset~\citep{Emerson2017-je,Pavlovic2021-sn}. We then inject a
``causal'' motif, by modifying $q^z_i$ such that a small fraction of
TCRs will have a specific subsequence. Meanwhile, we set the fitness
$r_i(x)$ of sequences $x$ containing a ``confounded'' motif such that
the fitness depends on the unobserved confounder $u_i$. Next, we
generate $q^a_i$ according to the selection mechanism
\Cref{eqn:fitness}. 
\add{We draw sequences from $q^z_i$ and $q^a_i$, with different numbers of sequences for different patients, following the sequencing depths of the real dataset.}
We generate the outcome $y_i$ based on the
confounder $u_i$, and based on the fraction of sequences in the mature
repertoire with the ``causal'' motif. In short, we design the
simulation such that the presence of two motifs is associated with the
outcome variable, \add{but only sequences with one motif actually have non-zero treatment effect $\textsc{ate}(a_\star, \epsilon)$}. Full
details are in \Cref{apx:semisynthetic}.

\subsection{Evaluation}

To evaluate CAIRE, we determine how well the estimated effect
$\ateest(a_\star, \epsilon)$ can discriminate sequences with the
causal motif from those without it. Intuitively, we ask how well the
method can recover effective therapeutic sequences---sequences that
actually cause good patient outcomes---from among patient repertoires.
We quantify classification performance with the area under the
precision recall curve (PR-AUC) on a test set of sequences from
held-out repertoires. Full details are in \Cref{apx:semisynth_eval}.

\subsection{Results}

\begin{table*}
  \centering
  \caption{\textbf{Method performance on semisynthetic data.} Values are mean and standard error of PR AUC across 15 independent datasets. } \label{tab:semisynth_perform}
  \begin{tabular}{ccc|cc}
  \hline
  CAIRE & Attention CAIRE & No propensity CAIRE & Uncorrected & DeepRC$^\star$\\
  \hline
  0.86 $\pm$ 0.04 & 0.82 $\pm$ 0.04 & 0.92 $\pm$ 0.03 & 0.56 $\pm$ 0.02 & 0.55 $\pm$ 0.03\\
  \hline
  \end{tabular}
\end{table*}

We compare CAIRE to several alternatives. We apply each candidate
method to 15 different independently generated semisynthetic datasets.
In each dataset, the motifs are injected into 1\% of pre-selection sequences, \add{since the fraction of a repertoire that recognizes a given antigen is often low \citep{Slabodkin2023-ba}.} For each
method on each dataset, we use Bayesian optimization to set
hyperparameters across 10 different
configurations~\citep{Balandat2020-pv}. In CAIRE, we optimize the
dimension of the latent fitness representation $\rho_i$ and sequence representation $h_a(x ;\theta)$,
as well as the size of the convolutional kernel in the convolutional
neural network layer of $h_a$.
Full details on the experiments are in \Cref{sec:cmv}.

We first demonstrate that CAIRE successfully adjusts for confounding. We
compare the method to an alternative that does not account for
confounding, i.e. it fixes $\rho_i = 0$ for all $i$. Without the confounding
correction, the method is much worse at distinguishing causal TCR sequences from non-causal sequences, \add{compared to when the confounding correction is included}
(\Cref{tab:semisynth_perform}, CAIRE vs uncorrected).

Next we compare CAIRE to DeepRC~\citep{Widrich2020-hj}, a
state-of-the-art repertoire classification method. 
DeepRC uses a transformer-based neural network to classify patient repertoires, e.g., learning to predict from repertoire sequences whether or not a patient has had a certain disease.
The key differences between DeepRC and CAIRE are that (a) DeepRC does not
correct for confounding, and (b) DeepRC computes its repertoire
embedding using an attention mechanism, up-weighting certain areas of
sequence space according to a learned attention score. We slightly
modified data weightings in DeepRC so the method can
provide valid effect estimates under the assumption of no confounding
(\Cref{apx:deeprc}). We call this modified version DeepRC$^\star$.
CAIRE substantially outperforms DeepRC$^\star$ at causal effect estimation
in the presence of confounding (\Cref{tab:semisynth_perform}).

We also compare CAIRE to a variant, ``Attention CAIRE,'' that corrects for confounding but uses DeepRC's attention mechanism. \add{Using the attention mechanism does not provide
better effect estimates than standard CAIRE, though it still outperforms alternatives that do not correct for confounding (\Cref{tab:semisynth_perform}).}

Finally, we consider the role of the propensity model. We found a
small but statistically insignificant advantage from removing the
propensity model (\Cref{tab:semisynth_perform}, ``no propensity CAIRE'';
permutation t-test p-value of 0.30). We
include the propensity model in our method because it offers
theoretical robustness guarantees, ensuring a safety net for real
data, \add{though it does come with additional computational cost.}

We next considered a simulation scenario where there is no confounding, i.e. $u$ does not affect $y$.
We evaluated CAIRE with and without its confounding correction, i.e. setting $\rho_i = 0$.
We find identical performance between the two methods: both achieve an
average PR-AUC of one across 10 independent simulations.
So, CAIRE performs well even in the absence of confounding.

A key concern for any TCR repertoire analysis method is that it should be sensitive to small repertoire changes,
since in practice the subpopulation of TCRs involved in an immune
response may be small~\citep{Widrich2020-hj,Slabodkin2023-ba}. We
lowered the motif injection rate from $1\%$ down to $0.5\%$ and
$0.1\%$ (\Cref{apx:semisynth_exp}, \Cref{fig:semisynth_scaling}). While the performance of both
CAIRE and its uncorrected version deteriorate slightly, CAIRE
continues to outperform at an injection rate of $0.5\%$. At $0.1\%$
both methods \add{provide poor effect estimates on average across independent simulations}, with no distinguishable difference between
them (permutation t-test p-value of 0.12).

 \section{Application: COVID-19 Severity} \label{sec:application}

\add{We now use CAIRE to
estimate the effects of TCRs on COVID-19 severity. Then we conduct an exploratory analysis of the estimates.} First, we compare
CAIRE's estimates to other sources of biological information,
including laboratory measurements of TCR binding. Second, we use
CAIRE's estimates to develop an overall picture of TCR-mediated
immune responses in COVID-19. Finally, we use CAIRE's estimates to
nominate therapeutic candidates, including both TCRs and vaccine
antigens.

The observational dataset contains $n=507$ COVID-19 patients from
2020~\citep{Nolan2020-fx,Snyder2020-qg}. The data contains
high-throughput sequencing results for the TCR $\beta$ CDR3 region \add{collected from blood samples},
with an average of 201,000 productive TCRs per patient (minimum 5,231,
maximum 733,792). The study also provides clinical data about the
patients' disease outcomes. Based on this data, we constructed an
overall measurement of disease severity on a three point scale, with
$y=+1$ corresponding to mild disease (no hospitalization), $y=0$
corresponding to moderate disease (hospitalization), and $y=-1$
corresponding to severe disease (ICU admission and/or death).
Additional details on data preprocessing and analyses
are in \Cref{apx:application_data,apx:application_eval,apx:application}.

\add{To evaluate predictive performance and quantify uncertainty, we refit CAIRE several times to different splits of the data. 
We use 8-fold stratified cross validation, repeated three times with different (random) partitions of the data.
We use the heldout data in each split to measure model fit.
As a point estimate of the average treatment effect, we report the average of $\textsc{ate}(a_\star, \epsilon)$ across the ensemble of 24 model fits.
To quantify our uncertainty in the effect of each $a_\star$, we fit a Gaussian to the ensemble of estimates, and report the probability $\tilde{p}$ that the sign of the point estimate is incorrect (\Cref{apx:effect_uncertainty}).
Note this ensembling approach provides only a rough heuristic approximation of a Bayesian posterior over the treatment effect, and does not ensure rigorous coverage guarantees~\citep{Lakshminarayanan2017-jy,Wilson2020-wb}.
}

We present our findings in detail in the following sections. The main results are as follows:
\begin{itemize}
	\item TCRs have diverse effects: among sequences found in patient repertoires, a substantial number have modest positive effects on patient outcomes, but some have strong positive effects and some have negative effects.
	\item Causal effects in patients are distinct from laboratory binding measurements: the causal effect of a TCR sequence is related to its binding properties but does not perfectly correlate, \add{suggesting the two are biologically distinct}.
	\item \add{Patients have heterogenous repertoires: all patients contain a mix of TCRs with not just positive but also negative causal effects, and some patients with severe COVID-19 having an especially large number of TCRs with negative effects.}
	\item Patient repertoires contain potential therapeutic candidates: we uncover TCR sequences that are (1) observed in patients, (2) possess \textit{in vitro} experimental evidence for binding SARS-CoV-2 epitopes, and (3) possess observational clinical evidence for efficacy, based on CAIRE.
\end{itemize}

\subsection{Effect estimates} \label{sec:effect_est_overview}

\begin{figure*}
\centering
\begin{subfigure}[c]{0.45\textwidth}
	\includegraphics[width=\textwidth]{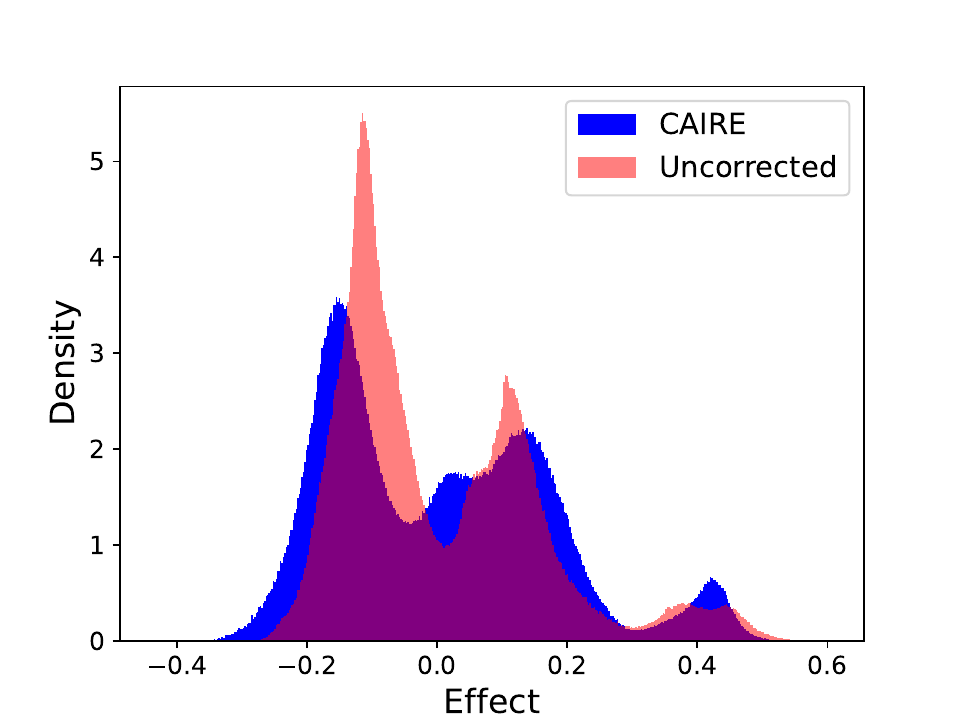}
\caption{}
\end{subfigure}
\begin{subfigure}[c]{0.45\textwidth}
	\includegraphics[width=\textwidth]{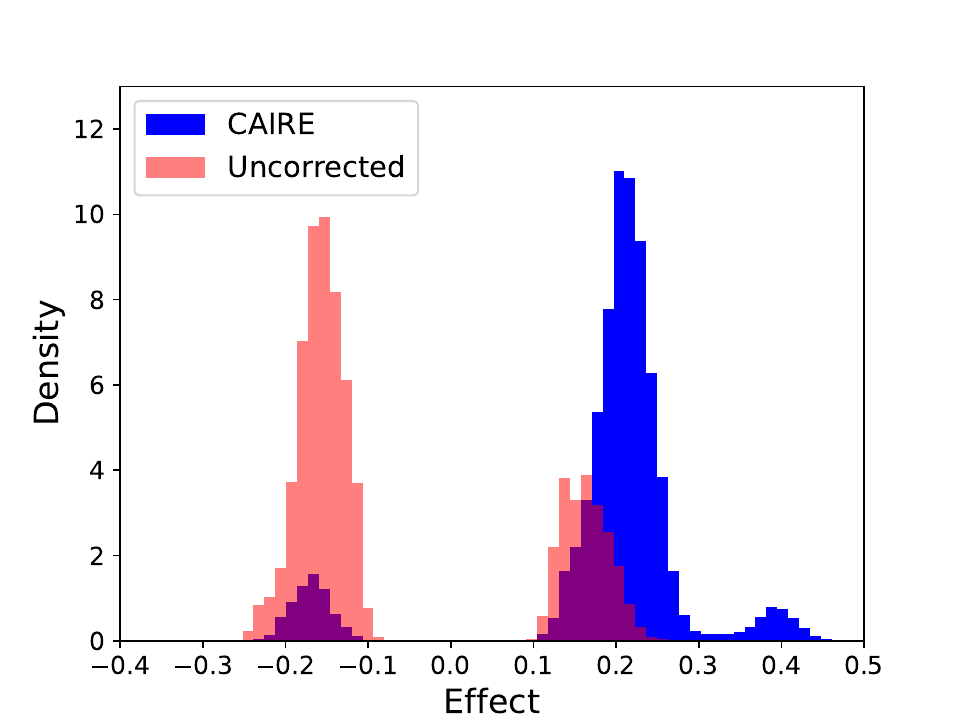}
\caption{}
\end{subfigure}
\caption{\textbf{Distribution of TCR effects in held-out patient repertoires.} (a) Effect distribution across all repertoire sequences. (b) Effect distribution across sequences with \add{more confident} effects ($\tilde p < 0.05$)} \label{fig:heldout_effect_distr}
\end{figure*}

We use CAIRE to estimate the effect $\textsc{ate}(a_\star;\epsilon)$ of interventions that use each of 11 million TCR sequences.
These sequences come from the repertoires of 71 COVID patients, \add{which were held-out from training} (\Cref{apx:heldout_cov}).
We set the dosage parameter $\epsilon$ to $0.1$, based on the dosage of existing TCR-based therapies (\Cref{apx:dosage}). Note that the effect at other $\epsilon$ values can also be obtained by linearly rescaling (\Cref{eqn:ate_est}).
We find 170,000 sequences (1.6\% of the total) with \add{more confident} effects on outcomes \add{($\tilde p < 0.05$ under our ensemble-based posterior approximation; note we do not apply a multiple hypothesis correction, since we are jointly modeling the effects of all sequences rather than testing each separately \citep{Gelman2012-pd})}. 
Among these, there is a small population of sequences with a strong positive effect ($\approx$0.4 on the three-point severity scale), a larger population with a weak positive effect ($\approx$0.2), and a small population with a weak negative effect ($\approx$-0.15) (\Cref{fig:heldout_effect_distr}).

We compare CAIRE's estimates of causal effects to those of a method
without the antigenic history correction (``Uncorrected''). We observe
a substantial shift in the distribution of effects: the uncorrected method predicts
that, among sequences with a \add{more confident} effect, 70\% have a negative
effect on patient outcomes, while CAIRE predicts only 10\%
(\Cref{fig:heldout_effect_distr}b). This suggests that many sequences
are statistically associated with severe disease not because they
cause severe disease, but rather due to confounding. For example, this
may be due to greater viral exposure in patients with severe COVID-19,
leading to greater positive selection of virus-responsive TCRs,
regardless of whether those TCRs are actually causing worse clinical outcomes. Overall, CAIRE's results are
consistent with the established idea that T cell-mediated immunity is
broadly important in combatting SARS-CoV-2, while the uncorrected
method's estimates are not~\citep{Moss2022-tr}.

CAIRE finds substantial confounding: 
its confounder term (i.e., $\gamma_r^\top \rho_i$) explains roughly twice as much of the variance in the outcome as the treatment term (\Cref{tbl:predictive_perf}, \Cref{fig:predictive_distribution}).
However, in this dataset, the confounding captured by CAIRE does not appear to reflect easy-to-measure demographic variables.
We do not find evidence that the confounder representation $\rho_i$ is associated with age, gender or ethnicity (\Cref{sec:confound_rep}). 

The causal effect estimates provided by CAIRE are largely robust to different model architectures. 
``Attention CAIRE'' achieves a similar fit to the data (\Cref{tbl:predictive_perf}, \Cref{fig:predictive_distribution}), and its effect estimates show a strong correlation with those of CAIRE (Pearson $R$ of 0.95, p value for non-correlation below floating point precision; \Cref{fig:effect_correlation_attention}).
We also compared CAIRE to a much simpler method, that does not use any non-linearities, though it still includes convolutions (\Cref{apx:nonneural}).
This ``Non-Neural CAIRE'' achieves a somewhat worse fit to the data (\Cref{tbl:predictive_perf}, \Cref{fig:predictive_distribution}), and its effect estimates are moderately correlated with CAIRE's (Pearson $R$ of 0.69, p value below precision; \Cref{fig:effect_correlation_nonneural}).

\subsection{Comparison to laboratory experiments} \label{sec:binding_compare}

We compare the causal effects of TCRs, estimated by CAIRE, to
laboratory measurements of TCR binding with SARS-CoV-2 antigens. While
\textit{in vitro} binding does not necessarily imply that a TCR
will have a strong effect on patient outcomes, it is reasonable to
expect the causal effect of a TCR to be related to its ability to bind
viral proteins. We use data collected from a high-throughput
experiment (MIRA, multiplex identification of T-cell receptor antigen
specificity) which identifies patient TCRs that bind different
antigens from across the SARS-CoV-2
genome~\citep{Nolan2020-fx,Snyder2020-qg}. The assay provides the
sequences of TCRs that bind each antigen (32,770 sequences total).

We evaluate how well CAIRE's effect estimates predict \textit{in
  vitro} TCR binding. This evaluation is not trivial because the assay
does not provide sequences found to \textit{not} bind viral antigens.
However, we do have repertoire sequencing data collected from the
patients whose T cells were used in the binding assay. That is, we
have access to sequence data without binding labels, but drawn from
the same underlying distribution of sequences as used in the binding
experiments. In \Cref{apx:application_eval} we show how this unlabeled
data can be used to evaluate a method's ability to discriminate
binders from non-binders. In particular, we show that the ROC AUC of a
classifier evaluated on discriminating known binders from unlabelled
sequences (MIRA hits from repertoire sequences) is a conservative
estimate of the classifier's AUC when  discriminating
binders from non-binders.

We find that that CAIRE's effect estimates can discriminate class I MHC binders with an ROC AUC of 0.563$\pm$ 0.003 (standard error; \Cref{apx:binding_comparison}).
Restricting the analysis just to sequences with \add{more confident} effect estimates ($\tilde{p} < 0.05$), the AUC is 0.604$\pm$0.027.
By contrast, the uncorrected method achieves a slightly lower AUC on all sequences (0.556$\pm$0.003) and a still lower AUC on sequences with high-confidence effects (0.534$\pm$0.020).
These results suggest that (a) CAIRE's effect estimates are somewhat consistent with, though not identical to, \textit{in vitro} experimental binding data, and (b) the antigenic history correction can lead to effect estimates that correspond more closely to experimental data. 
We discuss some implications of these results below in \Cref{sec:therap-implications}

\subsection{Balanced immune responses} \label{sec:application_balance}

\begin{figure*}
\centering
\begin{subfigure}[c]{0.45\textwidth}
\includegraphics[width=\textwidth]{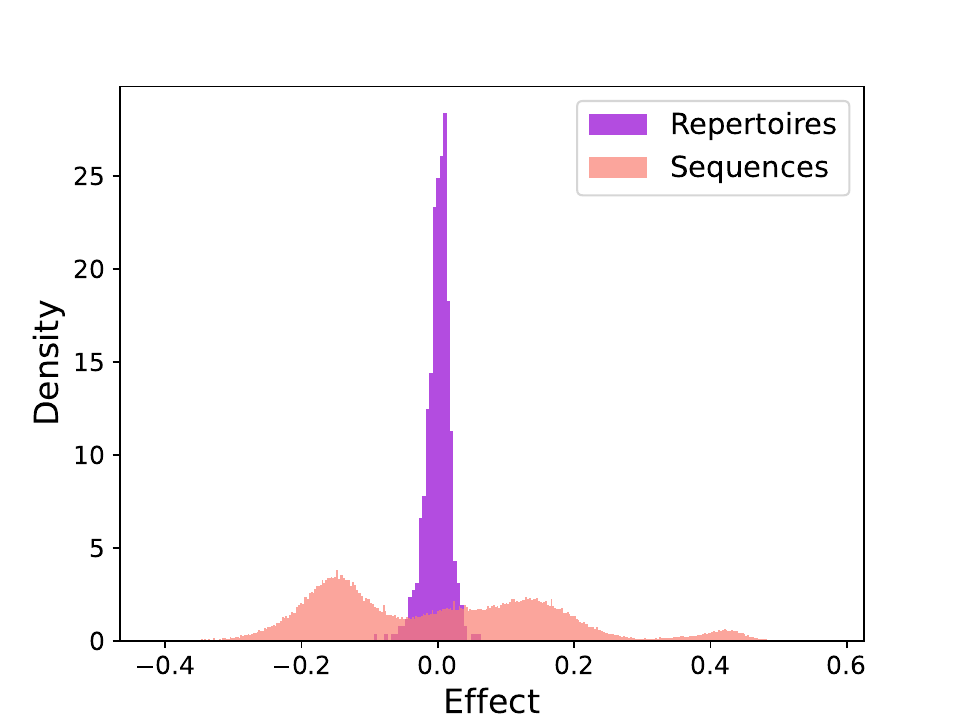}
\caption{} \label{fig:effect_distr_within_v_across}
\end{subfigure}
\begin{subfigure}[c]{0.45\textwidth}
\includegraphics[width=\textwidth]{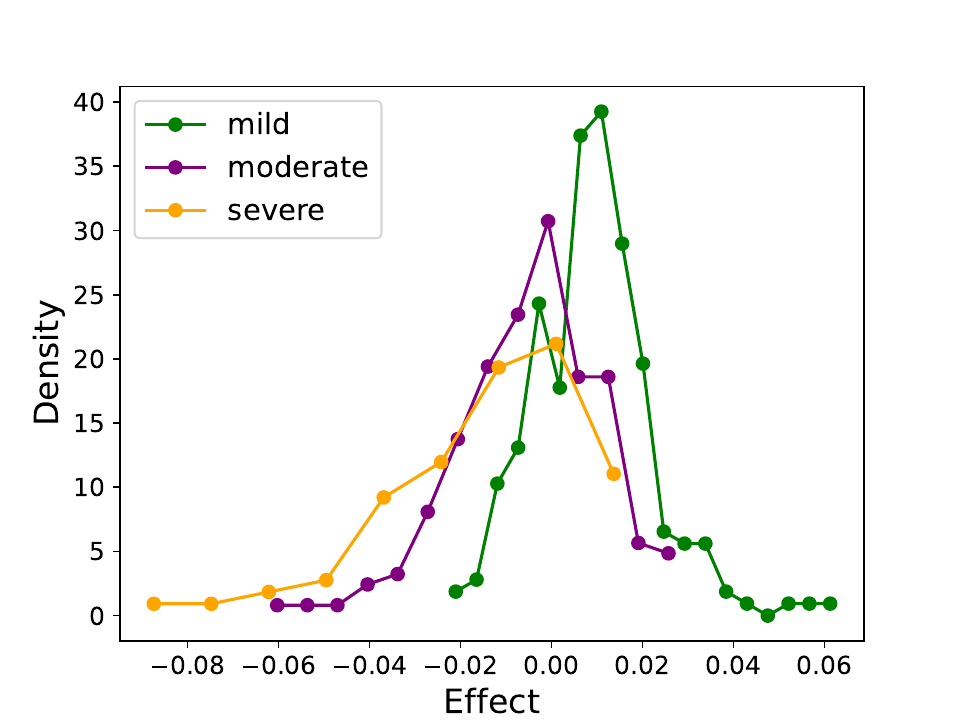}
\caption{} \label{fig:effect_distr_patient_average_outcome}
\end{subfigure}
\caption{\textbf{Effect heterogeneity within and between patients} (a) The distribution of average effects across repertoires (purple) and the average distribution of effects within repertoires (red). More precisely, each bar of the purple histogram covering interval $\mathcal{I}$ is an estimate of $\mathbb{P}_{Q^a \sim p(q^a)}[\mathbb{E}_{A \sim Q^a}[\textsc{ate}(A;0.1)] \in \mathcal{I}]$. Each bar of the red histogram is an estimate of $\mathbb{E}_{Q^a \sim p(q^a)}[\mathbb{P}_{A \sim Q^a}[ \textsc{ate}(A;0.1) \in \mathcal{I}]]$.
(b) Distribution of average effects across repertoires, from patients with different outcomes. Each point at interval $\mathcal{I}$ is an estimate of $\mathbb{P}_{Q^a \sim p(q^a \mid y)}[\mathbb{E}_{A \sim Q^a}[\textsc{ate}(A;0.1)] \in \mathcal{I}]$ for an outcome $y \in \{-1, 0, +1\}$.}
\end{figure*}

We next try to better understand how patients' repertoires, taken as a
whole, drive disease. Our results, presented below, support the
following conceptual model for TCR-mediated immune responses in
COVID-19. First, in general, patient repertoires provide TCRs capable
of fighting the virus. However, this success comes at a cost:
repertoires also carry a burden of TCRs with negative effects, likely
because these TCRs produce overly extreme immune responses. Some
burden of negative effect TCRs is present even in individuals with
mild disease, but in a subset of patients it is large, and helps drive
severe disease. It is the balance of the two factors -- TCRs that
drive safe immune responses and TCRs that drive damaging immune
responses -- that determines how the repertoire as a whole affects
patient outcomes.

We applied CAIRE to estimate the effects,
$\textsc{ate}(a_\star, \epsilon=0.1)$, of repertoire sequences drawn
from each of the $n=507$ COVID-19 patients in the training data. We
compared variability of effects \textit{between} patients to
variability \textit{within} patients
(\Cref{fig:effect_distr_within_v_across}). We find that there is much
more variation in the effects of individual TCRs within patients than
there is variation in the average effect of repertoires between
patients
($\sqrt{\mathbb{V}_{\pr(q^a)}[\mathbb{E}_{Q^a}[\textsc{ate}(A, 0.1)]]}
= 0.018$ versus
$\sqrt{\mathbb{E}_{\pr(q^a)}[\mathbb{V}_{Q^a}[\textsc{ate}(A, 0.1)]]}
= 0.12$, where $\mathbb{V}$ denotes variance). Indeed, most patients
contain a mix of TCRs with large positive and large negative effects,
but these balance each other out to create a more modest average
effect for the repertoire overall
(\Cref{fig:effect_distr_by_outcome_per_patient,fig:effect_distr_patient_average_outcome}).
(Note the effect of a patient's entire repertoire,
$\mathbb{E}[Y \s \rmdo(q^a_\star = q_i^a)] - \mathbb{E}[Y] =
\mathbb{E}_{q^a_i}[\textsc{ate}(A, 1)] = 10 \times
\mathbb{E}_{q_i^a}[\textsc{ate}(A, 0.1)]$ has an estimated average
absolute value of
$\mathbb{E}_{\pr(q^a)}|\mathbb{E}_{Q^a}[\textsc{ate}(A, 1)]| \approx
0.13$.)

Some TCRs are predicted to have a negative effect on disease outcomes, suggesting they do more harm than good in responding to viral infection.
Although the antigenic history correction in CAIRE reduces the size of this population, it does not eliminate it altogether.
Biologically, one possible explanation for negative causal effects is that these TCRs drive an overactive immune response. 
This is consistent with the clinical observation that some patients with severe COVID-19 suffer from severe hyperinflammation and immune reactions such as cytokine storm, leading to conditions such as acute respiratory distress syndrome or multisystem inflammatory syndrome~\citep{Cheng2020-ep,Lucas2020-dc,Kalfaoglu2021-kh,Mobasheri2022-dn,Moss2022-tr,Fajgenbaum2020-vz}.
COVID-19 patients also have increased risk of autoimmune disease post-infection~\citep{Sharma2023-wr}.

We find that all patients have some sequences with negative effects, but a subpopulation of patients with severe or moderate COVID-19 possess a greatly expanded set of TCRs with strong negative effects (\Cref{fig:effect_distr_patient_average_outcome,fig:negative_tail_weight_by_outcome}).
Moreover, TCRs with negative causal effects are more likely to bind SARS-CoV-2 antigens, as compared to TCRs with positive effects (\Cref{tbl:binding_predict} column 1, \Cref{fig:binder_unknown_distr_comparison}). 
This supports the hypothesis that TCRs can have a negative effect because they induce too wide and overactive an immune response~\citep{Stone2015-ol,Shakiba2022-ww}.
In summary, the role of a patient's TCR
repertoire depends on the balance between two populations
of TCRs: those that drive safe and effective immune responses versus
those that drive damaging immune responses.

\subsection{Implications for therapeutics and vaccines} \label{sec:therap-implications}

We next consider the implications of CAIRE's causal estimates for therapeutics.
We first consider therapies based on adding TCRs to a patient's repertoire, such as cell therapies or TCR bispecifics~\citep{Papayanni2021-wv,Papadopoulou2023-fk,Verhagen2021-lc}.
Applying CAIRE to binders identified in MIRA assays, we find 17 candidate therapeutic sequences with strong positive effects ($\textsc{ate}(a_\star; 0.1) > 0.3$, $\tilde{p} < 0.05$; \Cref{tab:top_tcrs}). 
These candidates are (1) observed in human patients (2) possess \textit{in vitro} experimental support for their binding activity (from MIRA), and (3) possess observational clinical support for therapeutic efficacy (from CAIRE).
Our results caution against relying exclusively on \textit{in vitro} assays of on-target binding when developing therapeutics: 60\% of binders found in MIRA assays are estimated to have \textit{negative} clinical effects, while another 37\% are predicted to have only weak clinical effects ($0 < \textsc{ate}(a_\star; 0.1) < 0.3$).

\begin{figure*}
\centering
	\includegraphics[width=\textwidth]{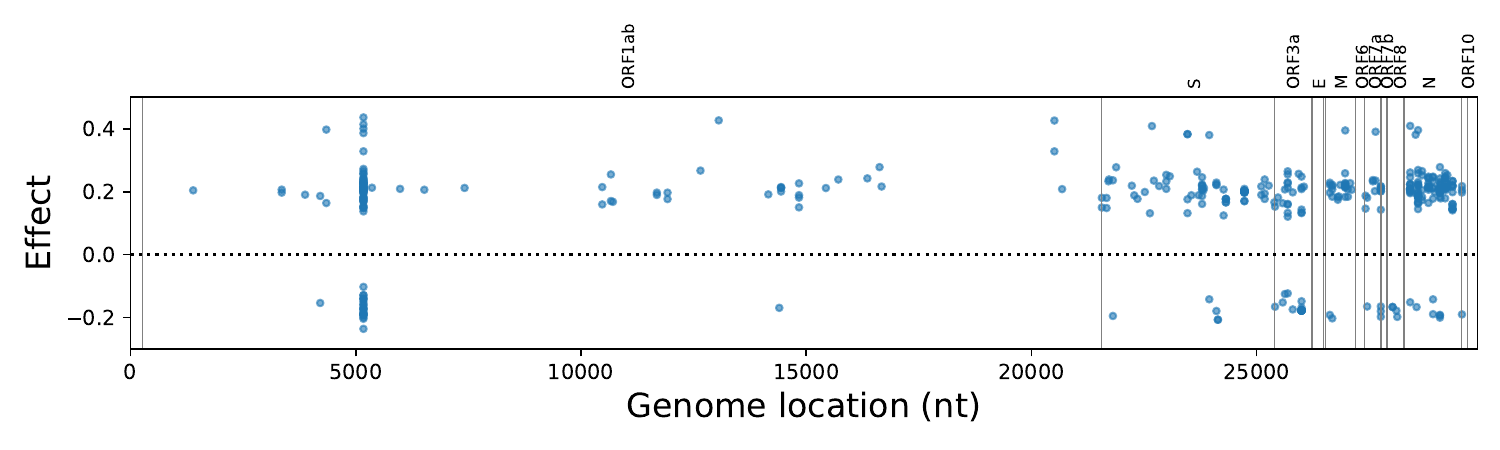}
\caption{\textbf{Distribution of TCR effects across the SARS-CoV-2 genome.} x-axis: antigen location in the SARS-CoV-2 genome, indexed by nucleotide. y-axis: estimated causal effect of TCRs that bind that antigen. Each dot represents an individual TCR with a \add{more confident} effect ($\tilde p < 0.05$) that was found to bind an epitope encoded at the given location in the SARS-Cov-2 genome.} \label{fig:binder_significant_effect_loc}
\end{figure*}

We next consider the implications of CAIRE's estimates for vaccine development.
A central question in vaccine design is how to choose an antigen that induces a strong, beneficial immune response. 
To address this question, we examine the effects of TCRs that bind each SARS-CoV-2 antigen studied in the MIRA binding assay.
We find many antigens that bind a diversity of TCRs, including TCRs with \add{more confident} positive and negative effects (\Cref{fig:binder_significant_effect_loc}).
However, a subset of antigens were enriched for TCRs with \add{more confident} positive effects (\Cref{tab:top_antigens}; binomial test, Benjamini-Hochberg adjusted p-value below 0.05).
These may be promising vaccine candidates.
Among these antigens are \add{one epitope from the virus' spike protein, as well as three from its nucleocapsid protein}, including NP$_{105-113}$. This last epitope was identified in experimental studies as a candidate for protection against severe COVID-19 \citep{Peng2020-au,Peng2022-jh}.
Overall, these results support the hypothesis that, besides the spike protein, the nucleocapsid protein may be a good candidate for improved T-cell vaccines~\citep{Moss2022-tr}.
 
 \section{Discussion} \label{sec:discussion}

We propose a new method for causal inference from
observational TCR repertoire data. We use V(D)J recombination as a
source of randomness. We show that a standard biophysical model of T
cell development implies causal identification. We then develop an
estimation strategy using nonproductive TCR data. The method is
general purpose in that it only requires TCR sequencing data and
clinical outcomes. It is also scalable, running on data with $\sim$100
million sequences.

\parhead{Assumptions and limitations} As with all observational causal
inference techniques, our method comes with assumptions and
limitations. Our identification approach assumes that the process of
V(D)J recombination is unaffected by confounders. But we cannot
entirely exclude the possibility of hidden confounding: the biological
mechanisms that lead to variation among individuals' pre-selection
repertoires are poorly understood, and there may be some genetic or
environmental factors that affect both the process of V(D)J
recombination and patient outcomes~\citep{Slabodkin2021-wr}. Another
limitation is that CAIRE may control for instruments, as some factors
that affect TCR fitness are likely unrelated to patient outcomes;
correcting for these factors risks worse effect
estimates~\citep{Bhattacharya2007-xe}. \add{Moreover, our model of T cell
repertoire development (\Cref{asm:selection}) accounts only for selection, and ignores
processes such as genetic drift that may also play an important role
in development~\citep{Horns2019-jh,Koraichi2023-en}. 
Our approach assumes nonproductive TCRs found in T cells provide reliable estimates of the pre-selection repertoire, an assumption that may degrade if, for example, the process of V(D)J recombination changes quickly over time.}
\add{The compartment or tissue the T cells are sequenced from, when they are sequenced, and the depth of sequencing can introduce additional variability and bias into CAIRE's estimates.}
Finally, the performance of CAIRE depends on the performance of
the upstream computational tools it relies on, in particular the
methods used to reconstruct TCR clonotype sequences from raw
sequencing data, and to reconstruct the pre-selection repertoire from
nonproductive sequences.

\parhead{Future work} To advance precision medicine applications, it
will be important to develop conditional treatment effect estimates
rather than only study average treatment effects. A key challenge in
particular is to estimate the effect of TCRs conditional on a
patient's HLA type, as HLAs directly affect TCR-antigen binding.
\add{With paired TCR and single cell RNA sequencing, CAIRE could be extended to condition on the TCR $\alpha$ chain, the T cell type, expressed co-receptors, and transcriptional state.}
Finally, although we have focused on T cell receptors, CAIRE may also
be applied to B cell receptors. An additional challenge here is that
somatic hypermutation plays an important role in B cell receptors,
complicating the assumption that repertoire development proceeds by
selection alone. But effect estimates for B cell receptors have the
potential to inform antibody-based medicines, a major class of
therapeutics.

\add{Methodologically, CAIRE may be improved along several axes. 
One direction is to develop more rigorous and well calibrated uncertainty estimates, using e.g. improved Bayesian deep learning methods or conformal inference techniques \citep{Miani2025-ia,Lei2021-do}. Robustness could be improved, to protect against e.g. distribution shifts between different tissues \citep{Kuhn2024-qh,Miron2021-ga}.  Sensitivity analysis could evaluate the implication of errors in the model's causal assumptions \citep{Miani2025-ia,Lei2021-do,Jesson2022-za}. 
Another direction is to improve the hierarchical causal model's estimates by correcting for finite sample bias introduced by its nested Monte Carlo likelihood \citep{Rainforth2018-qp}. 
Large unlabeled TCR datasets might be leveraged to improve representation learning, even when clinical outcome data is missing \citep{Nagano2025-dd}.}

\parhead{Outlook} Overall, our work advances causal inference methods for scientific and technological challenges.
Rather than analyze the effects of a binary treatment, we aim to understand high-dimensional and structured treatments, namely proteins.
We exploit the underlying biology and chemistry  as a natural experiment, relying on the stochastic enzymatic reactions of V(D)J recombination for randomization.
Conventional IV assumptions and estimators are unsuitable for this setting.
Instead, we use fundamental biological theory to derive a new IV identification formula, and scalable estimators.
Going forward, we anticipate that progress on scientific and societal challenges such as drug design will demand such new causal inference methodology, integrating statistics, machine learning and natural science.

\parhead{Data availability}
This article relies on publicly available data accessed from \url{https://clients.adaptivebiotech.com/pub/emerson-2017-natgen} \citep{Emerson2017-je} and \url{https://clients.adaptivebiotech.com/pub/covid-2020} \citep{Snyder2020-qg,Nolan2020-fx}. Preprocessing details can be found at \url{https://github.com/EWeinstein/causal-tcrs}.

\parhead{Code availability} All code is available at \url{https://github.com/EWeinstein/causal-tcrs}.

\bibliographystyle{abbrvnat}
\bibliography{references}

\appendix

\crefalias{section}{appendix}
\crefalias{subsection}{subappendix}
\crefalias{subsubsection}{subsubappendix}

\renewcommand{\thefigure}{S\arabic{figure}}
\renewcommand{\thetheorem}{S\arabic{theorem}}
\renewcommand{\thedefinition}{S\arabic{definition}}
\renewcommand{\theproposition}{S\arabic{proposition}}
\renewcommand{\theassumption}{S\arabic{assumption}}
\renewcommand{\theHtheorem}{appendix.\arabic{theorem}}

\setcounter{figure}{0}
\setcounter{theorem}{0}

\pagebreak
\section{Details on Identification} \label{apx:identification}

\subsection{Hierarchical causal model} \label{apx:hcm}

In this section we define the causal model and interventions formally.
\begin{definition}[Repertoire IV model] \label{def:repertoire_iv}
\textit{
	The repertoire IV model has the graph shown in \Cref{fig:hcm} and hierarchical causal graphical model (HCGM) equations are,
\begin{equation} \label{eqn:hcm}
\begin{split}
	u_i &\sim \pr(u)\\
	r_i &\sim \pr(r \mid u_i)\\
	q_i^z &\sim \pr(q^z) \quad \quad \quad \quad \quad \quad \quad \quad \,\, z_{ik} \sim q_i^z\\
	q_i^a &\sim \pr(q^a \mid \{z_{ik}\}_{k=1}^b, r_i) \quad \quad \quad a_{ij} \sim q_i^a\\
	y_i &\sim \pr(y \mid \{a_{ij}\}_{j=1}^m, u_i),
\end{split}
\end{equation}
for $i \in \{1, \ldots, n\}$, $j \in \{1, \ldots, m\}$ and $k \in \{1, \ldots, b\}$.}
\end{definition}
\noindent We use the notation $\{\cdot\}_{j=1}^m$ to indicate that the causal mechanisms $\pr(y \mid \{a_{ij}\}_{j=1}^m, u_i)$ and $\pr(q^a \mid \{z_{ik}\}_{k=1}^b, u_i)$ cannot depend on the ordering of the TCRs, just on the unordered set of sequences.\footnote{Technically, $\{a_{ij}\}_{j=1}^m$ is a multiset, as some sequences can be identical and in this case the number of repeated sequences is still relevant.}
This reflects the idea that there is no natural ordering of a patient's T cells.

We consider interventions where a single TCR sequence $a_\star$ is added to a patient's repertoire. Formally, this corresponds to a stochastic or \textit{soft} intervention, since there is still some randomness in the treatment variable after the intervention, coming from the rest of the repertoire~\citepapx[Chap. 4][]{Pearl2009-fh,Dawid2002-yq,Munoz2012-vy,Correa2020-df}. 
\begin{definition}[Intervention by adding a TCR]
\textit{
	After intervention, each patient's repertoire distribution is generated according to the conditional distribution $\sigma_{a_\star,\epsilon}(q^a_\star \g r)$, defined as
\begin{equation} \label{eqn:soft_intervene}
\begin{split}
	q^a &\sim \pr(q^a \g r)\\
	q^a_\star &= (1-\epsilon) q^a + \epsilon \delta_{a_\star},
\end{split}
\end{equation}
where $\pr(q^a \g r)$ is the causal mechanism in the un-intervened model (\Cref{def:repertoire_iv}). In other words, $\sigma_{a_\star,\epsilon}$ is the pushforward of $\pr(q^a \g r)$ through the function $(1-\epsilon) q^a + \epsilon \delta_{a_\star}$. After intervention, each patient's repertoire is generated as
$$ q_i^a \sim \sigma_{a_\star,\epsilon}(q^a_\star \g r_i) \quad \quad \quad a_{ij} \sim q_i^a$$
for $i \in \{1, \ldots, n\}$, $j \in \{1, \ldots, m\}$. }
\end{definition}

\subsection{Fitness and natural selection} \label{apx:selection}

In this section, we review the mathematical model of evolution under natural selection used in \Cref{eqn:fitness}, and the derivation of \Cref{eqn:relative-fitness}~\citepapx{Neher2011-ls,Bertram2019-ok}.

\textit{Fitness} is a measure of how much an organism with a particular genotype reproduces. In our setting, the organism is a T cell and the genotype is their TCR sequence. Fitness $g_i(x)$ is the size of the population of mature cells with TCR gene sequence $x$ divided by the size of the population of pre-selection cells with sequence $x$.
If $s_i$ is the total number of cells in the pre-selection repertoire, then $s_i q^z_i(x)$ is the total number of cells with sequence $x$, and $g_i(x) s_i q^z_i(x)$ is the total number of cells with sequence $x$ in the post-selection repertoire. So, the distribution of sequences $q^a_i(x)$ in the post-selection repertoire must be proportional to $g_i(x) q^z_i(x)$ up to a normalizing constant, i.e. $q^a_i(x) \propto g_i(x) q^z_i(x)$.

To describe fitness in a way that is agnostic to the normalizing constant, we employ not \textit{absolute} fitness $g_i(x)$ but instead \textit{relative} fitness $r_i(x)$, which is the ratio of the fitness of sequence $x$ to the fitness of an arbitrarily chosen reference sequence, $x_0$, i.e. $r_i(x) = g_i(x)/g_i(x_0)$. 
From $q_i^a(x) \propto g_i(x) q_i^z(x)$ and $\sum_x q_i^a(x) = 1$ we have,
$$ q^a_i(x) = \frac{g_i(x) s_i }{\sum_{x' \in \mathcal{X}} g_i(x') s_i q^z(x')} q_i^z(x),$$
and \Cref{eqn:fitness} follows.

Since $r_i(x_0) = 1$, we also have from plugging $x_0$ into \Cref{eqn:fitness} that $\sum_{x' \in \mathcal{X}} r_i(x') q^{z}_i(x') = q_i^z(x_0)/q_i^a(x_0)$. So,
\begin{equation*}
q^a(x) = \frac{r_i(x)}{q_i^z(x_0)/q_i^a(x_0)} q^z(x).
\end{equation*}
Solving for $r_i$ yields \Cref{eqn:relative-fitness}.

\subsection{Proof of identification (\Cref{thm:softID})} \label{apx:id}

In this section we prove \Cref{thm:softID}. The first step is to rewrite the hierarchical causal model in the format of Def. 3 in \citeapx{Weinstein2024-cg}. 
Next we introduce a regularity assumption which ensures the infinite repertoire limit exists, and derive the collapsed model.
Then we apply do-calculus to identify the effect of hard interventions on $q^a$. 
Finally we apply $\sigma$-calculus to identify the effect of the intervention of interest.

\parhead{Rewritten HCM} \label{sec:id_rewrite}

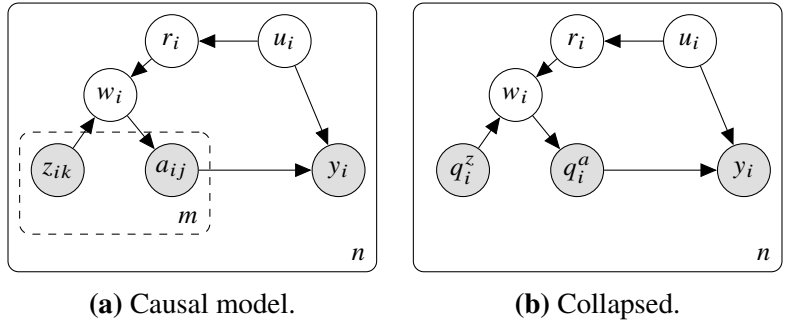
\begin{figure}
\centering
\begin{subfigure}[c]{0.32\textwidth}
\centering
\begin{tikzpicture}

\node[obs] (y) {$y_{i}$};
  \node[obs, left=1.5cm of y] (a) {$a_{ij}$};
  \node[latent, left=.1cm of a, yshift=1cm] (w) {$w_i$};
  \node[obs, left=.8cm of a] (z) {$z_{ik}$};
  \node[latent, above=of a, xshift=1.5cm]  (u) {$u_i$};
  \node[latent, above=of a]  (r) {$r_i$};

\edge {a,u} {y} ;
  \edge{u}{r};
  \edge {z,r} {w} ;
  \edge {w} {a};

\plate[dashed] {za} {(a)(z)} {$m$} ;
  \plate {zayu} {(a)(z)(u)(y)(w)(za)} {$n$} ;

\end{tikzpicture}
\caption{Causal model.} \label{fig:ehcm}
\end{subfigure}
\begin{subfigure}[c]{0.32\textwidth}
\centering
\begin{tikzpicture}

\node[obs] (y) {$y_{i}$};
  \node[obs, left=1.5cm of y] (a) {$q^{a}_i$};
  \node[latent, left=.1cm of a, yshift=1cm] (w) {$w_i$};
  \node[obs, left=.8cm of a] (z) {$q^{z}_i$};
  \node[latent, above=of a, xshift=1.5cm]  (u) {$u_i$};
  \node[latent, above=of a]  (r) {$r_i$};

\edge {a,u} {y} ; 
  \edge{u}{r};
  \edge {r,z} {w} ;
  \edge{w}{a};

\plate {zayu} {(a)(z)(u)(y)(w)(za)} {$n$} ;

\end{tikzpicture}
\caption{Collapsed.} \label{fig:cehcm}
\end{subfigure}
\caption{\textbf{The repertoire IV causal model, rewritten.}}
\end{figure}

We rewrite the causal model (\Cref{def:repertoire_iv}) so that it explicitly meets Def. 3 of \citeapx{Weinstein2024-cg}, and we can apply the identification techniques developed in that paper.
We assume $b=m$, i.e. the number of pre-selection repertoire sequences matches the number of mature repertoire sequences.
Since the identification result takes the number of sequences to infinity, this is a minor restriction; it can be relaxed and identification will still hold.
\begin{definition}[Rewritten repertoire IV model] \label{def:rewritten}
\textit{
The repertoire IV model (\Cref{def:repertoire_iv}) is equivalent to an HCGM with graph in \Cref{fig:ehcm} and equations,
\begin{equation}
\begin{split}
	u_i &\sim \pr(u)\\
	q_i^z &\sim \pr(q^z) \quad \quad \quad \quad \quad \quad z_{ik} \sim q_i^z\\
	w_i &\sim \pr(q^a \mid \{z_{ik}\}_{k=1}^m, u_i)\\
	q_i^a & \sim \delta_{w_i}\quad \quad \quad \quad \quad \quad \quad a_{ij} \sim q_i^a\\
	y_i &\sim \pr(y \mid \{a_{ij}\}_{j=1}^m, u_i).
\end{split}
\end{equation}
for $i \in \{1, \ldots, n\}$, $j \in \{1, \ldots, m\}$.}
\end{definition}
\noindent Here $w_i$ is an unobserved \textit{interferer} variable. Intuitively, there is interference between the pre-selection and mature repertoires because multiple mature T cells can descend from the same thymocyte, resulting in multiple copies of the same TCR in the mature repertoire.
This model is equivalent to the original repertoire IV model, as we have simply rewritten the mechanism for $a$ in terms of an intermediate variable $w$. We can see the model meets Def. 3 in \citeapx{Weinstein2024-cg}.

 \parhead{Infinite subunit limit}
 We study identification in the limit of infinite subunits, i.e. sequences.
We assume the limit exists.
 \begin{assumption}[Mechanism convergence] \label{asm:infinite}
 	Assume the mechanisms of the model in \Cref{def:rewritten} converge in Kullback-Leibler divergence, such that the model meets the conditions of Theorem 1 of \citeapx{Weinstein2024-cg}.
 \end{assumption}
\noindent The exact conditions are defined and discussed in detail in \citeapx{Weinstein2024-cg}.
 
 Theorem 1 of \citeapx{Weinstein2024-cg} implies that the hierarchical causal model converges in the limit $m \to\infty$ to a collapsed model with graph in \Cref{fig:cehcm} and equations,
 \begin{equation}
\begin{split}
	u_i &\sim \pr(u)\\
	q_i^z &\sim \pr(q^z)\\
	r_i &\sim \pr(r \g u_i)\\
	w_i &\sim \pr(q^a \mid q_i^z, r_i)\\
	q_i^a & \sim \delta_{w_i}\\
	y_i &\sim \pr(y \mid q_i^a, u_i).
\end{split}
\end{equation}
Marginalizing out $w_i$ and imposing \Cref{asm:selection} on the mechanism $\pr(q^a \mid q^z, r)$, we recover the collapsed repertoire IV model (\Cref{def:collapsed}).

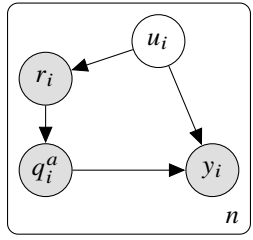
\begin{figure}
\centering
\begin{tikzpicture}

\node[obs] (y) {$y_{i}$};
  \node[obs, left=1.5cm of y] (a) {$q^{a}_i$};
  \node[obs, above=.5cm of a] (r) {$r_{i}$};
  \node[latent, above=of a, xshift=1.5cm]  (u) {$u_i$};

\edge {a,u} {y} ; 
  \edge {u} {r} ;
  \edge {r} {a} ;

\plate {zayu} {(a)(u)(y)} {$n$} ;

\end{tikzpicture}
\caption{\textbf{Marginalized collapsed repertoire IV model.}} \label{fig:mhcm}
\end{figure}

\parhead{Hard interventions on repertoires}
We first identify the effect of an intervention that draws the repertoire sequences for each patient from a fixed distribution, $\pr(y \s \rmdo(a \sim q_\star^a))$.
This is equivalent to a hard intervention on the repertoire distribution $\pr(y \s \rmdo(q^a = q_\star^a))$.
\begin{theorem}[Repertoire effects are identified] \label{thm:repertoire_id}
	Assume the repertoire IV model (\Cref{def:repertoire_iv}) satisfies \Cref{asm:infinite} and \Cref{asm:selection}. Further assume positivity, i.e. $\pr(q^a_\star \mid r) > 0$ a.s. for $r \sim \pr(r)$. Then, the causal effect of hard interventions on the repertoire is identified from $\pr(y, q^a, q^z)$ as,
	\begin{equation} \label{eqn:repertoire_id}
		\pr(y \s \rmdo(q^a = q^a_\star)) = \int \pr(y \mid q^a = q^a_\star, r) \pr(r) \di r,
	\end{equation}
	where $\pr(r)$ is derived from $\pr(q^a, q^z)$ via \Cref{eqn:relative-fitness}.
\end{theorem}
\begin{proof}
	By Thm. 1 of \citeapx{Weinstein2024-cg}, it suffices to identify the effect in the collapsed model, \Cref{def:collapsed}, since the effect in the hierarchical causal model will be equivalent to the effect in the collapsed model in the limit $m \to \infty$.
	Now, marginalize out the variable $q_i^z$ from the collapsed model, so that the mechanism generating $q^a$ from its parent $r$ is stochastic (\Cref{fig:mhcm}). 
\Cref{eqn:repertoire_id} follows by an application of do-calculus to \Cref{fig:mhcm}, under the positivity conditions of \citeapx{Shpitser2006-jg} (Assumption 3 in \citeapx{Weinstein2024-cg}).
\end{proof}

\add{We have assumed that $r(x) = 1$ outside the support of $q^z$, i.e. where $q^z(x) =0$. To see why this assumption does not impact causal identification, consider an arbitrary $r(x)$, and split it into two variables, an identified section $r^{obs}(x) = \{r(x) \text{ if } q^z(x) > 0, 1 \text{ otherwise}\}$ that can be reconstructed with \Cref{eqn:relative-fitness}, and an unidentified section $r^{un}(x) = \{r(x) \text{ if } q^z(x) = 0, 1 \text{ otherwise}\}$, such that $r(x) = r^{obs}(x)r^{un}(x)$ for all $x$.
Although $r^{un}$ is unobserved it has no causal descendants, since it does not impact $q^a$ or any other variable, so $r^{un}$ can be marginalized out of the causal graph. Then the same identification result, \Cref{eqn:repertoire_id}, holds with $r^{obs}$ in place of $r$.}

\parhead{Identification proof}
We now prove the main identification result. Interventions that add a TCR correspond to soft interventions on the repertoire distribution.
So, we apply $\sigma$-calculus, an extension of do-calculus to soft interventions~\citepapx[Thm. 1]{Correa2020-df}.
\begin{proof}
Starting from the collapsed and marginalized model in \Cref{fig:mhcm}, we decompose the effect as,
\begin{equation*}
\begin{split}
	\pr(y &\s  \rmdo_\epsilon(a_\star)) = \pr(y ; \rmdo(q^a_\star \sim \sigma_{a_\star, \epsilon}))\\ = &\int \int \pr(y \mid r, q^a \s \rmdo(q^a_\star \sim \sigma_{a_\star, \epsilon})) \pr(q^a \mid r \s \rmdo(q^a_\star \sim \sigma_{a_\star, \epsilon})) \pr(r \s \rmdo(q^a_\star \sim \sigma_{a_\star, \epsilon})) \di r \di q^a.
\end{split}
\end{equation*}
From rule 2 of $\sigma$-calculus, we have $\pr(y \mid r, q^a \s \rmdo(q^a_\star \sim \sigma_{a_\star, \epsilon})) = \pr(y \mid r, q^a)$.
From rule 3 of $\sigma$-calculus, we have $\pr(r \s \rmdo(q^a_\star \sim \sigma_{a_\star, \epsilon})) = \pr(r)$.
So,
\begin{equation*}
	\pr(y \s  \rmdo(q^a_\star \sim \sigma_{a_\star, \epsilon})) = \int \pr(y \g q^a,  r) \pr(q^a \mid r \s \rmdo(q^a_\star \sim \sigma_{a_\star, \epsilon})) \pr(r) \di q^a \di r.
\end{equation*} 
The conditional distribution $\pr(q^a \mid r \s \rmdo(q^a_\star \sim \sigma_{a_\star, \epsilon}))$ is specified by the intervention, \Cref{eqn:soft_intervene}.
Plugging in \Cref{eqn:soft_intervene} yields the identifying equation,
\begin{equation*}
\begin{split}
	\pr(y \s  \rmdo(q^a_\star \sim \sigma_{a_\star, \epsilon})) = \int \pr(y \mid (1 - \epsilon) q^a + \epsilon \delta_{a_\star}, r) \pr(q^a, r) \di q^a \di r. 
\end{split}
\end{equation*}

To be able to compute the integral on the right hand side, it must be possible to identify $\pr(y \mid (1 - \epsilon) q^a + \epsilon \delta_{a_\star}, r)$ for all values of $q^a, r$ with $\pr(q^a, r) > 0$ (see e.g. Appx. H in \citeapx{Weinstein2024-cg} for further discussion).
This is guaranteed by the stated positivity condition, which ensures that the intervention has non-zero probability under the observational distribution: $\pr((1 - \epsilon) q^a + \epsilon \delta_{a_\star}, r) > 0$ for $q^a, r \sim \pr(q^a, r)$ a.s..
\end{proof}

\subsection{Positivity} \label{apx:positivity}

For causal identification to hold, the intervention must satisfy positivity, i.e. it must naturally occur under the observational distribution. In particular, the intervention must have positive probability regardless of the value of the adjustment variable, $r$. 
Here we show that it is indeed possible to satisfy the positivity condition.
We focus on the positivity condition used for hard repertoire interventions (\Cref{thm:repertoire_id}), as it is easier to interpret, but the same reasoning extends to interventions that add a TCR (\Cref{thm:softID}).
For simplicity, here we take the set of possible sequences $\mathcal{X}$ to be finite.
\begin{proposition}
Assume the set of possible sequences $\mathcal{X} \subset \cup_{\ell=1}^\infty \Bc^\ell$ is finite.
	Assume $\pr(q^z)$ has full support over distributions on sequences, i.e. $\pr(q^z) > 0$ for all $q^z \in \mathcal{P}(\mathcal{X})$. Assume that with probability 1 under $\pr(r)$, $r(x) > 0$ for all $x \in \mathcal{X}$. Then, $\pr(q^{a}=q_\star^a \mid r) > 0$ for $r \sim \pr(r)$ a.s..
\end{proposition}
\begin{proof}
Since $r$ must be strictly positive, the selection process can always be run in reverse with $1/r$ in place of $r$, giving
\begin{equation*}
	\tilde{q}^{z}(x) = \f(r^{-1}, q^a_\star) = \frac{r^{-1}(x) }{\sum_{x'} r^{-1}(x') q_\star^a(x')} q_\star^a(x).
\end{equation*}
This distribution $\tilde{q}^{z}(x)$ generates the intervention, $q_\star^a = \f(r, \tilde{q}^{z})$.
Since $\tilde{q}^{z}$ is a valid distribution, it has positive probability, $\pr(\tilde{q}^{z}) > 0$. The conclusion follows.
\end{proof}

It is useful to contrast this result with an alternative model, where instead of the relative fitness mechanism (\Cref{eqn:fitness}) we assume a mechanism of the form $q^a_i(x) = r_i(x) q^z_i(x)$, without the normalizer in \Cref{eqn:fitness}.
In this case, $r_i(x) = q^a_i(x)/q^z_i(x)$, the likelihood ratio.
Now, however, $\tilde{q}^z(x) = r^{-1}_i (x) q_\star^a(x)$ is not necessarily a valid distribution, as it may not be normalized.
As a result, positivity would be difficult to guarantee. 

\add{\subsection{Accounting for Cellular Context} \label{apx:id-cell-context}

In practice, the effects of a T cell receptor are modulated by the cellular context, e.g. whether the T cell is CD4+ or CD8+. 
Our causal effect implicitly marginalizes over this context. To see this explicitly, consider an expanded causal model which includes a variable $b_{ij}$ representing the cellular context, which may include for example any unmeasured sections of the TCR (e.g. the $\alpha$ chain), the cell type, and the cell state (e.g. gene expression levels) (\Cref{fig:hcm-b}).
By collapsing and marginalizing this HCM, and applying the arguments in \Cref{apx:id}, we obtain \Cref{fig:chcm-b}. 
From do calculus, the causal effect is still identified by the same formula (\Cref{eqn:softID}).
The causal effect takes an average over the distribution of contexts $q^{b \mid a}$ given the antigenic history encoded in $r$, $\pr(y \mid r \s \rmdo(q^a)) = \int \pr(y \mid q^{b \mid a}, q^a, r) \pr(q^{b \mid a} \mid  r) \di q^{b \mid a}$, where $q^{b \mid a}(b \mid a)$ describes the conditional distribution of cellular context given sequence within a person.
Post-intervention in this HCM, a persons' T cells follow the distribution,
\begin{align}
	a_{ij} &\sim (1 - \epsilon) q_i^a + \epsilon \delta_{a_\star}\\
	b_{ij} &\sim q_i^{b \mid a}(b \mid a_{ij}).
\end{align}
That is, the conditional distribution of cellular context given TCR sequence remains unchanged after the intervention.

\begin{figure*}
\centering
\begin{subfigure}[c]{0.36\textwidth}
\centering
\begin{tikzpicture}

\node[obs] (y) {$y_{i}$};
  \node[obs, left=1.5cm of y] (a) {$a_{ij}$};
  \node[latent, below=1cm of a] (b) {$b_{ij}$};
  \node[obs, left=1cm of a] (z) {$z_{ik}$};
  \node[latent, above=.5cm of a, xshift=-0.7cm] (r) {$r_{i}$};
  \node[latent, above=of a, xshift=1.5cm]  (u) {$u_i$};

\edge {a,u,b} {y} ; \edge {u} {r} ;
  \edge {r} {a,b} ;
  \edge {z} {a} ;

\plate[dashed] {ap} {(a)(b)} {$m$} ;
  \plate[dashed] {zp} {(z)} {$b$} ;
  \plate {zayu} {(ap)(zp)(u)(y)} {$n$} ;

\end{tikzpicture}
\caption{} \label{fig:hcm-b}
\end{subfigure}
\begin{subfigure}[c]{0.36\textwidth}
\centering
\begin{tikzpicture}

\node[obs] (y) {$y_{i}$};
  \node[obs, left=1cm of y] (a) {$q^{a}_i$};
  \node[latent, below=1cm of a] (b) {$q^{b \mid a}_i$};
  \node[obs, left=1cm of a] (r) {$r_{i}$};
  \node[latent, above=of a]  (u) {$u_i$};

\edge {a,u,b} {y} ; 
  \edge {u} {r} ;
  \edge {r} {a} ;
  \edge {r} {b} ;

\plate {zayu} {(a)(b)(u)(y)(r)} {$n$} ;

\end{tikzpicture}
\caption{} \label{fig:chcm-b}
\end{subfigure}
\caption{\add{\textbf{Causal graphs with cellular context $b$.} (a) Hierarchical causal model. (b) Collapsed and marginalized causal model. $u_i$: unobserved confounding. $z_{ik}$: pre-selection repertoire sequences. $a_{ij}$: mature repertoire sequences. $b_{ij}$: additional cellular context, e.g. gene expression. $y_i$: patient outcomes. $q^z_i$: pre-selection repertoire distribution. $q^a_i$: repertoire distribution. $r_i$: relative fitness. }}
\end{figure*}}

\subsection{Relationship to HCM IV and General Control Functions} \label{apx:relate_hcm}

We briefly discuss the relationship of the repertoire IV model presented here to the hierarchical IV model described in \citeapx{Weinstein2024-cg}.
Though both methods use hierarchical causal models with a subunit-level instrument and treatment variables, they rely on different types of datasets and different identification assumptions.

The HCM IV model in \citeapx{Weinstein2024-cg} assumes that it is possible to observe both the treatment and the instrument within each subunit.
This implies that it is possible to estimate the joint distribution over the treatment and instrument within each unit, $q_i(a, z)$.
The identification result depends crucially on knowledge of this joint, as it involves a backdoor correction on the conditional $q_i(a \mid z)$

In the repertoire IV model presented here, however, we do not have direct access to the joint distribution over pre-selection and mature sequences, since we cannot observe those pre-selection cells which die off during selection.
Instead, we use domain knowledge to constrain the relationship between the instrument and the treatment (\Cref{asm:selection}).
As a result, in the identification formula of \Cref{thm:repertoire_id}, the relative fitness $r_i$ occupies essentially the same role as the conditional $q_i(a \mid z)$ in the HCM IV identification formula.

Both $r_i$ in repertoire IV and $q_i(a \mid z)$ in HCM IV can be seen as \textit{general control functions}, since they can be used together with the instrument $q^z_i$ to reconstruct the treatment $q^a_i$~\citepapx[Theorem 1]{Puli2020-ud}.

 \section{Further Description of Estimation} \label{apx:estimate_details}

\subsection{Selection estimate} \label{apx:selection_est_details}
Here we describe our estimation strategy for relative fitness in more depth.

To construct the statistical model in \Cref{eqn:classifier},
we assign each of the mature repertoire sequences $a_{i1}, \ldots, a_{im_i}$ the label $s_{ij} = 1$, and then draw an equal number of samples $z_{i1}, \ldots, z_{im_i}$ from $\hat{q}_i^{z}$ and assign them the label $s_{ij} = 0$. 
Let $\{(x_{ij}, s_{ij})\}_{j=1}^{2m_i} = \{(a_{ij}, 1)\}_{j=1}^{m_i} \cup \{(z_{ij}, 0)\}_{j=1}^{m_i}$ denote the pooled dataset. 
We now construct the model in \Cref{eqn:classifier}, restated here:
\begin{equation*}
	s_{ij} \sim \mathrm{Bernoulli}(\sigma(\rho_i^\top h_r(x_{ij}; \phi) + \beta_i)),
\end{equation*}
for  $j \in \{1, \ldots, 2m_i\}$ and $i \in \{1, \ldots, n\}$.
Note for fixed $\phi$, this is a hierarchical logistic regression model, with $\rho_i$ and $\beta_i$ the per-patient regression coefficients and offset respectively.
If $h_r$ is unrestricted, then this model can describe any conditional distribution $\pr(s \mid x)$, since $s$ is binary.

We now relate the representation $\rho_i$ to the relative fitness $r_i$. 
Let $q_i(s, x)$ be the true joint distribution over sequences and labels for patient $i$, and assume the regression model accurately matches $q_i(s \mid x)$. Applying Bayes' rule,
\begin{equation}
	\frac{q_i^{a}(x)}{q_i^{z}(x)} = \frac{q_i(x \g s=1)}{q_i(x \mid s=0)} = \frac{q_i(s=1\g x)}{q_i(s=0 \g x)}\frac{q_i(s=0)}{q_i(s=1)} = \frac{\sigma(\rho_i^\top h_r(x; \phi) + \beta_i)}{1 - \sigma(\rho_i^\top h_r(x; \phi) + \beta_i)} = \exp(\rho_i^\top h_r(x; \phi) + \beta_i),
\end{equation}
where we have used the fact that $q_i(s = 0) = q_i(s = 1) = 0.5$ by construction.
Plugging this into \Cref{eqn:relative-fitness}, we find that, 
\begin{equation}
	r_i (x) = \exp( \rho_i^\top [h_r(x; \phi) - h_r(x_0;\phi)]).
\end{equation}
Note $\beta_i$ does not appear.
This equation implies a one-to-one relationship between $\rho_i$ and $r_i$, under the minor regularity condition that each feature of $h_r(x; \phi)$ is unique, i.e. there is no $k \neq k' \in \{1, \ldots, d_r\}$ such that $h_r(x; \phi)_k = h_r(x; \phi)_{k'}$ for all $x \in \mathcal{X}$. 
So, we use $\rho_i$ as our representation of $r_i$.
Intuitively, $\rho_i$ is a vector describing the amount of selection in patient $i$ on the sequence features $h_r(x; \phi)$. 

\subsection{Regularization and training} \label{apx:reg_train}

One possible concern is that our estimates may be poor if the outcome model is misspecified, or converges slowly to the truth.
To help address these concerns, we implement a propensity score correction.
Such corrections can help improve the efficiency and robustness of causal estimates, by reducing estimators' sensitivity to nuisance parameters, i.e. those parameters that do not enter into the effect itself.
We use a method based on the Robinson decomposition and Neyman orthogonality, ``structured intervention networks''~\citepapx{Kaddour2021-uu}.
This method allows for high-dimensional, structured treatments.
The basic idea is to build a propensity model of the treatment's representation, $\mathbb{E}_{q_i^a}[h_a(A ; \theta)]$, and use this to adjust the regularization of the outcome model.
 This approach, based on semiparametric theory, can offer many of the benefits of standard propensity score methods developed for low-dimensional treatments, such as reduced bias from regularization, improved robustness to model misspecification, and fast convergence guarantees.
 
 With this propensity correction in place, our complete model is,
\begin{equation} \label{eqn:main-model}
 \begin{split}
 	s_{ij} &\sim \mathrm{Bernoulli}(\sigma(\rho_i^\top h_r(x_{ij}; \phi) + \beta_i))\\
 	y_i &\sim \mathrm{Normal}\left (\gamma_a^\top (\mathbb{E}_{\hat{q}^a_i}[h_a(A ;\theta)] - W \cdot \rho_i - B) + \gamma_r^\top \rho_i + \gamma_y, \tau_y\right),
 \end{split}
\end{equation}
where $\{(x_{ij}, s_{ij})\}_{j=1}^{2m_i} = \{(a_{ij}, 1)\}_{j=1}^{m_i} \cup \{(z_{ij}, 1)\}_{j=1}^{m_i}$. 
The parameters $W$ and $B$ are learned from a propensity model which predicts the treatment representation $\mathbb{E}_{q_i^a}[h_a(A ; \theta)]$ from the confounder representation $\rho_i$,
 \begin{equation} \label{eqn:prop-model}
 	\mathbb{E}_{\hat{q}^a_i}[h_a(A ;\theta)] \sim \mathrm{Normal}(W \cdot \rho_i + B, \tau_e).
 \end{equation}
The intuition behind the propensity-corrected term $\mathbb{E}_{\hat{q}^a_i}[h_a(A ;\theta)] - W \cdot \rho_i - B$ in \Cref{eqn:main-model} is that it describes just those aspects of the treatment that are not explained by the confounder $\rho_i$.
Note that the propensity model describes the learned representation $\mathbb{E}_{\hat{q}^a_i}[h_a(A ;\theta)]$ rather than actual data; it is an auxiliary tool for de-biasing the main model (\Cref{eqn:main-model}), not part of the main model's generative description of the data.

To train the entire model on datasets with large numbers of patients and TCRs, we use a stochastic inference procedure.
We draw minibatches of patients, and for each patient we draw a minibatch of mature TCR sequences and a minibatch of simulated pre-selection repertoire sequences.
We use these minibatches to form an approximation of the log likelihood of the entire dataset.
The latent variables $\rho_i$ and $\beta_i$ are local, per-patient variables, so we use amortized inference with an encoder network.
We place priors on all parameters and perform maximum \textit{a posteriori} inference, learning point estimates of each.
For optimization we employ AMSGrad, an extension of Adam with improved convergence guarantees~\citepapx{Reddi2018-bz}.
 We train the main model (\Cref{eqn:main-model}) and the propensity model (\Cref{eqn:prop-model}) in tandem, following the procedure of \citeapx{Kaddour2021-uu}. In particular, we alternate between (1) updating the parameters of the propensity model, $W,B$ and $\tau_e$, based on the likelihood of the propensity model alone, and (2) updating the rest of the parameters based on the likelihood of the main model alone.
 The propensity model updates are performed less often than the main model updates (every 10 steps).
  \section{DeepRC and DeepRC$^\star$} \label{apx:deeprc}

In this section we describe the DeepRC repertoire classification model proposed by \cite{Widrich2020-hj}, and our modified version, DeepRC$^\star$, which can be used to estimate causal effects under the assumption of no confounding.

We first briefly introduce some additional definitions. Let $\{\tilde{a}_{ij} : j \in \tilde{m}_i\}$ denote the set of unique sequences in the data, and let $c_{ij}$ be the number of observed counts of sequence $\tilde{a}_{ij}$.
The empirical distribution of the data can then be written in terms of $c$ as $\hat{q}_i^a = \frac{1}{m_i} \sum_{j=1}^{m_i} \delta_{a_{ij}} = \frac{1}{m_i} \sum_{j=1}^{\tilde{m}_i} c_{ij} \tilde{a}_{ij}$, where $\tilde m_i =  |\{\tilde{a}_{ij} : j \in \tilde{m}_i\}|$ is the number of unique sequences.

\parhead{DeepRC}
The DeepRC architecture employs an attention mechanism. To define it, we introduce the function,
\begin{equation}
	g(a; \theta, \eta) \equiv \exp(\eta_2^\top \tilde h(h_a(a;\theta);\eta_1) / \sqrt{\tilde d})
\end{equation} 
Here $\tilde h: \mathbb{R}^{d_a} \to \mathbb{R}^{\tilde d}$ is a neural network parameterized by $\eta_1$, and $\eta_2 \in \mathbb{R}^{\tilde d}$ is an additional parameter.
In the language of attention, the output of $\tilde h(h_a(a;\theta);\eta_1)$ is described as a ``key'', $\eta_2$ is a ``query'', and the output of $g(a; \theta, \eta)$ is an (unnormalized) ``attention weight''.

DeepRC is designed for a binary outcome or label, and employs a cross-entropy loss, corresponding to a logistic model. 
To define the model, we represent the sequence $a_{ij}$ as a one-hot encoding, such that $a_{ijkl} = 1$ if the $k$th letter of the sequence is $l$ and $a_{ijkl} = 0$ otherwise. 
Then the conditional distribution of $y_i$ given a set of sequences $\{a_{i1}, \ldots, a_{im_i}\}$ is given by,
\begin{equation} \label{eqn:deeprc}
	Y_i \sim \mathrm{Bernoulli}\left(\sigma \left(\gamma_a^\top \frac{\sum_{j=1}^{\tilde m_i} h_a(\log(c_{ij}) \tilde a_{ij}; \theta) g(\log(c_{ij}) \tilde a_{ij} ; \theta, \eta) \mathbb{I}[g(\log(c_{ij}) \tilde a_{ij} ; \theta, \eta) > \mathcal{Q}_{0.9}(\hat{g}_i)]}{\sum_{j=1}^{\tilde m_i} g(\log(c_{ij}) \tilde a_{ij} ; \theta, \eta)\mathbb{I}[g(\log(c_{ij}) \tilde a_{ij} ; \theta, \eta) > \mathcal{Q}_{0.9}(\hat{g}_i)]} + \gamma_y \right) \right)
\end{equation}
where $\sigma = 1/(1 + \exp(-x))$ is the logistic function, $\hat{g}_i$ is the empirical distribution of the attention weights
$$\hat{g}_i = \frac{1}{\tilde m_i} \sum_{j=1}^{\tilde m_i} \delta_{g(\log(c_{ij}) \tilde a_{ij} ; \theta, \eta)},$$
and $\mathcal{Q}_{0.9}(\hat{g}_i)$ is the value of the 90th percentile of the distribution.
In the language of attention, $h_a(\log(c_{ij}) \tilde a_{ij}; \theta)$ is the ``value'' and $g(\log(c_{ij}) \tilde a_{ij} ; \theta, \eta)/\sum_{j=1}^{\tilde m_i} g(\log(c_{ij}) \tilde a_{ij} ; \theta, \eta)$ is the ``attention weight''.

The indicator $\mathbb{I}[g(\log(c_{ij}) \tilde a_{ij} ; \theta, \eta) > \mathcal{Q}_{0.9}(\hat{g}_i)]$ sets smaller attention weights to zero (note it is specific to DeepRC rather than attention-based models more broadly).
The motivation for including this term is to reduce computational cost: terms with $g(\log(c_{ij}) \tilde a_{ij} ; \theta, \eta) \le \mathcal{Q}_{0.9}(\hat{g}_i)$ can be ignored when computing the gradient of the log likelihood, reducing both time and memory requirements.

Training of DeepRC proceeds by drawing a minibatch of unique sequences $\{\tilde a_{i1}, \ldots, \tilde a_{i \tilde{m}_i}\}$ uniformly for each patient repertoire, batch-normalizing the encoded sequences~\citep{Ioffe2015-zc}, and using this sample to approximate the sum over unique sequences in the numerator and denominator of \Cref{eqn:deeprc}. 

\parhead{DeepRC$^\star$} 
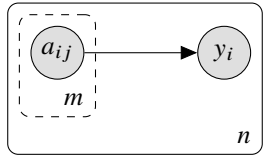
\begin{figure}
\centering
\begin{tikzpicture}

\node[obs] (y) {$y_{i}$};
  \node[obs, left=1.5cm of y] (a) {$a_{ij}$};

\edge {a} {y} ;
\plate[dashed] {pa} {(a)} {$m$} ;
  \plate {zayu} {(a)(y)(pa)} {$n$} ;

\end{tikzpicture}
\caption{\textbf{Causal model without confounding.}} \label{fig:drc_hcm}
\end{figure}
We modified DeepRC slightly such that it provides causal effect estimates under a causal model with no confounding.
In particular, consider the hierarchical causal model in \Cref{fig:drc_hcm}. After collapsing and applying $\sigma$-calculus as in \Cref{apx:id}, we can identify the causal effect as
\begin{equation}
	\pr(y \s  \rmdo(q^a_\star \sim \sigma_{a_\star, \epsilon})) = \int \pr(y \mid (1 - \epsilon) q^a + \epsilon \delta_{a_\star}) \pr(q^a) \di q^a
\end{equation}
To estimate the interventional effect under this model, we need to regress the outcome on the repertoire distribution, i.e. we need to estimate a model of $\pr(y \mid q^a)$.

Although DeepRC regresses from repertoires to outcomes, it does not directly provide such a model. The key issue is that DeepRC is not a function just of the empirical distribution of sequences $\hat q_i^a$. Instead, it depends also on the total number of sequences $m_i$.
In general, this number will depend on the experimental measurement procedure, rather than just the patient's repertoire (recall that the true number of T cells in the repertoire is many orders of magnitude larger than the number of samples that are actually recorded by sequencing).
We modify DeepRC to remove the dependence on $m_i$.

The DeepRC$^\star$ model (for a binary outcome) is,
\begin{equation}
	Y_i \sim \mathrm{Bernoulli}\left(\sigma\left( \gamma_a^\top \frac{\mathbb{E}_{q_i^a}[h_a(A;\theta) g(A ; \theta, \eta)]}{\mathbb{E}_{q_i^a}[g(A ; \theta, \eta)]} + \gamma_y\right) \right)
\end{equation}
Approximating the expectations with the empirical distribution of repertoire sequences yields,
\begin{equation}
	\frac{\mathbb{E}_{q_i^a}[h_a(A;\theta) g(A ; \theta, \eta)]}{\mathbb{E}_{q_i^a}[g(A ; \theta, \eta)]} \approx \frac{\mathbb{E}_{\hat{q}_i^a}[h_a(A;\theta) g(A ; \theta, \eta)]}{\mathbb{E}_{\hat{q}_i^a}[g(A ; \theta, \eta)]} = \frac{\sum_{j=1}^{\tilde m_i} c_{ij} h_a(\tilde a_{ij}; \theta) g( \tilde a_{ij} ; \theta, \eta) }{\sum_{j=1}^{\tilde m_i} c_{ij} g(\tilde a_{ij} ; \theta, \eta)}
\end{equation}
Written in this way, we can see that there are two key differences from DeepRC (\Cref{eqn:deeprc}). 
First, instead of multiplying the sequence representation by its log counts, i.e. $\log(c_{ij}) \tilde{a}_{ij}$, we multiply the sequence's attention weight by its counts, i.e. $g(\tilde{a}_{ij} ; \theta, \eta) c_{ij}$.
Second, we drop the quantile indicator function; in practice, we found the computational speedups provided by the quantile indicator to be unnecessary for training the model; this may be because we use improved GPU hardware as compared to that used in \cite{Widrich2020-hj}.

To train the model we draw a minibatch of samples from the full repertoire $\{a_{i1}, \ldots, a_{i m_i}\}$, rather than just the set of unique sequences, and approximate the expectations as,
\begin{equation}
	\frac{\mathbb{E}_{\hat{q}_i^a}[h_a(A;\theta) g(A ; \theta, \eta)]}{\mathbb{E}_{\hat{q}_i^a}[g(A ; \theta, \eta)]} \approx \frac{\sum_{j \in \mathcal{S}} h_a(a_{ij} ; \theta) g(a_{ij} ;\theta, \eta)}{\sum_{j \in \mathcal{S}} g(a_{ij} ;\theta, \eta)}
\end{equation}
where $\mathcal{S} \subseteq \{1, \ldots, m_i\}$ is the set of indices of the minibatch.
We do not use batch normalization, to avoid the complexities it introduces at test time, when the model is used to estimate the effects of held-out sequences.

 \section{Details on Semisynthetic Data} \label{apx:semisynthetic}

In this section we describe our procedure for constructing semisynthetic data in detail.

Let $\kappa \in \X$ denote a motif -- a string of amino acids -- of length $|\kappa|$. 
We define the \textit{motif injection} function $\mathcal{I}(q ; \kappa, L)$ which adds the motif $\kappa$ into a sequence distribution at position $L$: if $\tilde q = \mathcal{I}(q ; \kappa, L)$, then samples from $\tilde q$ are generated as
\begin{equation}
\begin{split}
	X &\sim q\\
X_{L:L+|\kappa|} &= \kappa.
\end{split}
\end{equation}
That is, for each sample from $q$, we overwrite whatever letters were at positions $L, \ldots, L+|\kappa|-1$ with the motif $\kappa$~\citepapx{Widrich2020-hj}.
We also define a \textit{motif recognition} function $D(x ; \kappa)$. This function takes value 1 if the motif $\kappa$ appears in the sequence, and zero otherwise, i.e.
\begin{equation}
	D(X ; \kappa) = \mathbb{I}\left[ \sum_{j=1}^{|X| - |\kappa|+1} \mathbb{I}(X_{j:j+ |\kappa|-1} = \kappa) > 0\right].
\end{equation}

To construct the synthetic pre-selection repertoire, we start with values $\tilde{q}^z_i$ learned from real repertoires via IGoR~\citepapx{Pavlovic2021-sn}.
Then we inject the confounded motif $\kappa^\con$ into all patients, and the causal motif $\kappa^\cau$ into some patients. The fraction of TCRs that contain each motif is $\eta$, which we will choose to be small ($\eta \le 0.01$).
\begin{equation}
\begin{split}
	\zeta_i &\sim \mathrm{Bernoulli}(0.4)\\
	q_i^z &= (1 - \eta - \eta \zeta_i) \tilde q_{i}^z  + \eta \mathcal{I}(\tilde q_{i}^z ; \kappa^\con) +  \eta \zeta_i \mathcal{I}(\tilde q_{i}^z ; \kappa^\cau)
\end{split}
\end{equation}
We take samples $z_{ij} \sim q_i^z$ to form a simulated pre-selection repertoire. 

Next we generate the confounder and relative fitness. We generate
\begin{equation}
\begin{split}
	U_{i} &\sim \mathrm{Bernoulli}(0.4)\\
	r_i(x) &= \exp\left((6 u_{i} - 5) D(x ; \kappa^\con)\right)
\end{split}
\end{equation}
Sequences containing the confounded motif are subject to positive selection ($r_i(x) > 1$) when $u_i = 1$, and negative selection ($r_i(x) < 1$) when $u_i = 0$.
The particular values of the confounder probability (0.4) and selection coefficient ($\exp(1)$ for positive selection and $\exp(-5)$ for negative selection) are chosen such that the confounded motif will have a similar occurrence pattern in the mature repertoire as the causal motif: 40\% of patients will have TCRs with the motif at prevalence $\approx \eta$, while in the rest of the patients the motif will be nearly absent.

Next we construct the mature repertoire.
To do so, we use a finite sample approximation to the selection equation (\Cref{eqn:fitness}). We first draw $m'$ fresh samples $z'_{i1}, \ldots, z'_{im'}$ from $q_i^z$ (independent of $\{z_{ij}\}_{j=1}^{m_i}$) and then take 
\begin{equation} \label{eqn:qA_semisynth}
	q_i^a = \frac{1}{\sum_{j=1}^{m'} r_i(z'_{ij})} \sum_{j=1}^{m'} r_i(z'_{ij}) \delta_{z'_{ij}}
\end{equation}
We sample the mature repertoire sequences as $a_{ij} \sim q_i^a$ for $j \in \{1, \ldots, m_i\}$.

Finally, the outcome variable $y$ depends on (a) the presence of a subpopulation of TCRs with the causal motif, and (b) the confounder:
\begin{equation}
	Y_i \sim \mathrm{Normal}(\gamma_a \mathbb{I}(\mathbb{E}_{q_i^a}[D(x, \kappa^\cau)] > \eta/2) + \gamma_u u_i + \gamma_0, \tau_y).
\end{equation}
We focus on the regime where $\gamma_u$ is substantially larger than $\gamma_a$ ($\gamma_u = 2$ versus $\gamma_a = 0.4$). 
This creates the impression, if confounding is ignored, that sequences with the confounded motif $\kappa^\con$ have a larger effect than sequences with the true causal motif $\kappa^\cau$. 
We set the standard deviation $\tau_y$ to be small compared to $\gamma_u$ and $\gamma_a$ ($\tau_y=0.1$).
Previous simulation studies of repertoire classification have focused on a regime with no noise, so using low values of $\tau_y$ ensures our results are at least somewhat comparable to these previous studies.

We used the real repertoire data from \citeapx{Emerson2017-je} (https://clients.adaptivebiotech.com/pub/emerson-2017-natgen) as the basis for our semisynthetic data.
The dataset contains 786 patients, and there is an average of 31,000 sequences per patient (minimum 177, maximum 183,000), \add{which we use to set the total number of patients and sequences in the synthetic data}. 
The pre-selection repertoire $\tilde q_i^z$ is estimated from the non-productive sequences via IGoR, following the protocol in \Cref{apx:igor}. 
We use motifs of length $|\kappa^\cau|=|\kappa^\con|=3$, and inject them at position $L=3$. 
To ensure that the motifs are not very common or extremely rare in existing sequences, we first sort existing length 3 subsequences in the initial repertoires by frequency. We then choose $\kappa^\cau$ and $\kappa^\con$ at random from among those subsequences whose frequency is between the 10th and 20th percentile of all subsequences.
We work only with the CDR3 region of the TCR$\beta$, starting immediately after the cysteine on 5' side and ending immediately before the phenylalanine on the 3' side.

It is important to note that in our semisynthetic experiments, we train CAIRE's fitness model directly on simulated productive sequences from the pre-selection repertoire $\{z_{ij}\}_{j=1}^{m_i}$, rather than run inference with IGoR on nonproductive sequences and then draw productive sequences from the estimated model.
This is because the underlying parametric model used by IGoR cannot describe the motif-injected repertoire $\tilde q$; we are also limited by the fact that inference in IGoR is computationally intensive.
Hence, our semisynthetic experiments should be understood as an evaluation of the CAIRE model alone (\Cref{eqn:main-model,eqn:prop-model}), and not in combination with IGoR.
 \section{Details on Architectures and Training} \label{apx:models}

In this section we detail CAIRE's model architecture and training, as well as that of other methods we compare to in experiments.

\subsection{Models} \label{apx:semisynth_models}

We consider the following models.

\parhead{CAIRE} The CAIRE model (\Cref{eqn:main-model,eqn:prop-model}), restated, is
\begin{align}
 	s_{ij} &\sim \mathrm{Bernoulli}(\sigma(\rho_i^\top h_r(x_{ij}; \phi) + \beta_i))\\
 	y_i &\sim \mathrm{Normal}\left (\gamma_a^\top (\mathbb{E}_{\hat{q}^a_i}[h_a(A ;\theta)] - W \cdot \rho_i - B) + \gamma_r^\top \rho_i + \gamma_y, \tau_y\right)
 \end{align}
along with the separate propensity model,
\begin{equation}
 	\mathbb{E}_{\hat{q}^a_i}[h_a(A ;\theta)] \sim \mathrm{Normal}(W \cdot \rho_i + B, \tau_e).
 \end{equation}

\parhead{No propensity CAIRE} This model is a variant of CAIRE without the propensity correction.
\begin{align}
 	s_{ij} &\sim \mathrm{Bernoulli}(\sigma(\rho_i^\top h_r(x_{ij}; \phi) + \beta_i))\\
 	y_i &\sim \mathrm{Normal}\left (\gamma_a^\top \mathbb{E}_{\hat{q}^a_i}[h_a(A ;\theta)] + \gamma_r^\top \rho_i + \gamma_y, \tau_y\right).
 \end{align}

\parhead{Uncorrected} This model does not correct for confounding.
\begin{equation}
 	y_i \sim \mathrm{Normal}\left (\gamma_a^\top \mathbb{E}_{\hat{q}^a_i}[h_a(A ;\theta)] + \gamma_y, \tau_y\right).
\end{equation}

\parhead{DeepRC$^\star$}
This model is a variant of DeepRC~\citepapx{Widrich2020-hj}, adapted for causal effect estimation (\Cref{apx:deeprc}).
Since the outcome variable is continuous, we use a Gaussian outcome distribution.
\begin{equation}
	y_i \sim \mathrm{Normal}\left( \gamma_a^\top \frac{\mathbb{E}_{\hat q_i^a}[h_a(A;\theta) g(A ; \theta, \eta)]}{\mathbb{E}_{\hat q_i^a}[g(A ; \theta, \eta)]} + \gamma_y, \tau_y \right) 
\end{equation}

\parhead{Attention CAIRE} This model combines CAIRE with the attention-based model parameterization from DeepRC$^\star$.
\begin{align}
 	s_{ij} &\sim \mathrm{Bernoulli}(\sigma(\rho_i^\top h_r(x_{ij}; \phi) + \beta_i))\\
 	y_i &\sim \mathrm{Normal}\left (\gamma_a^\top \left(\frac{\mathbb{E}_{\hat q_i^a}[h_a(A;\theta) g(A ; \theta, \eta)]}{\mathbb{E}_{\hat q_i^a}[g(A ; \theta, \eta)]} - W \cdot \rho_i - B\right) + \gamma_r^\top \rho_i + \gamma_y, \tau_y\right),
 \end{align}
along with the separate propensity model,
\begin{equation}
 	\frac{\mathbb{E}_{\hat q_i^a}[h_a(A;\theta) g(A ; \theta, \eta)]}{\mathbb{E}_{\hat q_i^a}[g(A ; \theta, \eta)]} \sim \mathrm{Normal}(W \cdot \rho_i + B, \tau_e).
 \end{equation}

 \subsection{Architecture details} \label{apx:semisynth_arch}
 
 In this section we detail the architecture of the neural networks in the models.
 
 \begin{itemize}
 	\item 
$h_a(\cdot;\theta)$: This network extracts features from TCR CDR3 sequences, for estimating causal effects. Its architecture follows the feature extractors in DeepRC~\citepapx{Widrich2020-hj}.

We first encode each sequence as follows: (1) we append to the amino acid sequence an end token, marking the final position, (2) we one-hot encode the entire sequence, such that $a_{ijkl} = 1$ if the amino acid in the $k$th position is $l$ and $a_{ijkl} = 0$ otherwise, and (3) we pad the one-hot encoded sequence with zeros out to the maximum sequence length observed in the dataset, i.e. $a_{ijkl} = 0$ for all $k$ greater than the sequence length, and (4) we add three position features describing the sequence start, end, and center, as described in detail in \citeapx{Widrich2020-hj}, Appendix A2. The fully encoded sequence $a_{ij}$ is a matrix of size $L_\mathrm{max} \times 24$,  where $L_{\mathrm{max}}$ is the maximum length of sequences in the dataset, and 24 = (20 amino acids) + (1 end token) + (3 position features).
 
 Given a sequence in this encoding, the network $h_a$ applies a 1D convolutional neural network (CNN) with a SELU nonlinearity~\citepapx{Klambauer2017-be}, then takes the maximum across positions: $h_a(a_{ij};\theta)_\ell = \max_k(\mathrm{SELU}(\mathrm{CONV1D}(a_{ij};\theta)_{k\ell}))$, where $\mathrm{CONV1D}$ is the 1D convolution function.
 The resulting output is in $\mathbb{R}^{d_a}$, with $d_a$ corresponding to the number of channels in the CNN.
 The trainable parameters $\theta$ are the weights and biases of the CNN, i.e. the convolutional filters. The convolution kernel size $\zeta$ (measured in number of amino acids) is a hyperparameter.
 
 \item $h_r(\cdot; \phi)$: This network extracts features from TCR CDR3 sequences, for estimating fitness. 
 	Its first layer has an identical architecture to $h_a$, that is it can be written in the form $h_r(a_{ij};\phi) = \tilde h_r( h_a(a_{ij}; \phi_0); \phi_1)$. (The parameters are not shared with $h_a$, however, i.e. $\phi_0 \neq \theta_0$.)
 	The remaining layers of $h_r$ are feedforward: $\tilde{h}_r$ is a feedforward neural network with SELU nonlinearities.
 
 \item $g(a;\theta, \eta)$: This network is used in the models DeepRC$^\star$ and Attention CAIRE, and outputs a scalar weight in $\mathbb{R}_+$. Its architecture follows the attention network used in DeepRC~\citepapx{Widrich2020-hj}.
 	In particular, the first layer is identical to, and shares parameters with, $h_a$. More precisely, we can write $g(a_{ij};\theta, \eta) = \exp(\tilde{g}(h_a(a;\theta);\eta))$. 
 	The remaining layers are feedforward: $\tilde g(\cdot ;\eta)$ is a two-layer feedforward neural network with SELU nonlinearities.

 \end{itemize}

 \subsection{Amortized inference}
 
 The selection representation $\rho_i$ and the offset term $\beta_i$ are per-patient latent variables. To achieve scalable, stochastic gradient-based inference on large datasets, we amortize inference of $\rho_i$ and $\beta_i$ across patients. The inference network is parameterized as,
 \begin{equation} \label{eqn:encoder}
 	(\rho_i, \beta_i) = \mathrm{enc}(\{(x_{ij}, s_{ij})\}_{j=1}^{2m_i}; \theta', \lambda) = \tilde{e}(\mathbb{E}_{\hat q_i^a}[h_a(A;\theta')] - \mathbb{E}_{\hat q_i^z}[h_a(A;\theta')]; \lambda), 
 \end{equation}
 where $\tilde{e}(\cdot ; \lambda)$ is a neural network. The neural network $h_a$ has the architecture described in \Cref{apx:semisynth_arch} (but $\theta' \neq \theta$).
 This encoder parameterization is intended to help the model focus on differences between the pre-selection and post-selection repertoires, by taking the difference between the embedding of $q_i^a$ (estimated from mature repertoire) and the embedding of $q_i^z$ (estimated from the pre-selection repertoire).
 The function $\tilde{e}$ is a single-layer feedforward neural network with SELU nonlinearity, i.e. it takes the form $\tilde{e}(\cdot ;\lambda) = \mathrm{Linear}_{\lambda_1}(\mathrm{SELU}(\mathrm{Linear}_{\lambda_0}(\cdot)))$ where $\mathrm{Linear}$ denotes a linear layer.
 The parameters $\lambda=(\lambda_0, \lambda_1)$ determine the weights and biases of the linear layers.

 In preliminary experiments, we compared to an alternative amortization network that pools the pre-selection and mature repertoire sequences, rather than take the difference in their representation. We found that the parameterization in \Cref{eqn:encoder} encourages the model to focus on features that distinguish between the pre-selection and post-selection repertoire, and helps avoid poor solutions where $\rho_i$ contains additional information about $q^a_i$ beyond the relative fitness $r_i$.
 
Our inference network provides an estimate of the maximum likelihood value of $\rho_i$ and $\beta_i$.
We also explored amortized variational inference of $\rho_i$ and $\beta_i$, following the logic of variational autoencoders~\citepapx{Kingma2019-iy}.
However, we found convergence to be quite poor in preliminary experiments.
Moreover, since there are many sequences per patient, uncertainty in the per-patient latent variables $\rho_i$ and $\beta_i$ appeared to be low.
We therefore did not pursue this variational inference approach further.

\subsection{Effect estimates}

After training, we can compute the estimated effect using \Cref{eqn:softID} and the empirical distribution of $\pr(q^a, r)$, or more precisely, the empirical distribution of the estimates $\hat{q}_i^a$ and $\rho_i$. For CAIRE, the estimated effect is,
\begin{align}
	\textsc{ate}(a_\star, \epsilon) \approx &\frac{1}{n}\sum_{i=1}^n \gamma_a^\top (\mathbb{E}_{(1-\epsilon)\hat{q}^a_i + \epsilon \delta_{a_\star}}[h_a(A ;\theta)] - W \cdot \rho_i - B) + \gamma_r^\top \rho_i + \gamma_y\\
	& - \frac{1}{n} \sum_{i=1}^n \gamma_a^\top (\mathbb{E}_{\hat{q}^a_i}[h_a(A ;\theta)] - W \cdot \rho_i - B) + \gamma_r^\top \rho_i + \gamma_y\\
	= & \epsilon \gamma_a^\top \left( h_a(a_\star \s \theta) - \frac{1}{n} \sum_{i=1}^n \mathbb{E}_{\hat{q}_i^a}[h_a(A \s \theta)]\right)
\end{align}
For the No propensity CAIRE model and the Uncorrected model, the estimated effect works out to the same expression.

For DeepRC$^\star$ and the Attention CAIRE model, the estimated effect becomes,
\begin{equation}
\begin{split}
	\textsc{ate}(a_\star, \epsilon) \approx \frac{1}{n} \sum_{i=1}^n \gamma_a^\top \bigg(&\frac{(1-\epsilon)\mathbb{E}_{\hat q_i^a}[h_a(A;\theta) g(A ; \theta, \eta)] + \epsilon h_a(a_\star \s \theta) g(a_\star)}{(1-\epsilon)\mathbb{E}_{\hat q_i^a}[g(A ; \theta, \eta)] + \epsilon g(a_\star ; \theta, \eta)}\\
	& - \frac{\mathbb{E}_{\hat q_i^a}[h_a(A;\theta) g(A ; \theta, \eta)]}{\mathbb{E}_{\hat q_i^a}[g(A ; \theta, \eta)]}\bigg).
\end{split}
\end{equation}

\subsection{Training details}

We use early stopping for regularization, with a validation set consisting of 12.5\% of the data.
As our stopping condition we use a score measuring the predictive performance of CAIRE.
A typical choice of early stopping score is the training objective; however, since there are large number of sequences in each individual's repertoire, the overall value of the training objective is dominated by the contribution of the selection model (\Cref{eqn:main-model}, line 1), and the contribution of the outcome model (\Cref{eqn:main-model}, line 2) is washed out.
To create a more balanced score, we heuristically combine performance metrics for the selection model and the outcome model, which are each between zero and one.
In particular, we sum (1) the accuracy of the selection model in predicting $s$ and (2) the coefficient of determination ($R^2$) of the outcome model with $y$.
We compute this score on the validation set after every 500 iterations, and take the model corresponding to the best recorded value over the entire training run.

We place priors on the main parameters of the model, $\gamma$, $\rho_i$, $\beta_i$, $W$, $B$ and $\tau$ (\Cref{apx:semisynth_hyperparameters}).
Following previous work on autoencoders, the prior on $\rho_i$ and $\beta$ is annealed during training: the log likelihood of the prior is weighted by a factor $\xi_t$ which increases linearly from 0 to 1~\citepapx{Fu2019-zt}.

The sequence feature extraction network $h_a(\cdot;\theta)$ is computed in low precision, using bfloat16~\citepapx{Kalamkar2019-hp}. This occurs wherever the $h_a$ function is used, i.e. not just in the model of $y_i$, but also in the first layer of $h_r$ and in $g$.
All other computations are at single precision, float32.

All models were implemented in the PyTorch-based probabilistic programming language Pyro (version 1.8.4)~\citepapx{Bingham2019-aa}.
All models were trained using an NVIDIA A100 GPU with 80GB memory.
All models were trained for the same, fixed amount of wall-clock time, 5 minutes. 
This was chosen based on preliminary experiments suggesting that it was sufficient for convergence (according to the diagnostic of \citeapx{Pesme2020-fs}) across the different model classes.

\section{Details on Semisynthetic Experiments} \label{sec:cmv}

\subsection{Evaluation details} \label{apx:semisynth_eval}

For evaluation, we hold out data from 12.5\% of patients.
We then look at the repertoires $\{a_{ij}\}_{j=1}^m$ of each held-out patient who had the causal motif injected, i.e. each patient $i$ with $\zeta_i = 1$ (\Cref{apx:semisynthetic}).
We examine the effect of interventions that use each of their repertoire sequences, i.e. $\ateest(a_\star = a_{ij}, \epsilon = 0.01)$ for $j \in \{1, \ldots, m\}$.
We then evaluate how well this effect estimate can discriminate causal sequences, $\{a_{ij}: D(a_{ij} \s \kappa^\cau) = 1\}$, from non-causal sequences,  $\{a_{ij}: D(a_{ij} \s \kappa^\cau) = 0\}$.
Performance is quantified with the area under the precision-recall curve (PR-AUC).
The final performance of the model is the average PR-AUC across the held-out motif-injected patients.

To evaluate the statistical significance of average differences between model performance, we use a permutation-based t-test (scipy \verb|ttest_ind| with 10000 permutations).

\subsection{Hyperparameters} \label{apx:semisynth_hyperparameters}

In this section we describe our handling of hyperparameters in the semisynthetic experiments.
Key hyperparameters (especially those governing the overall dimensionality of a model) were optimized for each method that we compare to, on each dataset.
Due to computational cost, optimizing all hyperparameters was infeasible; 
the remaining hyperparameters were fixed based on preliminary experiments, and based on previous work on DeepRC~\citepapx{Widrich2020-hj} and SIN~\citepapx{Kaddour2021-uu}.

Hyperparameter optimization was performed using BoTorch~\citepapx{Balandat2020-pv}, via the Ax interface (\url{https://github.com/facebook/Ax}, version 0.3.4). For CAIRE's search space, Ax's default strategy consisted of a Sobol sequence for six iterations, followed by Bayesian optimization with a Gaussian process and the expected improvement acquisition function.
We use 10 rounds of hyperparameter optimization for all experiments.

 \begin{table} 
  \centering
  \caption{\textbf{Hyperparameters for semisynthetic experiments.} Hyperparameters with a search range are optimized over that range using Bayesian optimization. Remaining hyperparameters are fixed at the given value.} \label{tab:hyperparams_semisynth} \label{tbl:hyperparam}
  \begin{tabular}{ccc}
  \hline
  Hyperparameter & Value or search range \\
  \hline
Dimension $d_r$ of the selection representation $\rho_i$ & [2, 32]\\
  Dimension $d_a$ of the repertoire representation $\mathbb{E}_{q_i^a}[h_a(A;\theta)]$ & [8, 32]\\
  Kernel size of CNNs in $h_a$, $h_r$, and $\mathrm{enc}$ & $\{5, 7, 9\}$ \\
  Number of channels of the CNNs in $h_r$ and $\mathrm{enc}$  & 8\\
  Number of layers in the selection network $\tilde{h}_r$ &  3\\
  Dimension of hidden layers in the selection network $\tilde{h}_r$  & 16\\
  Number of layers in the attention network $\tilde g$ &  2\\
  Dimension of hidden layers in the attention network $\tilde g$ &  32 \\
  Prior on entries of $\gamma_a$, $\gamma_r$ and $\gamma_y$ &  $\mathrm{Normal}(0, 100)$\\
  Prior on entries of $\rho_i$ &  $\mathrm{Normal}(0, 1)$\\
  Prior on $\beta_i$ &  $\mathrm{Normal}(0, 10)$\\
  Prior on entries of $W$ and $B$ & $\mathrm{Normal}(0, 10)$\\
  Prior on $\tau_y$ and $\tau_e$ & $\mathrm{LogNormal}(-1, 2)$\\
  Dimension of hidden layer in the encoder network $\tilde{e}$ & 8\\
  Batch size: patients &  8\\
  Batch size: sequences per patient &  16,384\\
  Learning rate  & 0.01 \\
  Weight decay & 0.01 \\
  Validation set size & 12.5\% of patients \\
  Test set size & 12.5\% of patients\\
  Training time & 5 minutes \\
  Annealing time & 3 minutes
  \end{tabular}
  \end{table}

\subsection{Experiments} \label{apx:semisynth_exp}

\parhead{Model comparison} In this experiment, we compare the methods in \Cref{apx:semisynth_models}. We draw 15 independent semisynthetic datasets, generated as in \Cref{apx:semisynthetic} (with motif injection rate $\eta = 0.01$). We then evaluate each method on each dataset.

\begin{figure}
\centering
\includegraphics[width=0.4\textwidth]{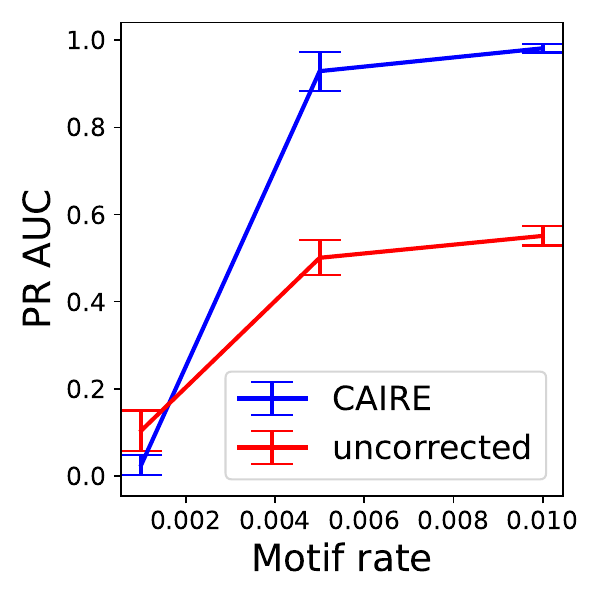}
	\caption{\textbf{Performance on semisynthetic data with increasing motif injection rate $\eta$.}} \label{fig:semisynth_scaling}
\end{figure}

\parhead{Low motif rates} In this experiment we compare CAIRE to the Uncorrected method (\Cref{apx:semisynth_models}). We evaluate on semisynthetic datasets in which the motif injection rate $\eta$ is set to 0.001, 0.005, and 0.01. For each value of $\eta$, we sample 10 independent semisynthetic datasets, and evaluate each method on each dataset.

\parhead{No confounding} In this experiment we compare CAIRE to the Uncorrected method (\Cref{apx:semisynth_models}). We sample 10 independent semisynthetic datasets in which confounding does not contribute to the outcome, i.e. with $\gamma_u = 0$. We set the motif injection rate $\eta$ to 0.01.
We then evaluate each model on each dataset.

 \section{Details on COVID Severity Data} \label{apx:application_data}

\citeapx{Snyder2020-qg} and \citeapx{Nolan2020-fx} constructed a TCR sequencing dataset from a large cohort of individuals currently or previously infected with SARS-CoV-2.
They also record clinical information about the patients, along with demographic information.
Details on the data, including the sequencing procedure and the patient populations, can be found in the original publications.  
Here, we describe how we preprocess this data to study the causal impacts of TCRs on disease.
The data is available at \url{https://clients.adaptivebiotech.com/pub/covid-2020}. The dataset we used was dated to July 2020, and we accessed it January 2024.

\subsection{Patient Outcomes}

We first constructed an outcome score for each patient measuring their overall disease severity.
The raw data was collected from several different patient cohorts at different study sites, and contains a heterogenous mixture of clinical information, with different variables missing for different patients.
We sought to construct a single summary of severity.
First, we determined whether a patient was hospitalized.
Patients with \verb|hospitalized == True|, \verb|days_in_hospital > 0| or \verb|covid_unit_admit == True| were labeled as \textit{hospitalized}. Those patients with \verb|hospitalized|,  \verb|days_in_hospital| and \verb|covid_unit_admit| all missing were labeled as \textit{missing hospitalization}. All other patients were labeled as having \textit{mild disease}.
Next, we determined whether a patient had severe disease.
Patients with \verb|icu_admit == True| or \verb|death == True| were labeled as having severe disease (regardless of the previous label).
All remaining \textit{hospitalized} patients were labeled as having \textit{moderate disease}.
All remaining \textit{missing hospitalization} patients were labeled as \textit{missing outcome}.
Patients with missing outcome data were dropped from further analysis.

The data contained a limited number of instances of repeated repertoire samples taken from the same patient (14\% of all repertoire sequencing samples).
In instances where the clinical information corresponding to each sample differed, we used the more severe patient outcome. E.g. if a patient is first hospitalized, and then dies, we label them as having \textit{severe disease}, and use the repertoire sequencing data labeled with this clinical outcome.
 When the outcome information was the same across different samples from a patient, we use the repertoire sequencing data corresponding to the earliest doctor visit, i.e. the time-point closest to disease onset.
 When multiple repertoire samples come from the same visit of the same patient, or information on time-points is missing, we take the repertoire sample with the largest number of TCR sequences.
 
 \subsection{Patient Demographics} \label{sec:patient_demo}

The final dataset consisted of $n = 507$ patients, of which 86 had severe disease (patients who were in the ICU or died), 187 had moderate disease (patients who were hospitalized) and 234 had mild disease (patients who were neither hospitalized nor died).

Among patients without missing demographic data, we found the average age was 56.9 (standard deviation 18.2), and 48\% were male. 84.4\% were non-Hispanic Caucasian, 9.0\% Hispanic, 1.8\% Asian or Pacific Islander, and 1.8\% Black or African American, with remaining racial and ethnic groups below 1\%.

\subsection{Mature Repertoires}

\begin{figure}
\centering
\begin{subfigure}[t]{0.4\textwidth}
	\includegraphics[width=\textwidth]{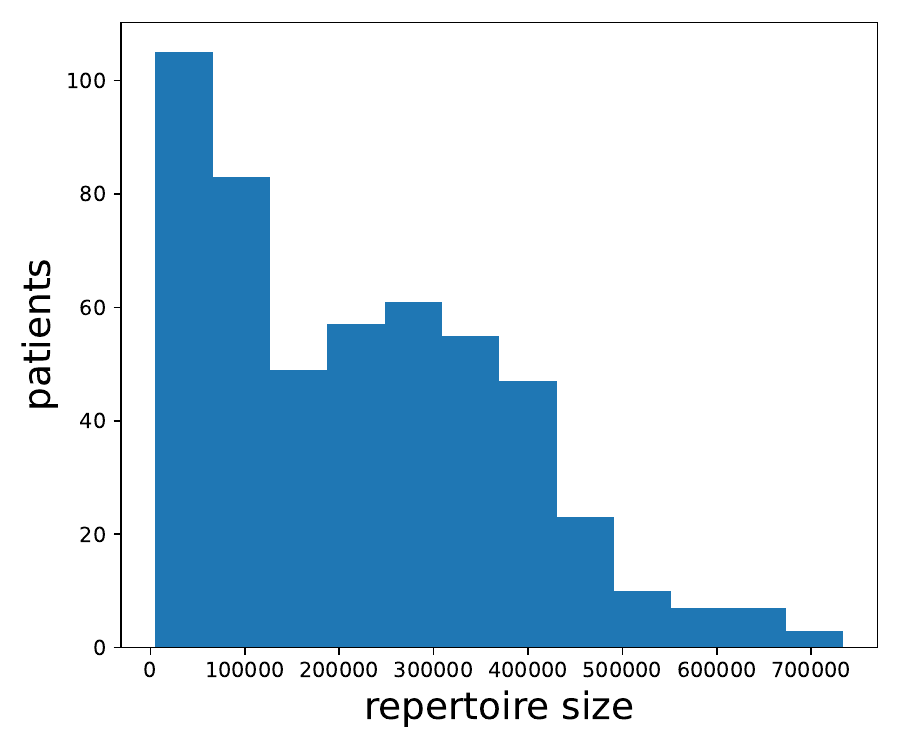}
	\caption{Mature.} \label{fig:snyder_mature_rep_sizes}
\end{subfigure}
\begin{subfigure}[t]{0.4\textwidth}
	\includegraphics[width=\textwidth]{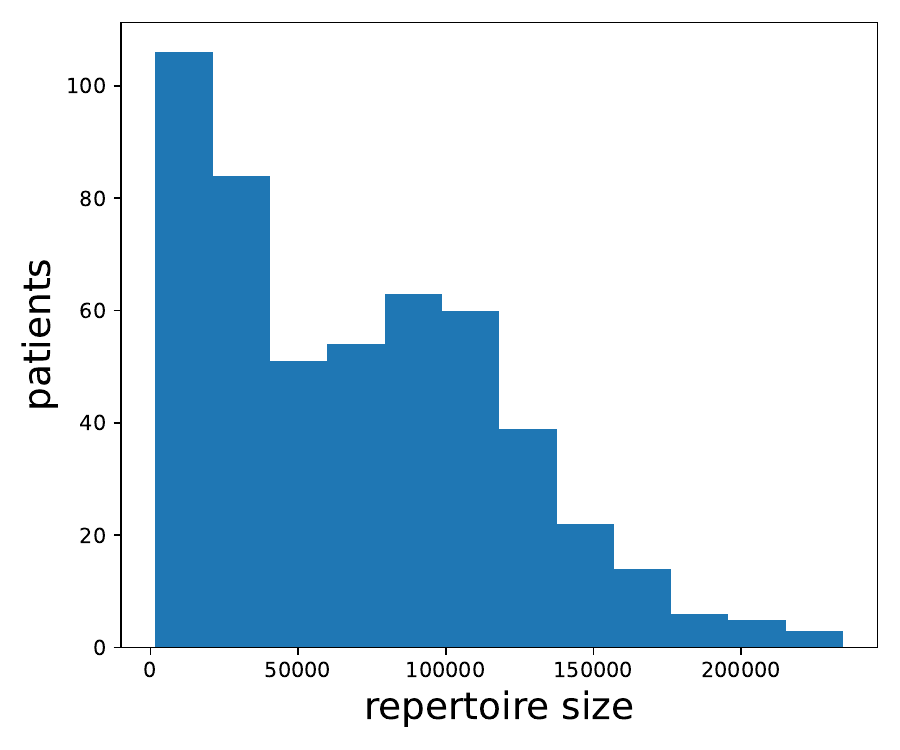}
	\caption{Pre-selection.} \label{fig:snyder_preselect_rep_sizes}
\end{subfigure}
\caption{\textbf{Distribution of repertoire sizes across patients.} Note the pre-selection repertoires are simulated. The histogram bin size is set according to the Freedman-Diaconis rule.} \label{fig:snyder_rep_sizes}
\end{figure}

For each patient we assemble a dataset of mature TCR sequences.
The immunoSEQ assay used for sequencing in \citeapx{Snyder2020-qg} is a bulk RNA sequencing assay, which records a segment of the TCR covering the TCR$\beta$ CDR3.
For the mature repertoire, we take only productive sequences (\verb|frame_type == 'In'|), excluding those with a stop codon or frameshift mutation.
We work with amino acids, taking the CDR3 region starting immediately after the conserved cysteine (C) on 5' side and ending immediately before the conserved phenylalanine (F) on the 3' side.
The distribution of mature repertoire sizes (in terms of number of unique sequences) is shown in \Cref{fig:snyder_mature_rep_sizes}. The mean CDR3 length across all sequences was 12.5 amino acids (standard deviation 1.8), excluding the terminal C and F.
We also record the frequency of each sequence (\verb|productive_frequency|), which is an estimate of how often it occurs in the population of T cells.

\subsection{Pre-selection Repertoires and IGoR} \label{apx:igor}

We estimate each patient's pre-selection TCR repertoire using non-productive sequence data, via IGoR~\citepapx{Marcou2018-so}.

For each patient, we first extract the non-productive sequences (\verb|frame_type != 'In'|), which have frameshift or truncation mutations.
We record the 65 nucleotides at the 3' end of each sequence (in immunoSEQ, priming is done in the constant region of the TCR, on the 3' side).

We then train an IGoR model on each patient's non-productive sequences. Due to the computational limitations of IGoR, patient repertoires with more than 10,000 sequences were randomly subsampled down to 10,000 sequences, and inference over rearrangement scenarios was done with a ``Viterbi-like'' algorithm (using the \verb|--MLSO| flag).
We use the default IGoR model architecture. 
We also use the default reference set of human V(D)J genes; note that our analysis is therefore limited in that it does not account for possible germline allelic variation in these genes, beyond the variation present in IGoR's reference set~\citepapx{Slabodkin2021-wr}).
IGoR reports low sequencing error rates across all patients, suggesting its inferences are reasonable (mean error across patients: 0.0052, 95th percentile: 0.0061, max: 0.0072)

\begin{figure}
\centering
\begin{subfigure}[t]{0.3\textwidth}
	\includegraphics[width=\textwidth]{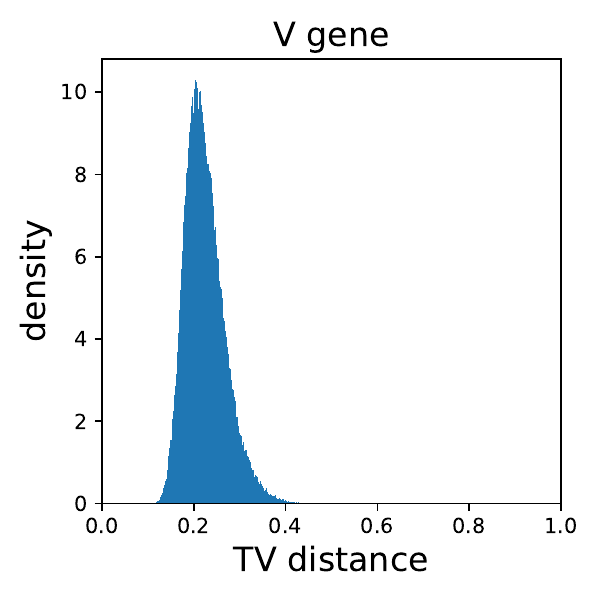}
	\caption{} \label{fig:igor_pairwise_tv_V_gene}
\end{subfigure}
\begin{subfigure}[t]{0.3\textwidth}
	\includegraphics[width=\textwidth]{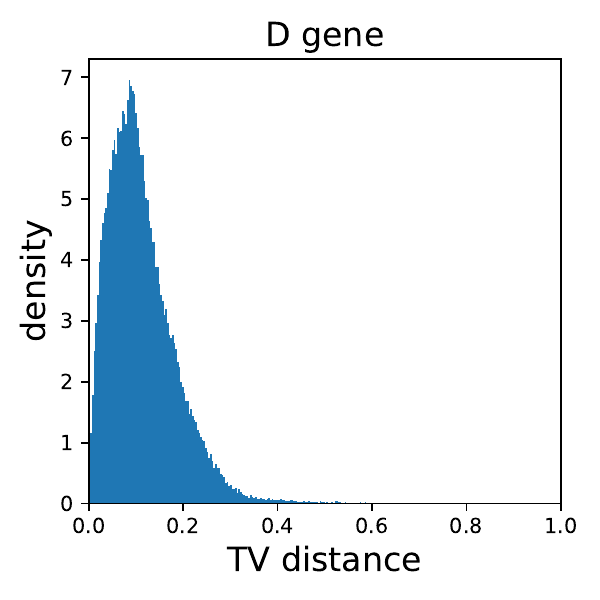}
	\caption{} \label{fig:igor_pairwise_tv_D_gene}
\end{subfigure}
\begin{subfigure}[t]{0.3\textwidth}
	\includegraphics[width=\textwidth]{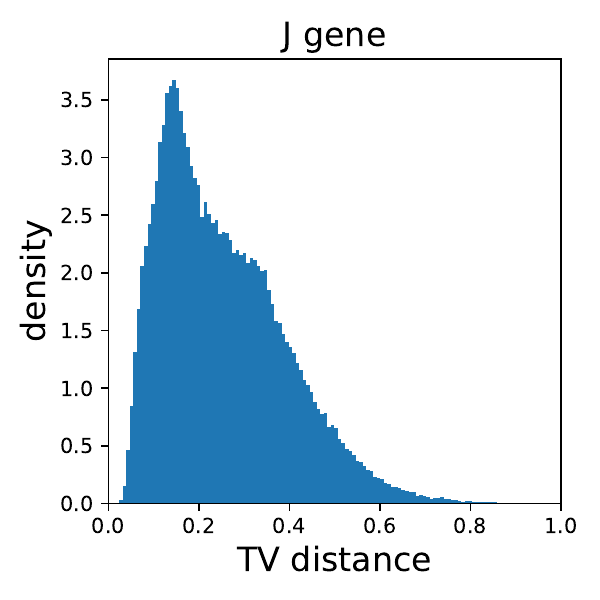}
	\caption{} \label{fig:igor_pairwise_tv_J_gene}
\end{subfigure}
\caption{\textbf{Distribution of gene usage differences among pairs of patients.} The histogram bin size is set according to the Freedman-Diaconis rule.}
\end{figure}

For the pre-selection repertoire to be an effective instrumental variable, it must vary across patients.
We therefore examined variation across patients in the pre-selection repertoire distribution estimated by IGoR.
In particular, we computed the total variation distance between the marginal distribution over V gene usage, among all pairs of individuals (that is, we look at the absolute value of the difference in estimated V gene frequency, summed over all genes)
The distribution exhibits a peak away from zero (\Cref{fig:igor_pairwise_tv_V_gene}).
The D and J gene distributions show similar patterns, with a wide range of distances between patients \Cref{fig:igor_pairwise_tv_D_gene,fig:igor_pairwise_tv_J_gene}.
So, based on these estimates, there is indeed variation among patients in the pre-selection repertoire~\citepapx{Slabodkin2021-wr}.

Once the IGoR model is trained, we use it to draw samples from the distribution of the pre-selection mature repertoire.
In particular, we first sample a number of sequences equal to the total number of TCR sequences measured in the patient.
We then retain only those sampled sequences that are productive (\verb|is_inframe == True|, \verb|anchors_found == True|, and no internal stop codon).
In other words, we use rejection sampling to obtain samples from the pre-selection distribution over productive sequences.
We translate the nucleotide sequences into amino acids, and take the CDR3 region from just after the 5' cysteine to just before the 3' phenylalanine, as with the mature repertoire sequences.
The distribution of generated pre-selection repertoire sizes (in terms of number of sequences) is shown in \Cref{fig:snyder_preselect_rep_sizes}. The mean CDR3 length across all sequences from all patients was 13.2 amino acids (standard deviation 2.3).

\subsection{MIRA data} \label{apx:mira}

\citeapx{Snyder2020-qg} and \citeapx{Nolan2020-fx} also collect a dataset of TCRs that bind SARS-CoV-2 epitopes, using MIRA assays~\citepapx{Klinger2013-rn,Klinger2015-yx}.
In these experiments, patient T cells are screened against viral peptides, and TCR sequences from cells that bind each epitope are recorded.
The epitopes were delivered exogenously, as pools of synthetic peptides.
The epitope sequences were chosen from among those viral peptides predicted by NetMHCpan to bind common class I and class II MHC HLA genes~\citepapx{Nielsen2003-xr,Andreatta2016-qx}. 
In some experiments overlapping or nearby peptides were pooled together, in which case one cannot distinguish precisely which epitope a TCR has bound.  

Our aim is to use this data to evaluate how well causal estimates from CAIRE predict \textit{in vitro} binding.
Following the strategy described in \Cref{apx:application_eval}, we will compare binding TCRs from patients to repertoire sequences drawn from the same patient.
For this reason, we focus just on MIRA experiments performed on patients for which we also have repertoire sequencing data.

We ignore nonproductive TCR sequences, under the assumption that they represent false positives.
We also focus just on MIRA data from a panel of class I MHC epitopes or a panel of class II MHC epitopes, dropping the smaller-scale ``minigene'' experiments, which do not provide exact knowledge of the epitope.

The final dataset contains a total of 32,770 binders and 11 million unlabeled repertoire sequences, drawn from 71 patients.
Binding assays with class I MHC epitopes were performed on 66 patients, and assays with class II were performed on 6 patients; 28,640 TCR sequences were found that bound class I in total, while 4,130 bound class II.
The average number of binders per patient was 462, with a minimum of 12 and a maximum of 2,344.
The average number of repertoire sequences per patient was 151,000, with a minimum of 13,910 and a maximum of 565,000.
There were 265 class I epitopes with at least one TCR binder, and 56 class II epitopes.

 \section{\textit{In vitro} Binding Comparison Method} \label{apx:application_eval}

In this section we detail our strategy for comparing causal effect estimates to \textit{in vitro} TCR binding data.
The key challenge is that the binding assay (\Cref{apx:mira}) only provides data on the sequences of TCRs that bind a target, and does not provide data on the sequences of TCRs that do not bind the target.
In other words, it provides only positive examples of binders, and not negative examples of non-binders.
To evaluate our models, we will therefore combine the MIRA data with unlabeled data from unbiased repertoire sequencing.
In this section, we formalize our evaluation mathematically.

We first describe the data generating process. We assume that TCR sequences are drawn from some underlying distribution $q_0(x)$, corresponding to the distribution of TCRs within a patient or group of patients.
We further assume that the probability that a given TCR $x$ binds the antigen is $q_0(d \mid x)$. Here, $d \in \{0, 1\}$ is a binary variable, with $d=1$ indicating binding.

We have access to unlabeled sequences, collected via repertoire sequencing. We assume that this data is drawn from the underlying TCR distribution,
\begin{equation}
	\tilde{x}_{1:\tilde{n}} \overset{i.i.d.}{\sim} q_0(x).
\end{equation}
We also have data on sequences that bind the antigen, which comes from screening patient T cells. We assume that this data is drawn as,
\begin{equation}
	x_{1:n} \overset{i.i.d.}{\sim} q_0(x \mid d=1). \end{equation}
This says that the sequences collected in the binding assay come from (1) drawing samples from $q_0$, and then (2) filtering to retain just samples that bind the target antigen ($d=1$).

Our goal is to understand how well a function $g(x)$ predicts binding. For example, $g(x)$ can be the treatment effect estimate from CAIRE, $g(x) = \widehat{\textsc{ate}}(x, \epsilon)$.
We will focus on the area under the receiver operating characteristic curve (ROC AUC) as our performance metric.
Let $(x, d)$ and $(x', d')$ be independent random variables distributed according to $q_0(x, d)$. The ROC AUC is,
\begin{equation}
	\textsc{A}(g) = \mathbb{P}_{q_0}(g(x) \ge g(x') \mid d=1, d'=0).
\end{equation}
In other words, the ROC AUC is the probability that a random positive example will be scored higher than a random negative example.
Standard methods for estimating the ROC AUC rely on samples from $q_0(x \mid d=0)$, that is, negative examples.
The challenge here is that negative examples are unavailable.

Instead, we will evaluate how well $g(x)$ discriminates positive examples from unlabeled examples, and use this evaluation to produce an estimate of the ROC AUC. In particular, we consider the ROC AUC for discriminating positive labeled data from unlabeled data,
\begin{equation}
	\tilde{\textsc{A}}(g) = \mathbb{P}_{q_0}(g(x) \ge g(x') \mid d=1).
\end{equation}
This quantity can be estimated by applying a standard ROC AUC estimator to positive labeled data (binders) and unlabeled data (repertoire sequencing data).
It is related to $\textsc{A}(g)$ as,
\begin{equation} \label{eqn:label_unlabel_auc}
\begin{split}
	\tilde{\textsc{A}}(g) &= \mathbb{E}_{X}[\mathbb{E}_{X'}[\mathbb{I}(g(X) \ge g(X'))] \mid d = 1]\\
	&=  \mathbb{E}_{X}\left[\mathbb{E}_{X'}[\mathbb{I}(g(X) \ge g(X')) \mid d' = 1] q_0(d=1) +  \mathbb{E}_{X'}[\mathbb{I}(g(X) \ge g(X')) \mid d' = 0] q_0(d=0) \,\big|\, d = 1\right]\\
	&= 0.5\, q_0(d=1) + A(g) q_0(d=0).
\end{split}
\end{equation}
where the last line comes from linearity of expectation and from the fact that the ROC AUC for discriminating two variables drawn from the same distribution is 0.5.
We can see that the ROC AUC for discriminating unlabeled examples depends on the ROC AUC for discriminating negative examples.

In practice, we do not have direct data on $q_0(d=1)$. However, \cite{Snyder2020-qg} estimate, based on various lines of evidence, that roughly 0.2\% of the repertoire is involved in binding SARS-CoV-2, so roughly speaking we expect $q_0(d=1) \approx 0.002$ (Figure 4).
This would make $\tilde{\textsc{A}}(g)$ a close approximation of $\textsc{A}(g)$. 
Even if this estimate of $q_0(d=1)$ is inexact, however, note that $\tilde{A}$ still provides a conservative underestimate of $A(g)$.
Moreover, if a predictor $g'$ has better performance than $g$ at discriminating binders, i.e. $A(g') > A(g)$, then the predictor will have better performance at discriminating unlabeled data, i.e. $\tilde{A}(g') > \tilde A(g)$, regardless of the value of $q_0(d=1)$. We can therefore use $\tilde{A}$ to compare binding predictors.

Note it is important in the derivation in \Cref{eqn:label_unlabel_auc} that the unlabeled data and the labeled data both come from the same distribution $q_0$.
If the unlabeled data came from a different distribution -- say, repertoire sequences drawn from a different group of patients -- then \Cref{eqn:label_unlabel_auc} need not hold.

In summary, we focus on $\tilde{A}(g)$ as an evaluation metric for how well effect estimates predict binding.
We report the empirical ROC AUC for discriminating binders (measured by MIRA) from unlabeled sequences (measured by repertoire sequencing).

 \section{Details on COVID Severity Analysis} \label{apx:application}

\subsection{Hyperparameters and training}

The repertoire sequencing data consists of a set of unique sequences along with their weights, which correspond to an estimate of the sequence's frequency in the population of T cells under study.
This gives the empirical distribution estimate of the mature repertoire distribution of the form: $\hat{q}_i(a) = \sum_{j=1}^{m_i} w_{ij} \delta_{a_{ij}}(a)$, where the repertoire sequences are $\{a_{i1}, \ldots, a_{im}\}$ and their corresponding weights are $\{w_{i1}, \ldots, w_{im}\}$ (these weights are normalized to one).
During training, we sample minibatches of sequences by drawing samples from  $\hat{q}_i(a) = \sum_{j=1}^{m_i} w_{ij} \delta_{a_{ij}}(a)$.

We use the same hyperparameter settings as in the semisynthetic experiments (\Cref{tbl:hyperparam}), except we increase the training time to 10 minutes. After preliminary optimization, we fix the dimension of the selection representation and the repertoire representation at $d_r = d_a = 32$, and the kernel size at 9. That is, we do not perform Bayesian optimization.
For the Uncorrected method and Attention CAIRE method, we use the same hyperparameter settings.

\subsection{Non-neural CAIRE} \label{apx:nonneural}

As an additional comparison method, we evaluated a variant of CAIRE in which the non-linear, neural-network-based components were stripped away, and replaced with low-dimensional linear components and biophysical inductive biases. In detail, we modify the model specifications in \Cref{apx:semisynth_arch} as follows:

\begin{itemize}
	\item $h_a(\cdot;\theta)$: Instead of one-hot encoding, amino acids are encoded with their corresponding row in the BLOSUM50 substitution matrix, so that amino acids with similar biophysical properties receive similar encodings~\citepapx{Henikoff1992-ks}. The nonlinearity and max-pooling are removed, leaving a linear convolution: $h_a(a_{ij};\theta)_\ell = \sum_k \mathrm{CONV1D}(a_{ij};\theta)_{k\ell}$.
	\item $h_r(\cdot;\phi)$: Instead of a feedforward neural network, we use a single linear layer.
\end{itemize}

The hyperparameters follow (\Cref{tbl:hyperparam}), except we fix the dimension of the selection representation to a small value ($d_r = d_a = 4$) and fix the size of the kernel to 3. Training time is set to 10 minutes.

\subsection{Held-out sequences} \label{apx:heldout_cov}

In \Cref{sec:effect_est_overview}, as held-out sequences, we use the 11 million repertoire sequences collected from the patients the MIRA experiments were performed on. (These are not the sequences found in the binding assay, but rather those collected via unbiased repertoire sequencing.)

\subsection{Effect uncertainty} \label{apx:effect_uncertainty}

To construct estimates of uncertainty, we interpret the ensemble of 24 models as approximate samples from the Bayesian posterior of the CAIRE model~\citepapx{Lakshminarayanan2017-jy,Wilson2020-wb}. For each candidate sequence $a_\star$ and dosage $\epsilon$, we compute the effect estimated provided by each model in the ensemble, $\widehat{\textsc{ate}}_1, \ldots, \widehat{\textsc{ate}}_{K}$ (where $K = 24$ since there are 24 models in the ensemble). 
We fit a Gaussian to this data, obtaining $\mathrm{Normal}(\hat{\mu}, \hat{\sigma})$ where $\hat{\mu}$ and $\hat{\sigma}$ are the empirical mean and standard deviation respectively.
Here, the average $\hat{\mu} = \frac{1}{K} \sum_{k=1}^K \widehat{\textsc{ate}}_k$ provides a point estimate for the effect, i.e. it is an estimate of the posterior mean.
Moreover, since each $\widehat{\textsc{ate}}_{k}$ is an approximate sample from the posterior over the effect, the distribution $\mathrm{Normal}(\hat{\mu}, \hat{\sigma})$ provides a rough approximation to the full posterior, $\pr(\textsc{ate}(a_\star, \epsilon) \mid \mathcal{D})$, where $\mathcal{D}$ denotes the dataset.
As a measure of significance, we estimate the posterior probability that that the sign of the effect is the opposite of the sign of $\hat{\mu}$, namely $\tilde{p} = \pr_\mathrm{Normal}(x \le 0 \mid |\hat{\mu}|, \hat{\sigma})$.

\subsection{Dosage calculation} \label{apx:dosage}

We use $\epsilon = 0.1$ in all the effect estimates we report.
We chose this value based on rough calculation of the dosage of existing TCR-based therapies.
In a review of clinical trials of TCR-engineered T cell therapy, \citeapx{Baulu2023-vt} report several trials that use dosages as high as $10^{11}$ cells per patient. There are roughly $10^{12}$ T cells in an adult human~\citepapx{Lythe2016-zh}.
This suggests that it is tractable to intervene on TCR repertoires such that about 10\% of the repertoire is a chosen sequence, giving $\epsilon = 0.1$.

\subsection{Outcome predictive performance}
\begin{figure}
\centering
\begin{subfigure}[t]{0.32\textwidth}
	\includegraphics[width=\textwidth]{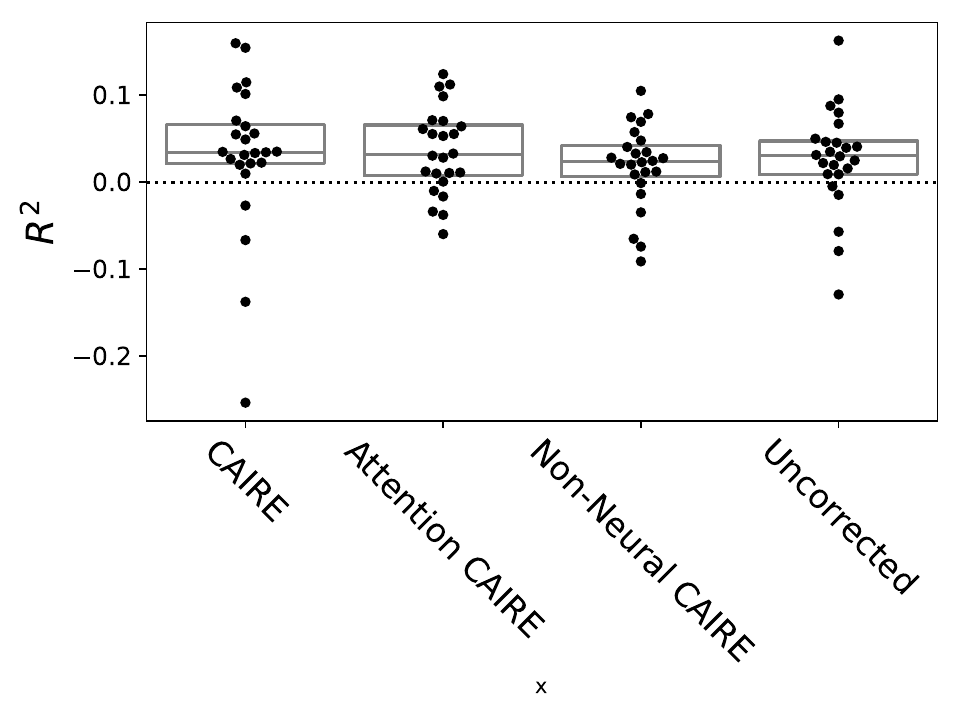}
	\caption{} \label{fig:outcome_r2_distributions}
\end{subfigure}
\begin{subfigure}[t]{0.32\textwidth}
	\includegraphics[width=\textwidth]{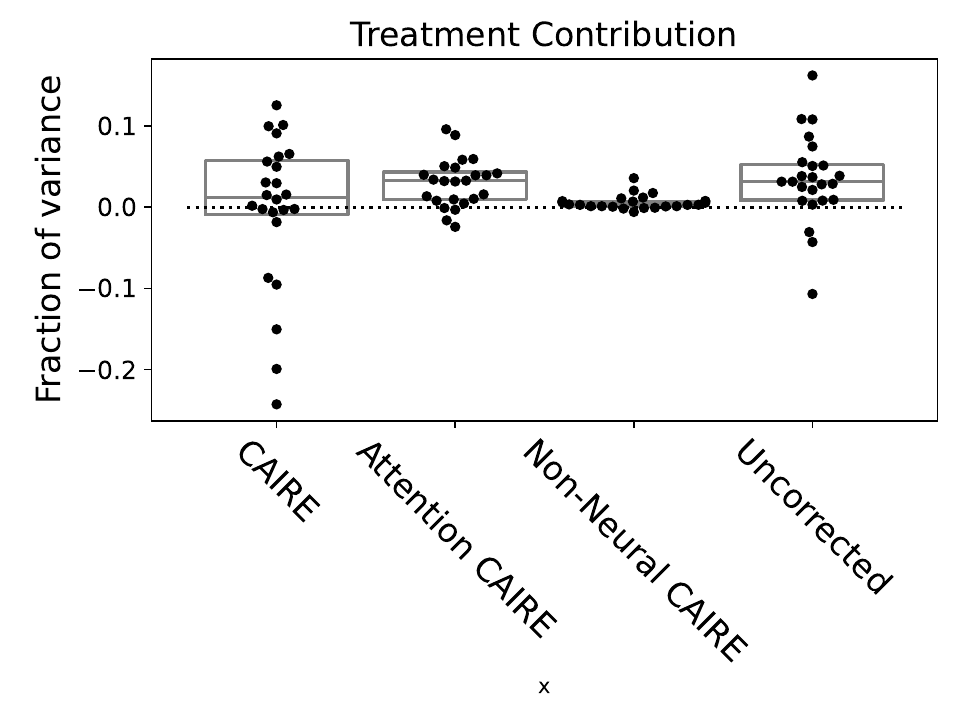}
	\caption{} \label{fig:treatment_explain_distributions}
\end{subfigure}
\begin{subfigure}[t]{0.32\textwidth}
	\includegraphics[width=\textwidth]{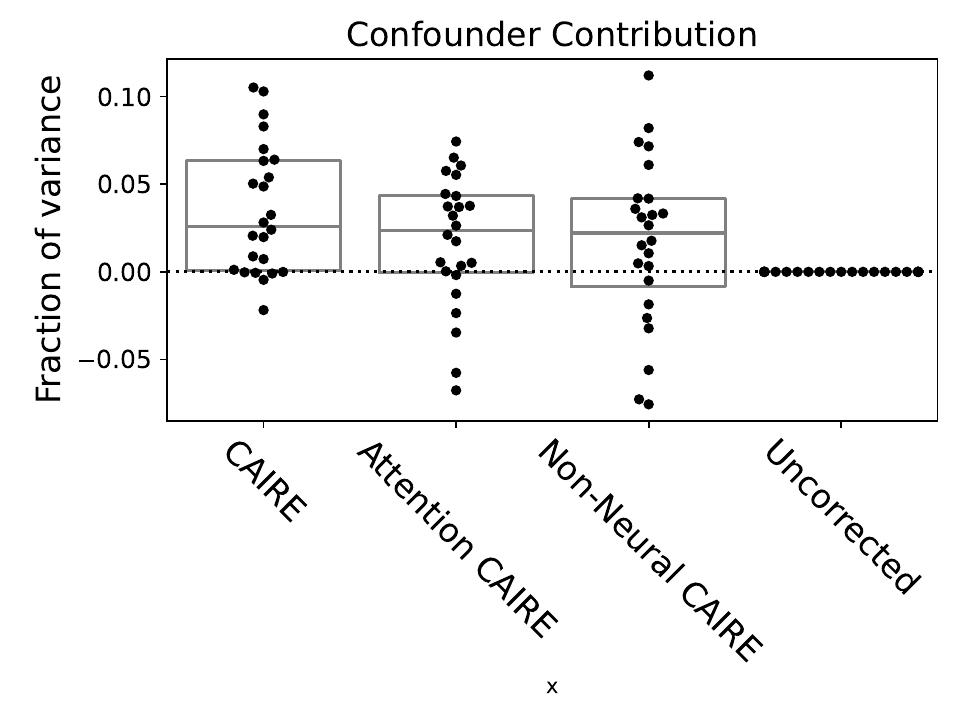}
	\caption{} \label{fig:confound_explain_distributions}
\end{subfigure}
\caption{\textbf{Predictive performance across the COVID model ensemble}. Distribution across the model ensemble of the coefficient of determination $R^2$ with the outcome (\Cref{fig:outcome_r2_distributions}), variance in the outcome explained by the treatment term (\Cref{fig:treatment_explain_distributions}) and variance in the outcome explained by the confounder term (\Cref{fig:confound_explain_distributions}).} \label{fig:predictive_distribution}
\end{figure}

\Cref{tbl:predictive_perf} and \Cref{fig:predictive_distribution} report the performance of CAIRE and comparison models at predicting the outcome $y$ on held-out data.
We find CAIRE, its version with attention, and the uncorrected method show similar performance. Non-Neural CAIRE performs considerably worse.
 \Cref{tbl:predictive_perf} and \Cref{fig:predictive_distribution} also report the variance in $y$ explained by the treatment term of CAIRE, namely $\gamma_a^\top (\mathbb{E}_{\hat{q}^a_i}[h_a(A ;\theta)] - W \cdot \rho_i - B)$, and the variance explained by the confounder term, $\gamma_r^\top \rho_i$.
 Both are also calculated on held-out data.
 
\begin{table}
\centering
\caption{\textbf{Predictive performance of COVID models.} The first column gives the coefficient of determination between the model's predictions and the outcome, while the second and third columns give the variance explained by the treatment and confounder terms respectively. All values are calculated on heldout data. For each, we report the median value across the model ensemble, along with a 95\% confidence interval of the median (calculated based on the quantiles of the binomial distribution). We use the median as a robust statistic, due to outliers in the data; see \Cref{fig:predictive_distribution} for the raw data.} \label{tbl:predictive_perf}	
\begin{tabular}{cccc}
& Outcome $R^2$ & Treatment var. explained & Confounder var. explained\\
\hline
CAIRE & 0.035 [0.022, 0.059] & 0.012 [0.00, 0.05] & 0.026 [0.01, 0.06]\\
Attention CAIRE & 0.032 [0.010, 0.062] & 0.032 [0.01, 0.04] & 0.024 [0.00, 0.04]\\
Non-Neural CAIRE & 0.024 [0.011, 0.036] & 0.003 [0.00, 0.01] & 0.022 [0.00, 0.04]\\
Uncorrected & 0.031 [0.014, 0.046] & 0.031 [0.02, 0.05] & 0.000 [0.00, 0.00]
\end{tabular}
\end{table}

\subsection{Confounder representation} \label{sec:confound_rep}

We sought to determine whether the latent representation of fitness learned by CAIRE, $\rho_i$, contained information about other patient covariates.
In particular, we investigated whether the learned representation contained information about patient age, gender or ethnicity.
Such demographic information is often used to correct for confounding, when alternative strategies are unavailable.

To determine whether the latent fitness representation contains information about patient age, we train a Bayesian ridge regression model to predict patient age from the representation $\rho_i$.
We focus just on the $n = 505$ patients for which age information was available (non-missing).
We hold out 12.5\% of patients as test set.
We repeat this process for the latent representations from each of the ensemble of 24 CAIRE models, holding out a randomly chosen set of patients each time.
On average over the ensemble, we find that the coefficient of determination ($R^2$) on the test set is 0.004, with a standard error of the mean of 0.009.

We next consider gender, training a logistic regression model to predict whether a patient is male or female. 
There were $n = 506$ patients with gender information available.
The average accuracy over the ensemble of CAIRE models was 0.47 (standard error 0.01), while the average percentage of the test set that was female was 0.51.

Finally we consider ethnicity. We focus on predicting whether or not a patient is Caucasian, since other groups make up just a small percentage of the data (\Cref{sec:patient_demo}).
Using logistic regression, we find that the average accuracy over the ensemble was 0.84 (s.e. 0.01), while the average Caucasian fraction of the test set was 0.84.

Overall, then, we find no evidence to support the idea that the latent fitness representation reflects patient demographics.
Note, however, that it also does not appear to the case that the latent fitness representation is arbitrary or purely random.
First, it is predictive of the outcome variable (\Cref{tbl:predictive_perf}, \Cref{fig:predictive_distribution}).
Second, we computed the Euclidean distance matrix between patients' representations, that is $M_{ij} = \|\rho_i - \rho_j\|_2$ and found that it was stable across the ensemble of CAIRE models.
In particular, to compare two models in the ensemble, we examined the Pearson correlation between the non-diagonal entries of their distance matrices, $M$.
On average across pairs of models from the ensemble, the Pearson correlation was 0.70 (all p values for testing non-correlation were below floating point precision).

\begin{figure}
\centering
\begin{subfigure}[t]{0.32\textwidth}
	\includegraphics[width=\textwidth]{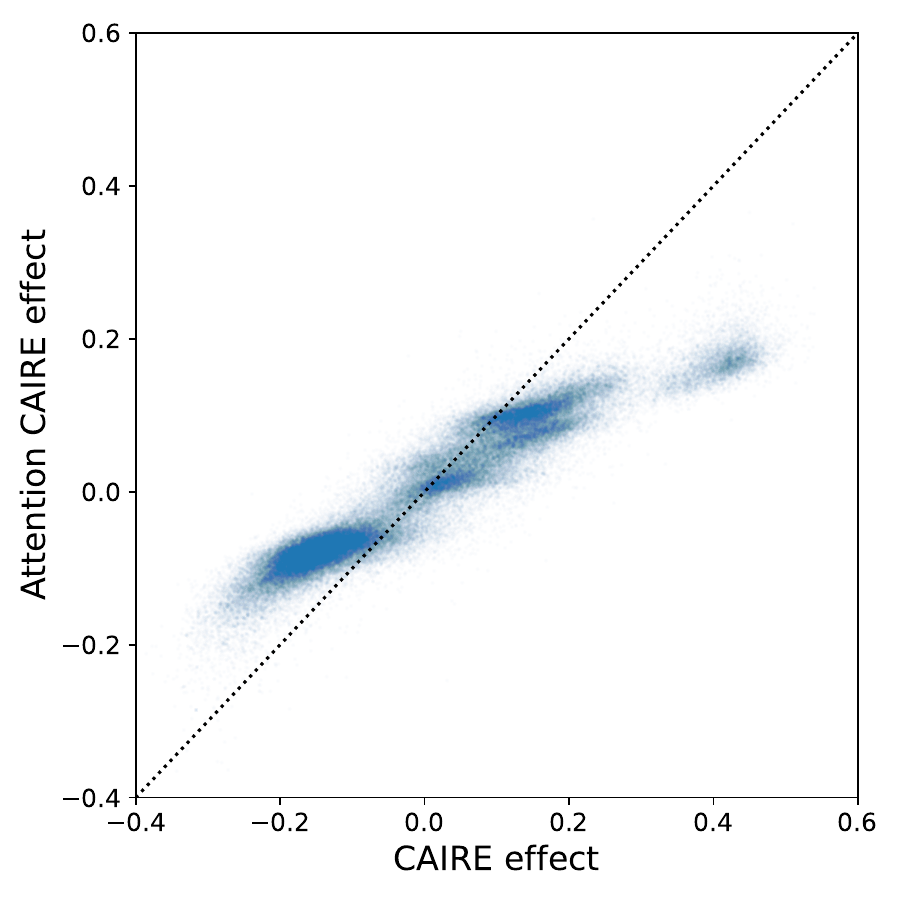}
	\caption{Pearson r: 0.95} \label{fig:effect_correlation_attention}
\end{subfigure}
\begin{subfigure}[t]{0.32\textwidth}
	\includegraphics[width=\textwidth]{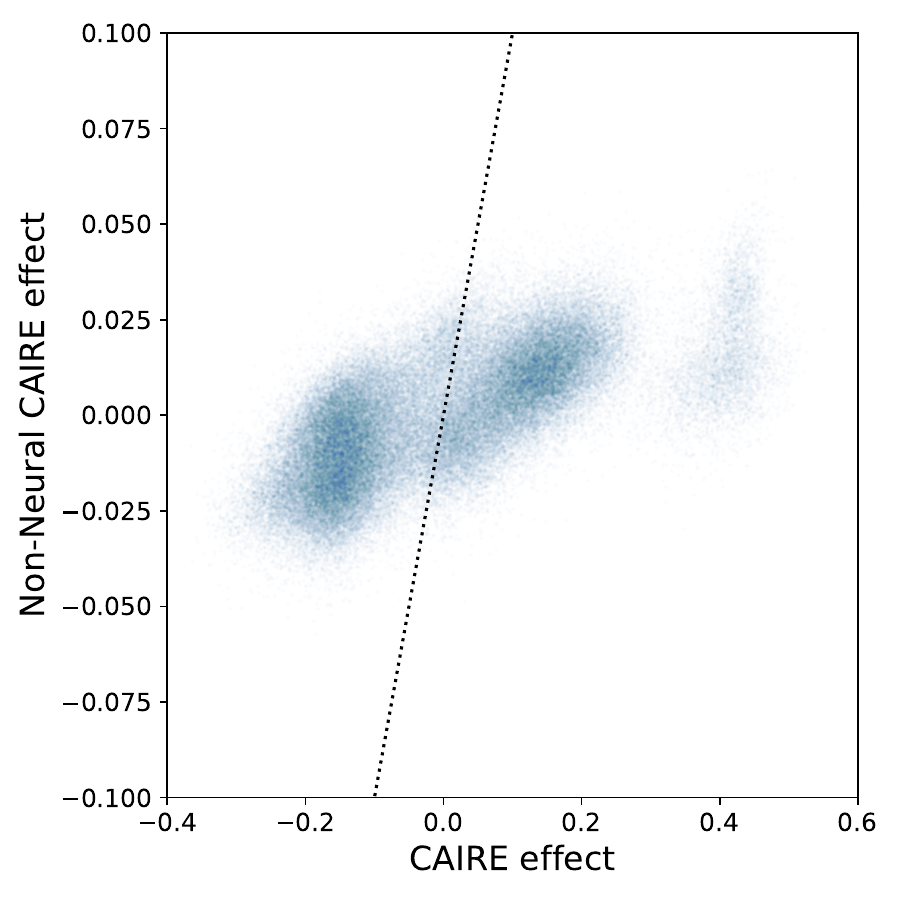}
	\caption{Pearson r: 0.69} \label{fig:effect_correlation_nonneural}
\end{subfigure}
\begin{subfigure}[t]{0.32\textwidth}
	\includegraphics[width=\textwidth]{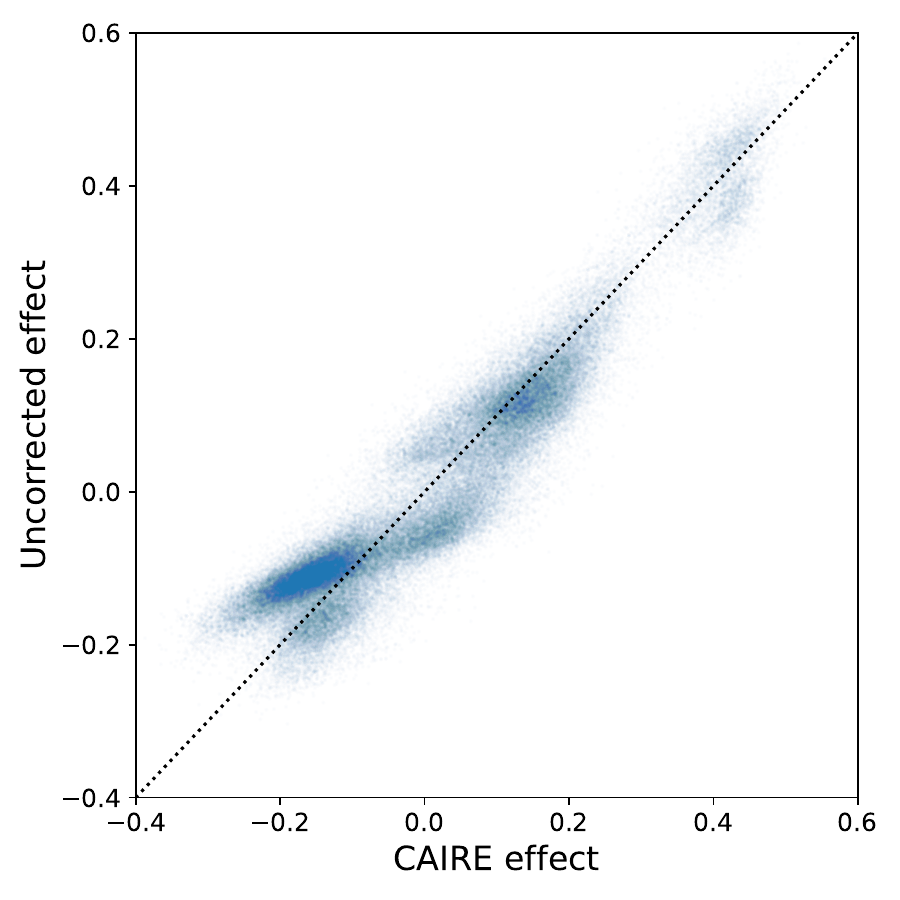}
	\caption{Pearson r: 0.94} \label{fig:effect_correlation_uncorrected}
\end{subfigure}
\caption{\textbf{Comparison of COVID effect estimates from different methods.} We compare effect estimates from CAIRE with those from Attention CAIRE (\Cref{fig:effect_correlation_attention}), Non-Neural CAIRE (\Cref{fig:effect_correlation_nonneural}) and the Uncorrected method (\Cref{fig:effect_correlation_uncorrected}). Scatter plot shows 100,000 sequences randomly subsampled from the 11 million held out TCR sequences. The dotted black line shows the identity, where the two effects are equal.} \label{fig:effect_correlation}
\end{figure}

\subsection{Binding prediction evaluation} \label{apx:binding_comparison}

To evaluate a model's ability to predict binding, we calculate the ROC AUC for separating binders from repertoire sequences, as described in \Cref{apx:application_eval}.
We use MIRA binding data together with repertoire sequencing data from the same patients (\Cref{apx:mira}).
We first consider separating sequences found to bind class I MHC epitopes from repertoire sequences drawn from the same set of patients (\Cref{tbl:binding_predict}, column: Class I (all)).
Then, we consider the same metric for class II MHC epitopes (\Cref{tbl:binding_predict}, column: Class II (all)).
We also repeat these calculations just for those sequences that the model is confident has an effect ($\tilde p < 0.05$) (\Cref{tbl:binding_predict}, columns: Class I and II (signif.)).
We report the AUC standard error as $1/2 \sqrt{\mathrm{min}(B,U)}$, where $B$ is the number of binders and $U$ is the number of repertoire sequences with unknown binding; note this is considered a somewhat conservative estimate of the true uncertainty~\citepapx{Cortes2004-rv}.

For class I, we find that CAIRE's estimates are predictive of binding. In particular, we find that sequences with more negative effects are more likely to bind an antigen (as indicated by a negative sign in \Cref{tbl:binding_predict}).
This is in line with the idea that sequences have negative effects because they give rise to overactive immune responses (\Cref{sec:application_balance}): TCRs that bind more may be more likely to create an extreme immune response and damage healthy tissue.

For class II, we do not find clear evidence that CAIRE's effect estimates are predictive of binding (\Cref{tbl:binding_predict}), though note that less data is available for class II than class I (4,130 sequences bind class II, identified from binding experiments on 6 patients; 32,770 sequences bind class I, identified from binding experiments on 66 patients).

One possible concern is that these evaluation metrics pool data from many patients. Due to experimental variability, it may be the case that the distribution over sequences in the pooled set of repertoire sequences is slightly different from that in the pooled set of binders.
As a robustness check, we also calculated the ROC AUC for discriminating binders (both class I and class II) from repertoire sequences \textit{within} each patient, and then take the average ROC AUC across patients.
The results confirm that CAIRE's estimates are predictive of binding (\Cref{tbl:binding_predict}, column: Patients (all)).

\subsection{Other therapeutic approaches} \label{apx:cov_other_therapies}

We further considered the implications of CAIRE's estimates for therapeutic approaches that deplete T cells with TCRs that bind a specific antigen~\citepapx{Moisini2008-so,Norville2023-eb}.
Here, a key design question is how to select an antigen that interacts with pathogenic TCRs.
We therefore looked for antigens studied in the MIRA binding experiments that were enriched for TCRs with significant negative effects.
We did not find any (binomial test, Benjamini-Hochberg adjusted p-value threshold of 0.05).
This suggests that it is difficult to target and deplete the population of TCRs with negative effects, without depleting even more TCRs that have positive effects on clinical outcomes. 
To develop therapies that deplete TCRs with negative effects, it may be necessary to look outside the SARS-CoV-2 genome, for instance at human antigens.

\begin{table}
\caption{\textbf{Binding prediction performance.} ROC AUC values, with standard error. The sign of the prediction is in parenthesis: $(-)$ indicates that sequences with more negative effects are more likely to bind, and $(+)$ indicates that sequences with more positive effects are more likely to bind. NA values are reported in situations where no sequences have significant effects ($\tilde p < 0.05$). Note that for all class II values, 0.5 is within the 95\% confidence interval (1.96$\times$ standard error).} \label{tbl:binding_predict}
\begin{tabular}{cccccc}
& Class I (all) & Class I (signif.) & Class II (all) & Class II (signif.) \\
\hline
CAIRE & 0.563$\pm$0.003 (-) & 0.604$\pm$0.027 (-) & 0.505$\pm$0.008 (-) & 0.564$\pm$0.058 (+) \\
Attention CAIRE & 0.568$\pm$0.003 (-) & 0.526$\pm$0.109 (+) & 0.507$\pm$0.008 (-) & 0.651$\pm$0.500 (+) \\
Non-Neural CAIRE & 0.566$\pm$0.003 (-) & NA & 0.502$\pm$0.008 (+) & NA \\
Uncorrected & 0.556$\pm$0.003 (-) & 0.534$\pm$0.020 (-) & 0.513$\pm$0.008 (-) & 0.557$\pm$0.052 (-) \\
\end{tabular}
\begin{tabular}{cc}
& Patients (all) \\
\hline
	CAIRE & 0.553$\pm$0.009 (-)\\
Attention CAIRE & 0.557$\pm$0.008 (-)\\
Non-Neural CAIRE & 0.556$\pm$0.008 (-)\\
Uncorrected & 0.549$\pm$0.007 (-)\\
\end{tabular}

\end{table}

\begin{figure}
\centering
\begin{subfigure}[c]{0.45\textwidth}
\includegraphics[width=\textwidth]{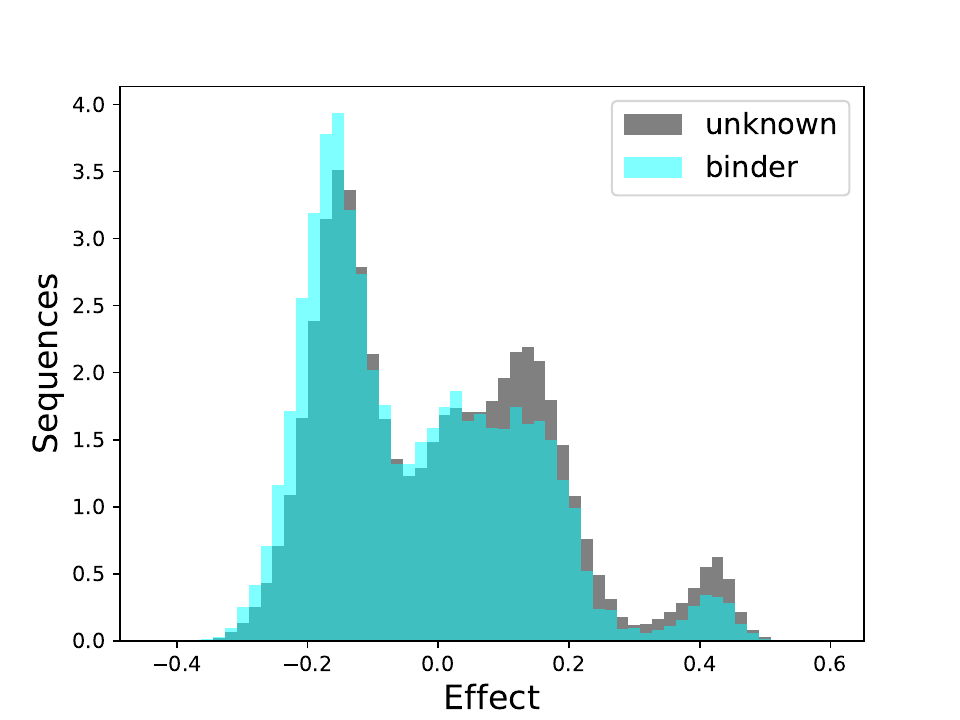}
\caption{All} \label{fig:binder_unknown_distr}
\end{subfigure}
\begin{subfigure}[c]{0.45\textwidth}
\includegraphics[width=\textwidth]{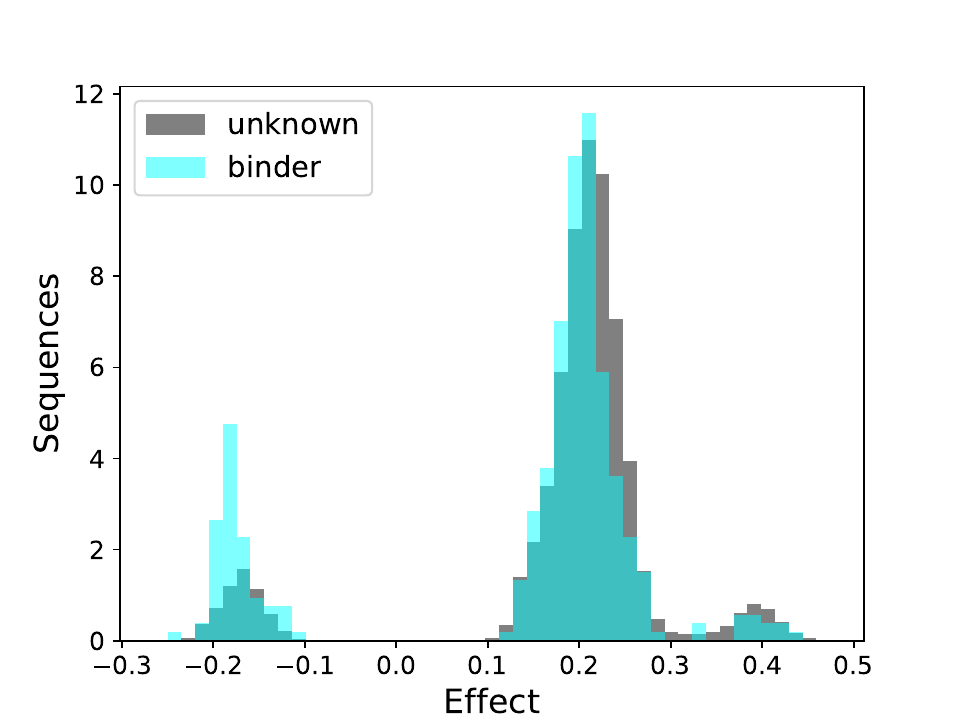}
\caption{Significant} \label{fig:binder_unknown_distr_significant}
\end{subfigure}
\caption{\textbf{Effect distribution across binders.} Distribution of estimated effects among sequences found to bind class I antigens in the MIRA assay (binders, blue) versus among general repertoire sequences, from the same patients the binders were found in (unknown binders, gray). (a) All sequences. (b) Only sequences with significant non-zero effects, $\tilde p < 0.05$.
} \label{fig:binder_unknown_distr_comparison}
\end{figure}

\begin{figure}
	\centering
\begin{subfigure}[c]{0.45\textwidth}
\includegraphics[width=\textwidth]{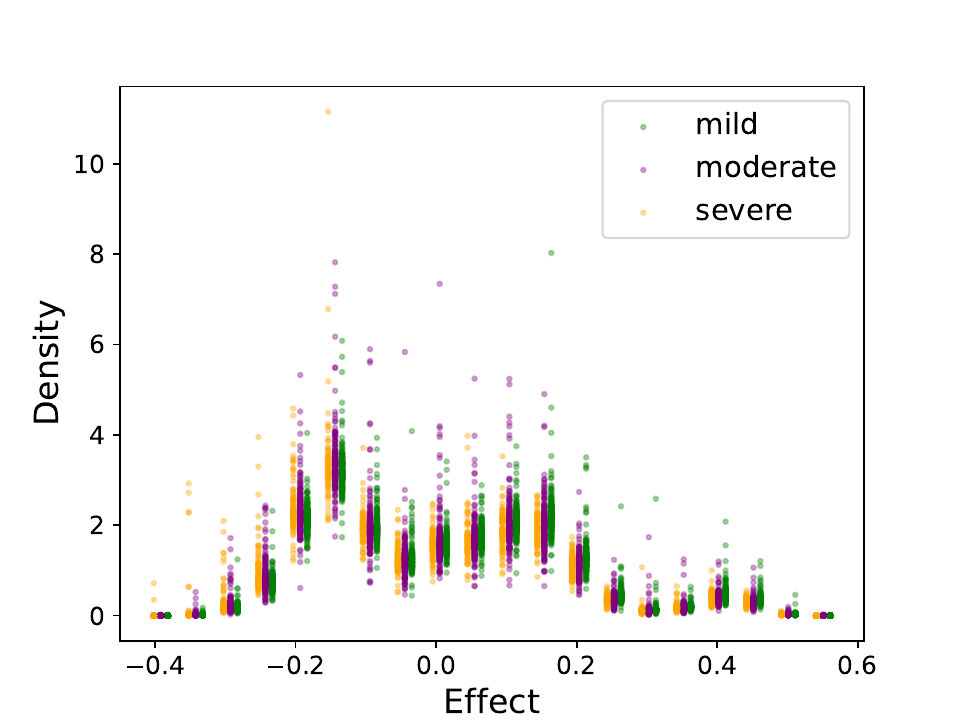}
\caption{} \label{fig:effect_distr_by_outcome_per_patient}
\end{subfigure} 
\begin{subfigure}[c]{0.45\textwidth}
\includegraphics[width=\textwidth]{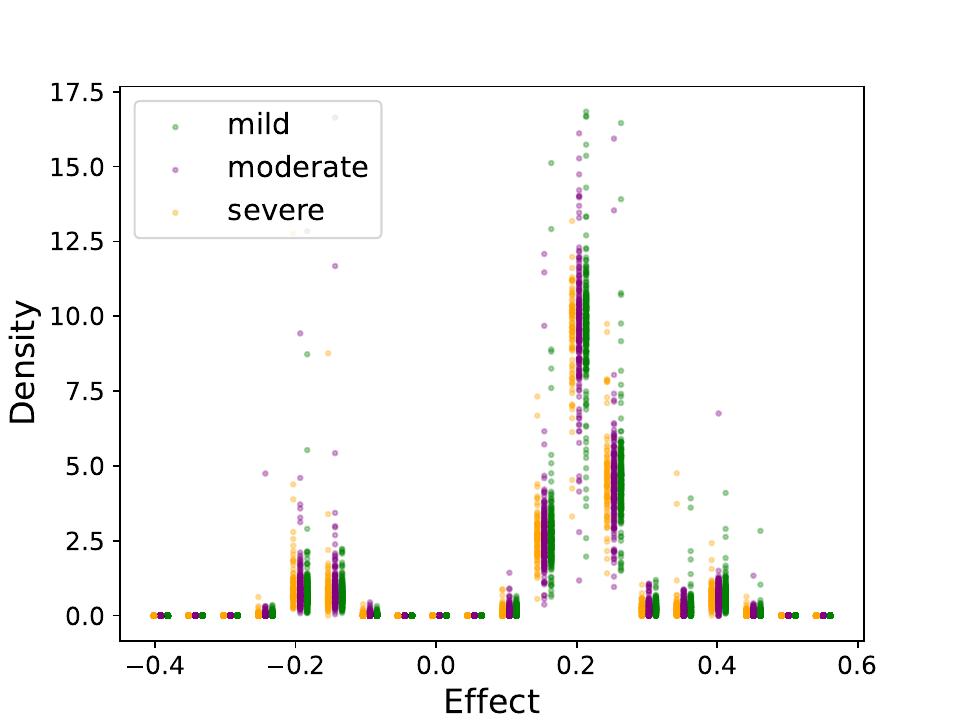}
\caption{} \label{fig:effect_distr_by_outcome_per_patient_signif}
\end{subfigure}
\begin{subfigure}[c]{0.45\textwidth}
\includegraphics[width=\textwidth]{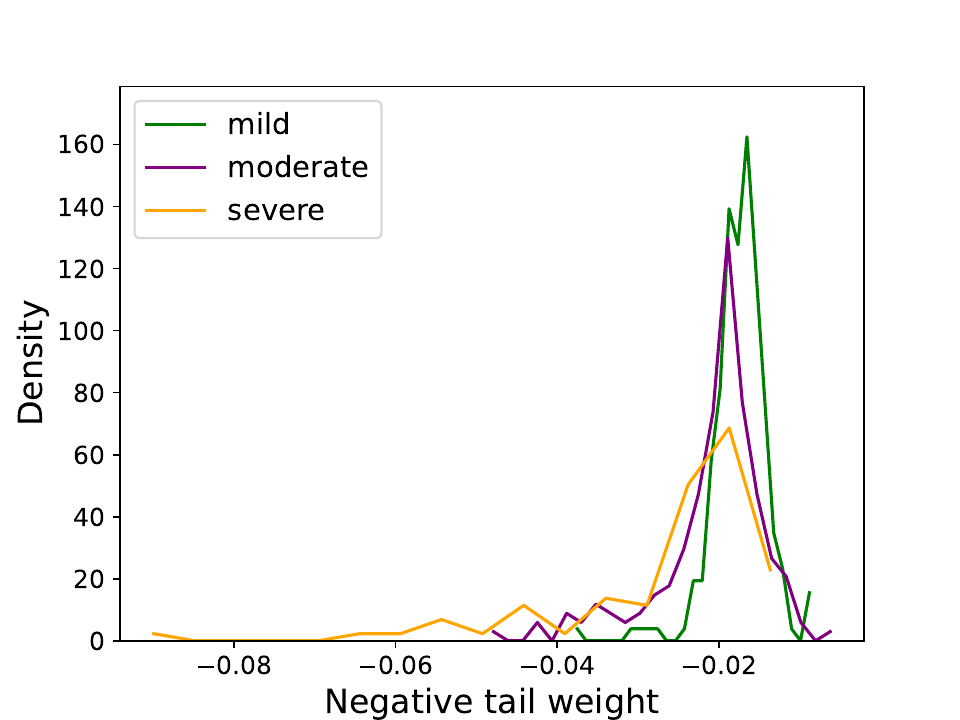}
\caption{} \label{fig:negative_tail_weight_by_outcome}
\end{subfigure}
\caption{\textbf{Distribution of effects across patients.} (a) The distribution of TCR effects across individual patient repertoires. Each point at interval $\mathcal{I}$ is an estimate of $\mathbb{P}_{A \sim q^a_i}[ \textsc{ate}(A;0.1) \in \mathcal{I}]$ for a patient $i$.
(b) The distribution of TCR effects among TCRs with significant effects, across individual patient repertoires. Each point is an estimate of $\mathbb{P}_{A \sim q^a_i}[ \textsc{ate}(A;0.1) \in \mathcal{I} \mid \tilde{p} < 0.05]$ for a patient $i$.
(c) Distribution of the burden of repertoire sequences with negative effects, across patients with different outcomes. Each point at interval $\mathcal{I}$ is an estimate of $\mathbb{P}_{Q^a}[\mathbb{E}_{A \sim Q^a}[\textsc{ate}(A;0.1) \mathbb{I}(\textsc{ate}(A;0.1) < -0.2)]] \in \mathcal{I} \mid y]$ for an outcome $y \in \{-1, 0, +1\}$.}
\end{figure}

\begin{table}
  \centering
  \caption{\textbf{TCRs with the largest positive effects.} Effect: the effect estimate from CAIRE, for $\epsilon=0.1$. CDR3$\beta$: the CDR3$\beta$ region of the TCR. Framework: the V and J gene for the TCR. Epitope(s): the SARS-CoV-2 epitope(s) the TCR was found to bind. Some entries have multiple epitopes because the experiment pooled them, so it is unknown precisely which of epitope the TCR binds. The table is divided into TCRs that were found to bind class I and class II MHC epitopes.} \label{tab:top_tcrs}
  \begin{tabular}{cccc}
  \hline
    Effect & CDR3$\beta$ sequence & Framework & Epitope(s)\\ \hline
    &&& \textit{Class I}
\\ \hline
0.436 & CASTKEGRVATNEKLFF & V28-01+J01-04 & HTTDPSFLGRY\\0.427 & CASRRGQENEKLFF & V05-04+J01-04 & \makecell[tc]{TVLSFCAFA \\ VLSFCAFAV} \\0.426 & CASSLATTGENEKLFF & V05-06+J01-04 & \makecell[tc]{LLDDFVEII \\ LLLDDFVEI} \\0.414 & CASCETHPVGYPNEKLFF & V27-01+J01-04 & HTTDPSFLGRY \\0.400 & CASSDRQGTNEKLFF & V27-01+J01-04 & HTTDPSFLGRY \\0.397 & CASSITGRANEKLFF & V19-01+J01-04 & \makecell[tc]{ILGTVSWNL \\ SNEKQEILGTVSW} \\0.396 & CASSYRAGGNEKLFF & V06-05+J01-04 & \makecell[tc]{LSPRWYFYY \\ SPRWYFYYL} \\0.386 & CAWKSEDRQGFNEKLFF & V30-01+J01-04 & HTTDPSFLGRY \\0.383 & CASSPNQQGTNEKLFF & V27-01+J01-04 & \makecell[tc]{STGSNVFQTR \\ TGSNVFQTR \\ VYSTGSNVF} \\0.381 & CASSDDQVGTANEKLFF & V18-01+J01-04 & YYRRATRRIR \\0.328 & CASSLTGIEKLFF & V27-01+J01-04 & HTTDPSFLGRY \\0.328 & CASSQKTGGREKLFF & V04-01+J01-04 & \makecell[tc]{LLDDFVEII \\ LLLDDFVEI} \\\hline
&&& \textit{Class II}
\\ \hline
0.409 & CASSQDQTGDNEKLFF & V03-01/02+J01-04 & \makecell[tc]{\small EDLKFPRGQGVPINTNSSP \\ \small PNNTASWFTALTQHGKEDL \\ \small QGVPINTNSSPDDQIGYYR \\ \small SSPDDQIGYYRRATRRIRG \\ \small TALTQHGKEDLKFPRGQGV} \\0.409 & CASSRTGGNEKLFF & V11-02+J01-04 & \makecell[tc]{\small ASFSTFKCYGVSPTKLNDL \\ \small GDEVRQIAPGQTGKIADYN \\ \small NDLCFTNVYADSFVIRGDE \\ \small PGQTGKIADYNYKLPDDFT \\ \small YADSFVIRGDEVRQIAPGQ \\ \small YGVSPTKLNDLCFTNVYAD} \\0.395 & CAISDTTGRGANEKLFF & V10-03+J01-04 & \makecell[tc]{\small GRCDIKDLPKEITVATSRT \\ \small LRIAGHHLGRCDIKDLPKE \\ \small PKEITVATSRTLSYYKLGA \\ \small SRTLSYYKLGASQRVAGDS} \\0.391 & CASSQQPTTNEKLFF & V05-05+J01-04 & \makecell[tc]{\small FLIVAAIVFITLCFTLKRKTE \\ \small SPKLFIRQEEVQELYSPIFL}  \\0.380 & CASSQVTIANEKLFF & V04-01+J01-04 & \makecell[tc]{\small DFGGFNFSQILPDPSKPSK \\ \small DLLFNKVTLADAGFIKQYG \\ \small LADAGFIKQYGDCLGDIAA \\ \small PSKRSFIEDLLFNKVTLAD \\ \small QILPDPSKPSKRSFIEDLL \\ \small QYGDCLGDIAARDLICAQK}  \\\hline
  \end{tabular}
\end{table}

\begin{figure}
\centering
	\includegraphics[width=\textwidth]{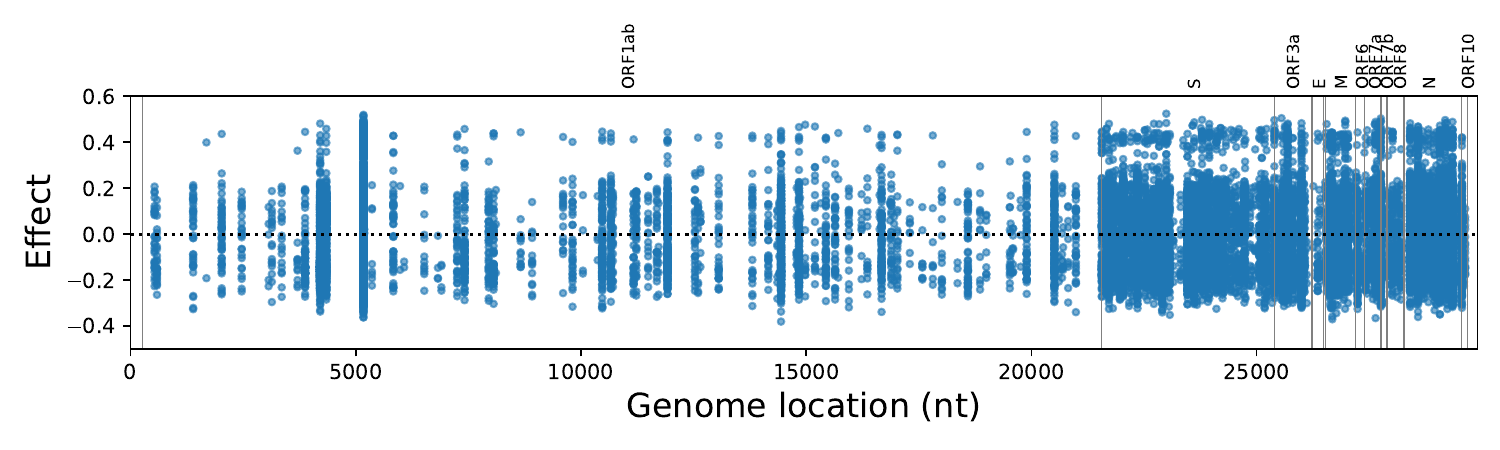}
\caption{\textbf{TCR effects across the SARS-CoV-2 genome.} Same as \Cref{fig:binder_significant_effect_loc}, except including all binders, not just those with significant effects.} \label{fig:binder_all_effect_loc}
\end{figure}

\begin{table}
  \centering
  \caption{\textbf{Antigens that bind TCRs with strong positive effects.} We examined the causal effect of the TCRs that bind each epitope. Focusing just on TCRs with a significant causal effect, we found that these antigens were specifically enriched for TCRs with beneficial effects on patient outcomes (binomial test, Benjamini-Hochberg adjusted p-value below 0.05). Mean effect sign: average sign of the estimated effect of TCRs that bind the epitope(s). Epitope(s): the specific SARS-CoV-2 antigen. Some entries have multiple overlapping epitopes because the experiment pooled these epitopes, so it is unknown precisely which of these epitopes the TCR binds. Note all discovered epitopes here are class I. Gene: the SARS-CoV-2 protein the epitope comes from (S: spike, N: nucleocapsid).} \label{tab:top_antigens}
  \begin{tabular}{ccc}
  \hline
    Mean effect sign & Epitope(s) & Gene \\ \hline
    0.40 & HTTDPSFLGRY & ORF1ab \\ 1.00 & \makecell[tc]{APHGVVFL,APHGVVFLHV\\GVVFLHVTY,VVFLHVTYV} & S \\ 1.00 & LSPRWYFYY,SPRWYFYYL & N \\ 1.00 & AYKTFPPTEPK,KTFPPTEPK & N \\ 1.00 & \makecell[tc]{APSASAFFGM,AQFAPSASA\\ASAFFGMSR,SASAFFGMSR} & N \\ \hline
  \end{tabular}
\end{table}

\end{document}